\documentclass[pdflatex,sn-mathphys-num,iicol]{sn-jnl}

\usepackage{graphicx}%
\usepackage{multirow}%
\usepackage{amsmath,amssymb,amsfonts}%
\usepackage{amsthm}%
\usepackage{mathrsfs}%
\usepackage[title]{appendix}%
\usepackage{xcolor}%
\usepackage{textcomp}%
\usepackage{manyfoot}%
\usepackage{booktabs}%
\usepackage{algorithm}%
\usepackage{algorithmicx}%
\usepackage{algpseudocode}%
\usepackage{listings}%

\usepackage[utf8]{inputenc}
\usepackage[T1]{fontenc}
\usepackage{tabularx}
\usepackage[table]{xcolor}
\usepackage{longtable}
\usepackage{subcaption}

\usepackage{stfloats}

\usepackage{xurl}
\usepackage{hyperref}

\usepackage{scalerel}
\usepackage{tikz}
\usetikzlibrary{svg.path}

\definecolor{orcidlogocol}{HTML}{A6CE39}
\tikzset{
  orcidlogo/.style={
    target length=10pt,
    2.1.0,
    svg={M256,128c0,70.7-57.3,128-128,128S0,198.7,0,128S57.3,0,128,0S256,57.3,256,128z 
         M91.1,190.1H109v-73.1H91.1V190.1z M100,107.5c6.4,0,11.6-5.2,11.6-11.6S106.4,84.3,100,84.3s-11.6,5.2-11.6,11.6 
         S93.6,107.5,100,107.5z M180.8,118.4c-6.8-5.3-15.5-8-25.7-8h-32.9v79.7h34.7c10,0,18.4-2.5,25.1-7.6 
         c6.7-5.1,10.1-12.4,10.1-22.1C192,132.8,188.1,124.1,180.8,118.4z M170.8,163.6c-3.1,3-7.9,4.5-14.4,4.5h-15.3v-38.3h16.2 
         c6.1,0,10.7,1.4,13.7,4.3c3,2.9,4.5,7.1,4.5,12.7C175.6,154.5,174,160,170.8,163.6z}
  }
}

\definecolor{orcidlogocol}{HTML}{A6CE39}
\newcommand\orcidicon[1]{%
  \href{https://orcid.org/#1}{%
    \begin{tikzpicture}[baseline=-0.6ex] %
      \draw[orcidlogocol, fill=orcidlogocol] (0,0) circle [radius=0.16];
      \node at (0,0) {\textcolor{white}{\fontsize{3.5}{4}\selectfont \textbf{iD}}};
    \end{tikzpicture}%
  }%
}

\usepackage{hyperref}

\theoremstyle{thmstyleone}%
\theoremstyle{thmstyletwo}%

\theoremstyle{thmstylethree}%

\begin{document}

\title[Perception of Social Robots as Communication Partners in Healthcare for Older Adults]{Perception of Social Robots as Communication Partners \\in Healthcare for Older Adults}

\author*[1,2]{\fnm{Hana} \sur{Yamamoto}\orcidicon{0009-0008-7408-0046}}\email{hana.yamamoto@kit.edu}
\author[3,4]{\fnm{Carlotta Julia} \sur{Mayer}\orcidicon{0000-0003-3102-3566}}\email{carlotta.mayer@psychologie.uni-heidelberg.de}
\author[3]{\fnm{Charlotte} \sur{Raithel}\orcidicon{0009-0005-3470-6858}}\email{charlotte.raithel@med.uni-heidelberg.de}
\author[5]{\fnm{Theresa} \sur{Buchner}\orcidicon{0009-0002-8296-3507}}\email{theresa.buchner@med.uni-heidelberg.de}
\author[5]{\fnm{Christian} \sur{Werner}\orcidicon{0000-0003-0679-3227}}\email{christian.werner@med.uni-heidelberg.de}
\author[2]{\fnm{Yasuhisa} \sur{Hirata}\orcidicon{0000-0002-5931-0471}}\email{yasuhisa.hirata.b1@tohoku.ac.jp}
\author[3]{\fnm{Monika} \sur{Eckstein}\orcidicon{0000-0002-1846-4992}}\email{monika.eckstein@med.uni-heidelberg.de}
\author[1]{\fnm{Katja} \sur{Mombaur}\orcidicon{0000-0003-1353-0943}}\email{katja.mombaur@kit.edu}


\affil*[1]{\orgdiv{Institute for Anthropomatics and Robotics}, \orgname{Karlsruhe Institute of Technology}, \orgaddress{\street{Adenauerring 12}, \city{Karlsruhe}, \postcode{76131}, \state{Baden-W\"urttemberg}, \country{Germany}}}

\affil[2]{\orgdiv{Department of Robotics}, \orgname{Tohoku University}, \orgaddress{\street{Aramaki Aza Aoba, Aoba-ku}, \city{Sendai}, \postcode{980-8579}, \state{Miyagi}, \country{Japan}}}

\affil[3]{\orgdiv{Institute of Medical Psychology}, \orgname{Heidelberg University}, \orgaddress{\street{Bergheimer Str. 20}, \city{Heidelberg}, \postcode{69115}, \state{Baden-W\"urttemberg}, \country{Germany}}}

\affil[4]{\orgdiv{Institute of Psychology}, \orgname{Heidelberg University}, \orgaddress{\street{ Haupt Str. 47-51}, \city{Heidelberg}, \postcode{69117}, \state{Baden-W\"urttemberg}, \country{Germany}}}

\affil[5]{\orgdiv{Geriatric Center, Medical Faculty Heidelberg}, \orgname{Heidelberg University}, \orgaddress{\street{Rohrbacher Str. 149}, \city{Heidelberg}, \postcode{69126}, \state{Baden-W\"urttemberg}, \country{Germany}}}

\abstract{
Addressing the global caregiver shortage through socially assistive robots necessitates a deep understanding of their psychological and physiological impacts on older adults during human-robot interaction (HRI). This study addresses whether social robots can serve as effective interaction partners compared to humans, and if "positive prompts" can similarly enhance these interactions. We conducted a comparative study with 35 participants (aged 70+). Our multi-modal analysis, integrating facial expression data, heart rate variability, and subjective questionnaires, revealed no significant differences in overall stress levels between human and robot interactions. Facial expression analysis confirmed that the robot was accepted as a valid interaction partner, while physiological data showed slightly lower heart rates during robot interactions, suggesting a more relaxed state compared to human-led sessions. These findings indicate that social robots can engage older adults without inducing psychological strain and are capable of alleviating caregiver burden by performing structured tasks, such as health-sensing surveys. Future work should address the identified "appearance-content mismatch" in robot design to facilitate even more natural and effective interactions.}

\maketitle
\section{Introduction}\label{sec1}
\subsection{Social Background}\label{sec1-1}
The global population is rapidly aging, and the declining birth rate and the aging population are particularly problematic in developed countries \cite{population}. In Germany, the target region of this study, the need for nursing care is increasing as the population ages. In 2023, the number of nursing workers is expected to reach approximately 5.7 million and will continue to grow \cite{aging}. And the demand for nursing staff is expected to increase by 30\% over the next 30 years, from 2019 to 2049. Meanwhile, due to a declining labor force, a shortage of up to 700,000 nursing staffs is predicted by 2049 \cite{demand_supply}. Against this backdrop, the median monthly salary of full-time workers in the healthcare and nursing care sector in Germany increased by 43.1\% between 2014 and 2024, with a 61.6\% increase in the care sector for older adults \cite{elderly_care}. This significant wage increase indicates that demand significantly exceeds supply in the labor market. However, despite rising wages as an incentive to increase labor supply, the labor shortage continues and is predicted to worsen in the future. This suggests that this problem, a deep-rooted social issue of structural labor force decline due to a declining birthrate and aging population, cannot be solved simply by economic incentives. This indicates a critical situation affecting the productivity and caregiving capacity of society as a whole.

The shortage of caregivers not only affects the volume of care services provided, but also increases the physical and mental burden on existing caregivers. For example, the physical demands of transferring care recipients from beds to wheelchairs are significant, causing many caregivers to suffer from musculoskeletal disorders such as back pain \cite{backpain1, backpain2}. Furthermore, caregivers must constantly monitor the physical and mental condition of those receiving care through conversation, paying close attention to even the slightest changes. Caregivers are required to communicate appropriately based on the patient's needs and condition, which has a direct impact on the quality of care and patient satisfaction \cite{mental}. However, such work, which requires advanced interpersonal skills and mental concentration, can place a heavy burden on caregivers. Under these circumstances, it could lead to a decline in the quality of care services and ultimately to a decline in the quality of life for society as a whole. Innovative approaches to solving this problem are essential. As a result, there is a growing need for robotics technology to complement and replace human manpower, and attempts are already being made to utilize robots \cite{survey1, survey2}.

Existing robotics offer various forms of support—ranging from physical assistance to psychological interaction. Given that social interaction is essential for building trust and maintaining the quality of care, this study focuses on the role of robots as interaction partners. We aim to explore the possibility of introducing these technologies in healthcare for older adults by evaluating the psychological and physiological impact of robotic interaction. The following section reviews the current landscape of nursing care robots to provide a broad context for how they are currently utilized in care environments.

\subsection{Technological Background}\label{sec1-2}
Nursing care robots are typically designed for specific functional roles, ranging from physical assistance and monitoring to communication and therapeutic interaction. 

The "ROBEAR" (RIKEN) robot uses high-precision sensors and asctuators to handle heavy-lifting tasks, such as patient transfers and standing assistance, minimizing caregiver back strain and injury risks \cite{robear1, robear2}. Conversely, Lio-a reduces daily workloads by several hours, enabling staff to focus on individualized patient care by automating repetitive duties such as item delivery and night patrols and processing data locally for privacy, \cite{lio}. Furthermore, communication therapy robots like the seal-shaped "PARO" provide psychological support through passive tactile interaction. PARO offers benefits similar to animal therapy, reducing loneliness and maintaining cognitive function among the older adults. Studies in dementia care have demonstrated its therapeutic effects on mood and engagement \cite{paro2, paro4}. Long-term interventions have shown significant improvements in psychological outcomes, social interaction, and the achievement of care goals, ultimately enhancing the quality of life for older adults \cite{paro3, paro5}. More broadly, socially assistive robots (SARs) have been recognized for their potential to improve the overall quality of life (QOL) of the older adults by providing both emotional and functional support \cite{qol}.

While these robots demonstrate functional and therapeutic benefits, their integration into care settings increasingly depends on their ability to understand and facilitate Human-Robot Interaction (HRI). Recent studies have explored the effectiveness of SARs in specialized contexts, such as motivating physical exercise among older adults, demonstrating that robotic intervention can successfully encourage health-promoting behaviors \cite{healthcare}. Beyond physical and tactile support, verbal communication has emerged as a vital interface. With advancements in Text-to-Speech (TTS) technology, an increasing number of social robots are being equipped with voice-based interaction capabilities. In studies primarily involving healthy adults, it has been demonstrated that interaction through spoken language is not merely a matter of convenience; it is a crucial element that facilitates conveying information to maintain and repair trust toward the robot \cite{trust}. For instance, experiments conducted in laboratory settings or public spaces have reported that proactive verbal inquiries and feedback from robots can lower psychological barriers for users, thereby eliciting more sustained engagement \cite{engagement}.

However, the evaluation of a robot as an interaction partner is not purely determined by its functional performance. Research on general populations indicates that factors such as the robot's physical appearance \cite{appearance} and specific design elements \cite{appearance2} function as immediate "biases" that shape initial user expectations regarding its role and capabilities. These psychological predispositions are heavily modulated by age; recent studies highlight that older adults often exhibit higher levels of implicit negative attitudes toward robots compared to younger generations, despite maintaining a similar level of curiosity \cite{age}. This tendency is consistent with prior findings that older adults evaluate assistive social agents not only in terms of functional performance, but also through socio-psychological factors such as social presence and trust \cite{elderly}. Observational studies in nursing homes further suggest that the actual interaction between residents and robots is highly contingent on individual and contextual factors, highlighting the need for a deep understanding of how older adults users perceive robotic presence in their daily environment \cite{nursing_home}.

Consequently, in order to implement nursing care robots in society, research into the acceptability of interaction between older adults and robots is necessary. Experimental evaluations of robot services have shown that user satisfaction is deeply tied to the perceived reliability and ease of interaction \cite{trust2}. However, research into HRI, primarily conversational, aimed at older adults is still insufficient. The implementation of socially assistive robots in care for older adults is not yet widespread, partly because of the need for comprehensive user pre-training and the consideration of ethical and practical issues. For widespread acceptance, it is crucial to build a high level of trust between older patients and robots, which requires future designs to prioritize privacy and security \cite{HRI}.

\subsection{Study Purpose}\label{sec1-3}
Based on the background discussed above, the overarching research questions of this study are: 
\begin{itemize}
    \item {Can social robots be effective interaction partners compared to humans?}
    \item {Does a "positive prompt" enhance these interactions similarly for both?}
\end{itemize}
To address these questions, we compare the psychological and physiological responses of older participants during human-human and human-robot encounters. Drawing on research regarding the neurological benefits of positive thinking \cite{positive_prompt}, the experiment specifically investigates the impact of "positive prompts"—encouraging participants to think positively of others—on these interactions.To achieve a comprehensive evaluation of the robot’s potential as a care partner, we integrate findings from three key dimensions: facial emotional responses (RQ1), physiological data (RQ2), and subjective perceptions of acceptance (RQ3). While a related study \cite{mayer} investigates neuroendocrine markers and subjective affect, the present study focuses on the following sub-questions, which are evaluated by comparing interaction partners (Robot vs. Human) and the presence of positive prompts:

\medskip
\noindent\textbf{RQ1: Facial Expressions}:\\
How do these interaction conditions impact emotional response represented by facial emotions?\\

\noindent\textbf{RQ2: Heart rate}:\\
How are these encounters reflected in the physiological data, specifically the mean and standard deviation of heart rate?\\

\noindent\textbf{RQ3: General Perception}:\\
What general perceptions and levels of acceptance toward the robot can be observed in older participants through subjective evaluations?\\

\section{Methods}\label{sec2}
\subsection{Experimental Design}\label{sec2-1}
The study employed a $2 \times 2$ mixed (between- and within-subjects) design. As shown in Table \ref{condition}, the between-subjects factor was the presence of "positive prompts" (No Prompt vs. With Prompt). The within-subjects factor was the interaction partner (Robot vs. Human), with the order of interaction counterbalanced across two groups to mitigate order effects.

In Conditions 3 and 4, the experimenter used verbal prompts to induce a positive mindset. The first prompt (Interaction 1) asked participants to describe positive aspects of their family. The second (Interaction 2) encouraged deeper reflection specifically on the positive qualities of those same family members and shared memories with them. This aimed to serve as a psychological trigger to foster a positive mental state during the interactions.

\renewcommand{\tabularxcolumn}[1]{m{#1}}
\begin{table}[b] 
\caption{Conditions for the interaction experiment\\{\small $2 \times 2$ mixed design. Prompt presence is between-subjects; interaction partner is within-subjects. Orders are counterbalanced to mitigate order effects.}}\label{condition}
\begin{tabularx}{\columnwidth}{|>{\hsize=1.4\hsize}X||>{\hsize=0.8\hsize\centering\arraybackslash}X|>{\hsize=0.8\hsize\centering\arraybackslash}X|}
\hline
\multicolumn{1}{|c||}{ } & No Prompt & With Prompt \\  
\noalign{\hrule height 1.5pt}
Robot first \newline Human second & Condition 1 & Condition 3 \\ \hline
Human first \newline Robot second & Condition 2 & Condition 4 \\ \hline
\end{tabularx}
\end{table}

\begin{figure}[b]
\centering
\includegraphics[width=0.2\textwidth]{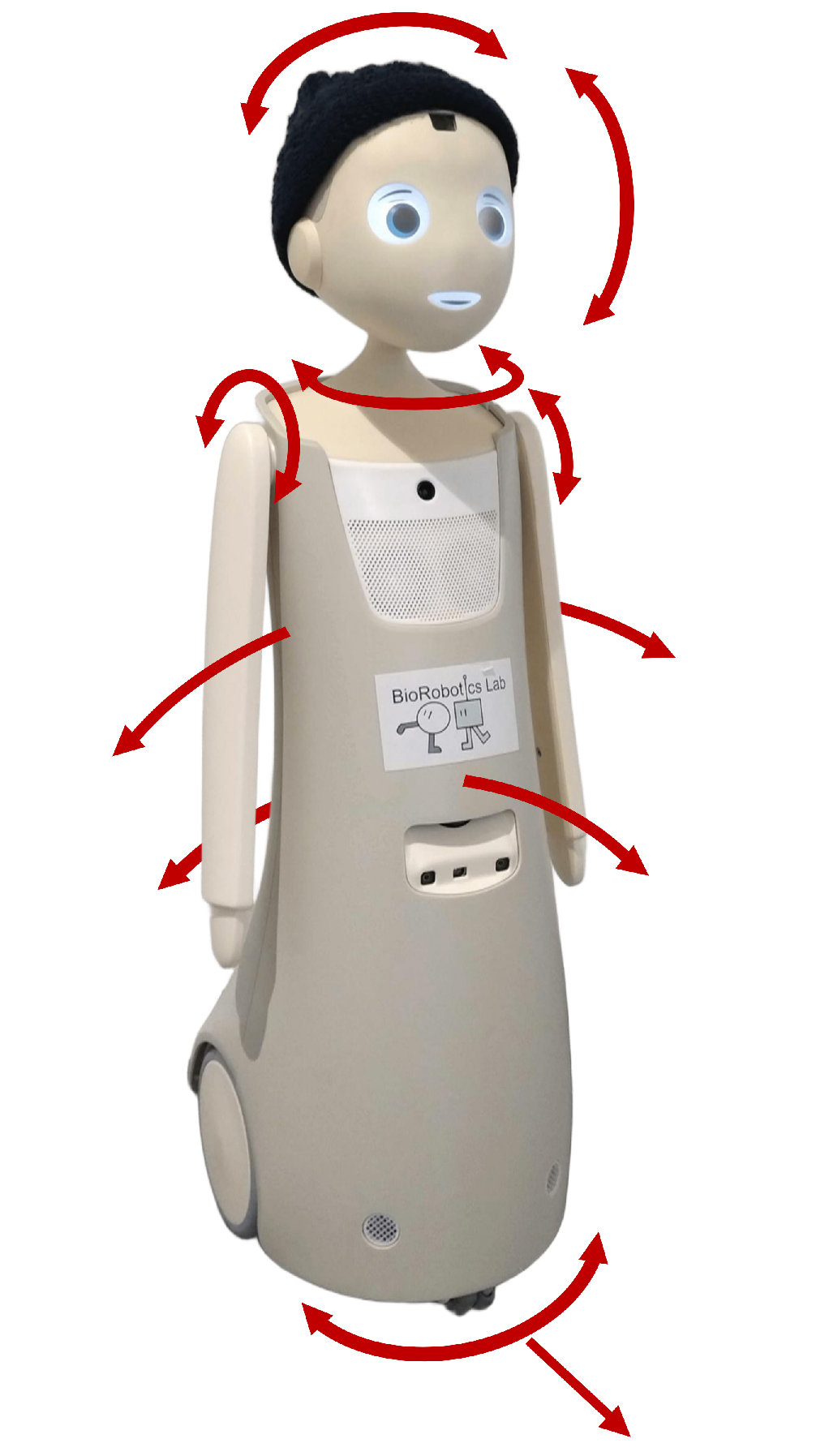}
\caption{Humanoid robot "Navel"\\ \small Its degrees of freedom (DoFs). The arrows indicate the 9 DoFs: 3 for the head, 2 for the shoulders, 2 for the body tilt, and 2 for the differential drive system.}\label{navel}
\end{figure}

\subsection{Used Robot}\label{sec2-2}
The humanoid robot "Navel (Navel Robotics GmbH)," was used in this research. As illustrated by the movement indicators in Fig. \ref{navel}, Navel features a total of nine degrees of freedom (DoFs) to support expressive social interaction. Specifically, the robot is equipped with a 3-DoF head for natural eye contact and nodding, 2-DoF shoulders for arm rotation, 2-DoF tilt functionality for upper-body leaning, and a 2-wheel differential drive system for locomotion. It features expressive 3D eyes using OLED displays, and is powered by an NVIDIA Jetson AGX Xavier SoC operating on a Linux-based system. Standing at a height of 72 cm, the robot is designed to be slightly below the eye level of a seated adult, providing a non-intimidating and friendly presence during interactions.

For this study, Navel was controlled through the Navel Control Studio GUI and customized using the Python Software Development Kit (SDK). The robot's interaction capabilities focused on real-time face tracking to maintain engagement and a TTS system for clear verbal communication through its integrated 4 W broadband speakers. Its abstracted humanoid design avoids the Uncanny Valley while supporting expressive communication \cite{navel}. As a platform providing physical embodiment, Navel’s suitability for evaluating artificial personality and intuitive interaction has been recognized in previous studies, where it was employed as a key comparative model for testing cognitive architectures against non-embodied agents \cite{navel2}. Such characteristics make it an appropriate choice for research involving older populations.

\begin{figure*}[b]
\makebox[\textwidth]{\includegraphics[width=1.0\textwidth]{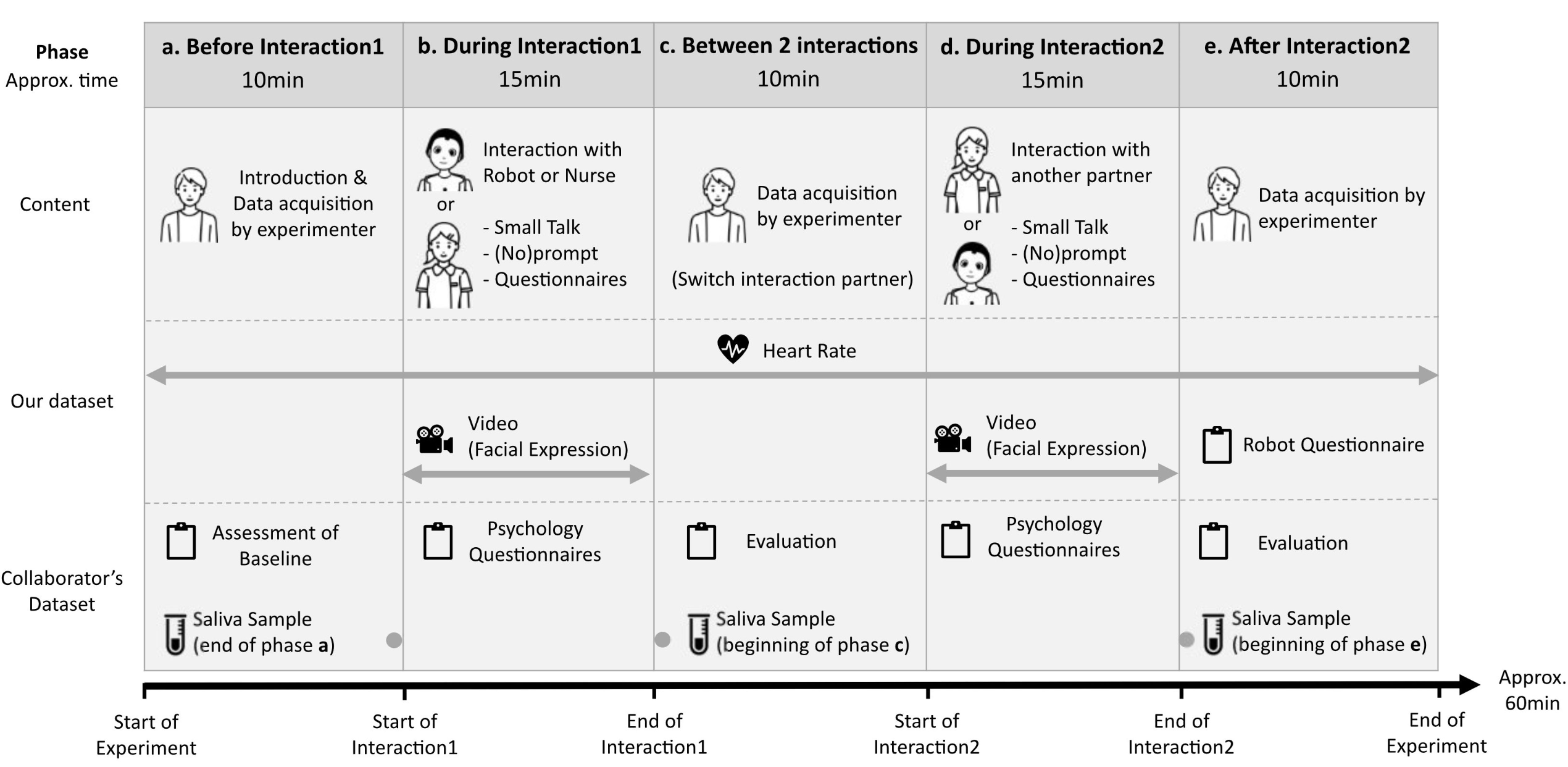}}
\caption{Experimental procedure\\{\small Protocol divided into five phases (a--e). Each segment indicates the specific data collection points for physiological, behavioral, and subjective measures.}}\label{procedure}
\end{figure*}

\begin{figure*}[t]
\centering
\includegraphics[width=1.0\textwidth]{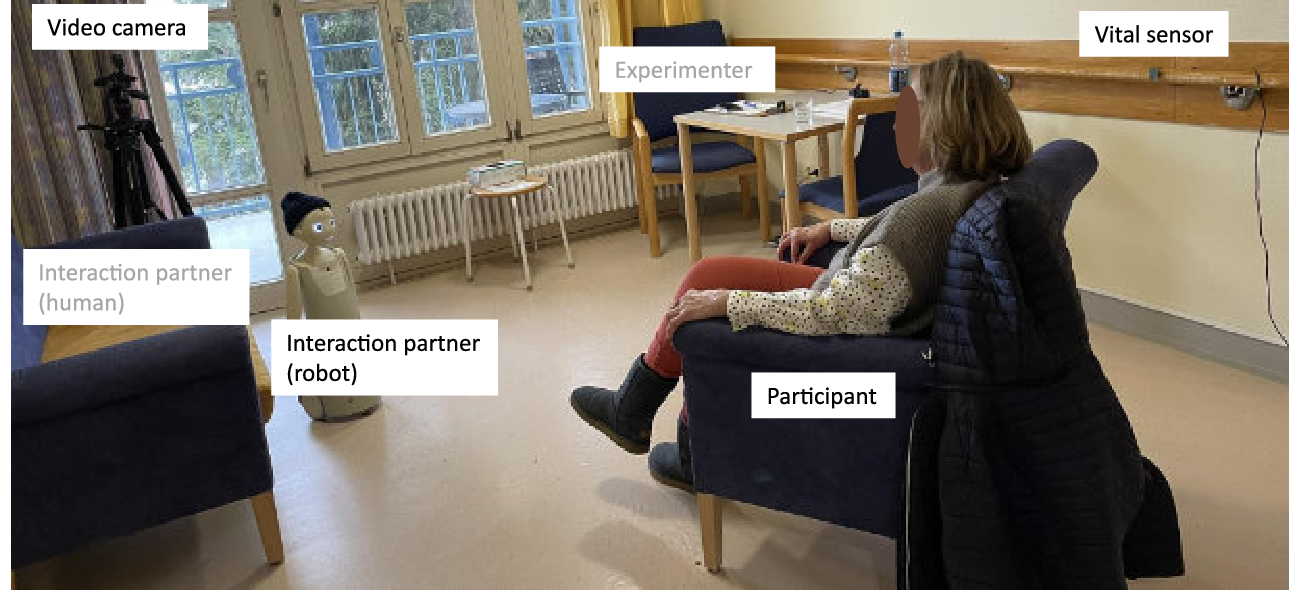}
\caption{Experimental setup\\{\small The participant, robot, and human partner are monitored via a camera and mmWave sensor. The robot is operated by an experimenter in a separate room using a restricted-perception Wizard-of-Oz (WoZ) approach.}}\label{setup}
\end{figure*}

\subsection{Participants}\label{sec2-3}

Eligible participants were German-speaking adults aged 70 or older with normal cognitive functioning, defined as a score of 24 or higher on the Mini-Mental State Examination (MMSE) \cite{MMSE}. Inclusion required the capacity to provide informed consent and complete a one-hour session without hearing impairment. Individuals with diagnosed dementia, severe mental illness, or history of substance abuse were excluded. Participants were also required to follow standardized pre-test restrictions regarding alcohol, caffeine, and nicotine. 

Participants were recruited through public advertisements and existing participant databases. A total of 35 older participants (8 men and 27 women) with an average age of 77.9 years took part in the experiment. Each participant interacted with both a human and the robot, and the order was randomly assigned for each participant. 

\subsection{Experimental Procedure}\label{sec2-4}

As illustrated in Fig. \ref{procedure}, the interaction protocol was structured into five distinct phases (a–e) to facilitate a granular analysis of physiological and emotional shifts. The figure provides a comprehensive overview of the data collection timeline, distinguishing between the primary datasets (e.g., heart rate and facial expressions) and the collaborator's supplementary psychological and physiological measures. These phases allowed for the targeted evaluation of facial expressions (RQ1) and heart rate (RQ2) at different stages of the encounter:\\

\noindent a. Before Interaction 1:\\
Participants entered the room, led by the experimenter, to perform initial saliva collection and complete a pre-questionnaire.\\

\noindent b. During Interaction 1:\\
The first encounter with either a human or the robot occurred, including the first positive prompt for participants in Conditions 3 and 4.\\

\noindent c. Between 2 Interactions:\\
A short break was provided for the middle saliva collection and to allow physiological signals to stabilize.\\

\noindent d. During Interaction 2:\\
The second encounter took place with the remaining partner, including the second positive prompt for Conditions 3 and 4.\\

\noindent e. After Interaction 2:\\
The final saliva collection was conducted, followed by a subjective evaluation questionnaire regarding the robot (RQ3).\\

The experimental environment is shown in the Fig. \ref{setup} . Participants entered the room, led by the experimenter. After participants sat down, a saliva sample was taken and they answered a mini-questionnaire. Then, the robot or a human interaction partner entered the room and began the interaction. When the robot entered or left the room, it moved automatically according to a pre-prepared program. During the experiment, it also tracked the participants' faces and made eye contact. Throughout the hour-long session, participants' facial expressions and heart rate were continuously monitored using the video camera and mmWave sensor setup.

To ensure uniform conditions, the robot operated on a predefined script, with responses selected via a Wizard-of-Oz (WoZ) method by an experimenter in a separate room. This approach allowed the robot to simulate high-level social intelligence while maintaining strict experimental control over the interaction flow. Limiting the robot’s responses through predefined scripts corresponds to the restricted-perception WoZ approach, which is commonly used to study social interaction strategies under controlled conditions \cite{oz}.\\


\subsection{Data Collection and Processing}\label{sec2-5}
To address the each research question and evaluate participants' reactions, the following data were collected:\\

\noindent\textbf{RQ1: Facial Expressions}

The experiment was captured using a video camera set in the room, and emotions were estimated from facial images using an open-source facial expression recognition model. In HRI, facial expressions serve as a vital indicator of a user's internal emotional state and can be used to assess robot acceptance and the quality of the interaction \cite{emotion1}. Since human emotional responses are often manifested through non-verbal cues before being consciously reported, analyzing these expressions allows for a more objective evaluation of the user's immediate psychological reaction to the robot \cite{emotion2}.

OpenCV's facial expression recognition system operates in an efficient two-stage pipeline. First, a lightweight YuNet model optimized for edge devices detects faces. YuNet is an excellent face detection model that combines high accuracy with millisecond-level inference speed \cite{facenet}. Second, a pre-trained model in ONNX format classifies seven emotions (anger, disgust, fear, happiness, neutral, sadness, and surprise) from the detected facial regions. Overall, it is a modular and practical computer vision solution. The detected facial expression data were processed into graphs.\\

\noindent\textbf{RQ2: Heart Rate}

Mean heart rate and heart rate variability indices, such as the standard deviation of inter-beat intervals, are widely established as objective physiological markers for detecting autonomic responses to mental stress \cite{hrv1, hrv2}. To measure heart rate, we used a MR60BHA1 60GHz mmWave sensor(Seeed Studio). It employs frequency-modulated continuous wave (FMCW) radar technology to detect subtle chest movements associated with heartbeat and breathing, making it particularly valuable in that it is less susceptible to environmental factors such as changes in light. Recent studies have validated that 60GHz FMCW radar can achieve high-precision heart rate monitoring comparable to traditional electrocardiogram (ECG) sensors, particularly in its ability to filter out environmental noise \cite{heartrate2}. This sensor is connected to a Raspberry Pi 5 operating on Ubuntu 24.04, allowing for the real-time monitoring and recording of heart rate data via serial communication. The use of this non-contact sensor ensures that participants can interact naturally with the robot or human caregiver without the discomfort of wearable devices. This non-invasive approach is crucial for maintaining ecological validity in social interaction studies, as wearable sensors can increase the perceived artificiality of the experiment and influence the participant's spontaneous emotional expression \cite{heartrate3}.\\

\vspace{5mm}
\noindent\textbf{RQ3: General Perception}

Participants' perceptions of the robot were assessed by a questionnaire. It used Likert-style questions, where respondents chose answers on a scale of 1-5, and a descriptive format. The Likert-style format is superior in terms of reliability and ease of response \cite{likert}. Unless otherwise noted, items were rated on a 5-point scale (1. Strongly Disagree - 5. Strongly Agree). The questions are as follows:\\

\noindent(a) This robot is friendly\\
(b) The size of this robot is (1. Too Small - 5. Too Big) \\
(c) I want to touch this robot. \\
(d) I want this robot to give me an attention. \\
(e) The way this robot speaks is natural. \\
(f) This robot is physically intrusive. \\
(g) This robot is psychologically calming.\\ 
(h) The sound of this robot moving is annoying. \\
(i) This robot seems safe. \\
(n) Free comment\\

\noindent\textbf{Overall Evaluation}

By integrating these multifaceted observations with RQ1: Facial Expressions, RQ2: Heart Rate, and RQ3: General Perception, we aims to determine the overall effectiveness of a robot in supporting the well-being of older populations compared to the interaction with humans. This integrative approach is grounded in the principle of triangulation, where combining behavioral (RQ1), physiological (RQ2), and subjective (RQ3) measures provides a more robust and objective assessment than any single modality alone \cite{multi}\cite{multi2}. To further situating these measures within a broader psychological context, this cross-disciplinary collaboration also gathered a comprehensive suite of data—ranging from baseline cognitive assessments to neuroendocrine markers—the detailed psychological analyses of which are presented in \cite{mayer}. These supplementary datasets include:

\begin{itemize}
    \item Assessment of Baseline: Includes cognitive ability by MMSE \cite{MMSE}, demographic and control variables (health, living conditions, technology use), Computer Anxiety Trait Scale (CATS) \cite{CATS}, and baseline affect (PANAS) \cite{PANAS}.
    \item Psychological Questionnaires: Measures of mental health and social context using the Geriatric Depression Scale (GDS) \cite{GDS} and the 6-item Lubben Social Network Scale (LSNS-6) \cite{LSNS}.
    \item Evaluation of Interaction: Assessment of the interaction through subjective ratings (Trustworthiness and Comfort) and physiological markers (salivary Cortisol, Alpha-Amylase, and Oxytocin).
\end{itemize}

\subsection {Statistical Analysis}\label{sec2-6}
Statistical analyses were performed to evaluate the participants' emotional, physiological, and subjective responses using Python with the SciPy and Pingouin libraries.

For the analysis of facial expressions (RQ1), participants were categorized into four groups (A--D) based on specific criteria, such as the initial reaction and the duration of "Happy" expressions. The distributions of these categories were compared between interaction partners (Robot vs. Human) and prompt conditions (No Prompt vs. With Prompt) using Chi-square tests ($df=1$). In cases where the expected cell frequencies were small, Fisher's exact tests were employed to ensure the accuracy of the $p$-values.

For the heart rate data (RQ2), the mean and standard deviation of heart rates were compared across five experimental phases (a--e). A one-way Analysis of Variance (ANOVA) was conducted to identify significant differences between experimental conditions and interaction partners at each phase.

Subjective evaluations from the robot perception survey (RQ3) were similarly analyzed using ANOVA to compare the scores (1--5 Likert scale) across the four experimental conditions. 

For all statistical tests, a $p$-value of less than .05 was considered statistically significant, while $p$-values between .05 and .20 were discussed as marginal trends where appropriate.

\section{Result and Discussion}

\subsection{Result}\label{sec3-1}
The 35 participants were randomly assigned to one of four experimental groups based on the interaction order and the presence of prompts: robot-first ($n=9$ without prompt; $n=7$ with prompt) and human-first ($n=10$ without prompt; $n=9$ with prompt). Regarding the video data, recording failures occurred in two sessions for the robot and two sessions for a human.  The parts that were not recorded were due to participants not facing the camera and a failure of the video camera recording. Therefore, the results of these experiments have been appropriately excluded from the conclusions that need to be compared below. The excluded data are indicated in the tables.


\subsubsection{RQ1: Facial Expressions}

\begin{figure}[htbp]
  \centering

  \begin{subfigure}{1.0\linewidth}
    \centering
    \includegraphics[width=\linewidth]{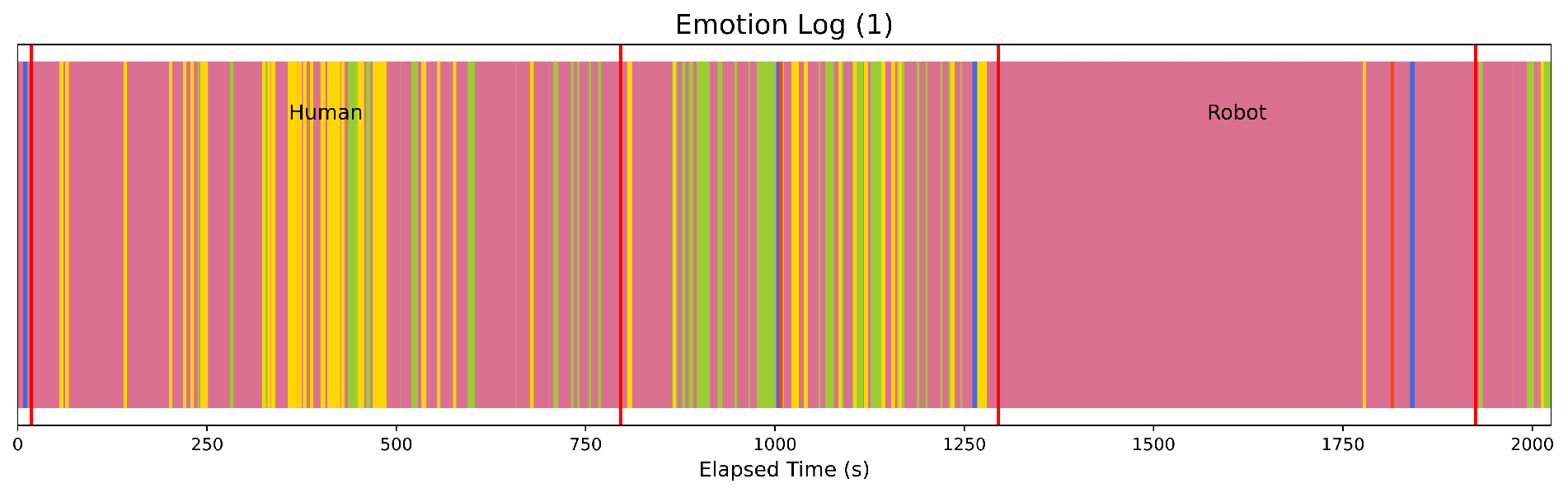}
    \caption{Emotion log}
    \label{result_1_log}
  \end{subfigure}

  \vspace{1em}

  \begin{subfigure}{0.8\linewidth}
    \centering
    \includegraphics[width=\linewidth]{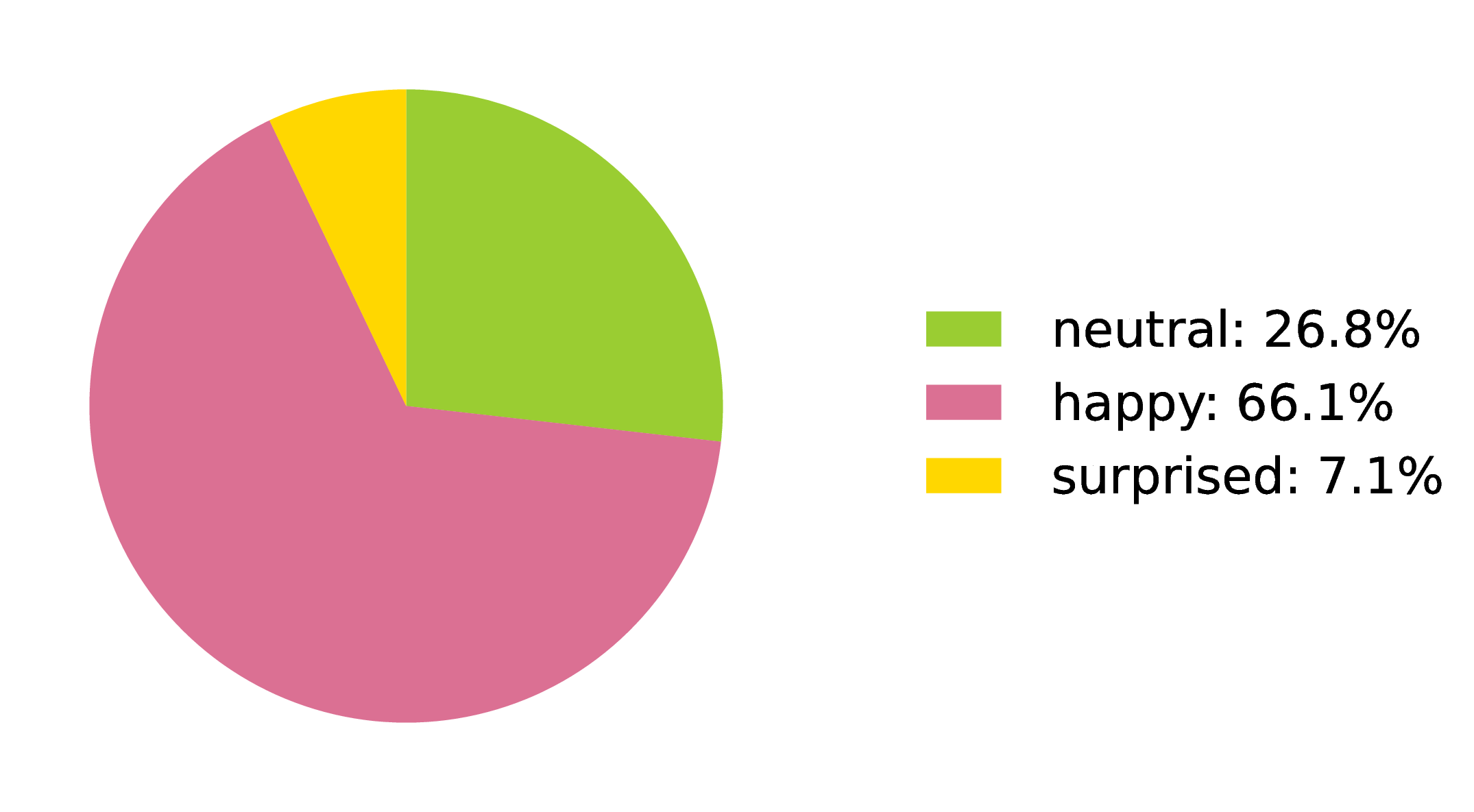}
    \caption{Emotion pie chart interacted with a human}
    \label{result_1_nurse}
  \end{subfigure}

  \vspace{1em}

  \begin{subfigure}{0.8\linewidth}
    \centering
    \includegraphics[width=\linewidth]{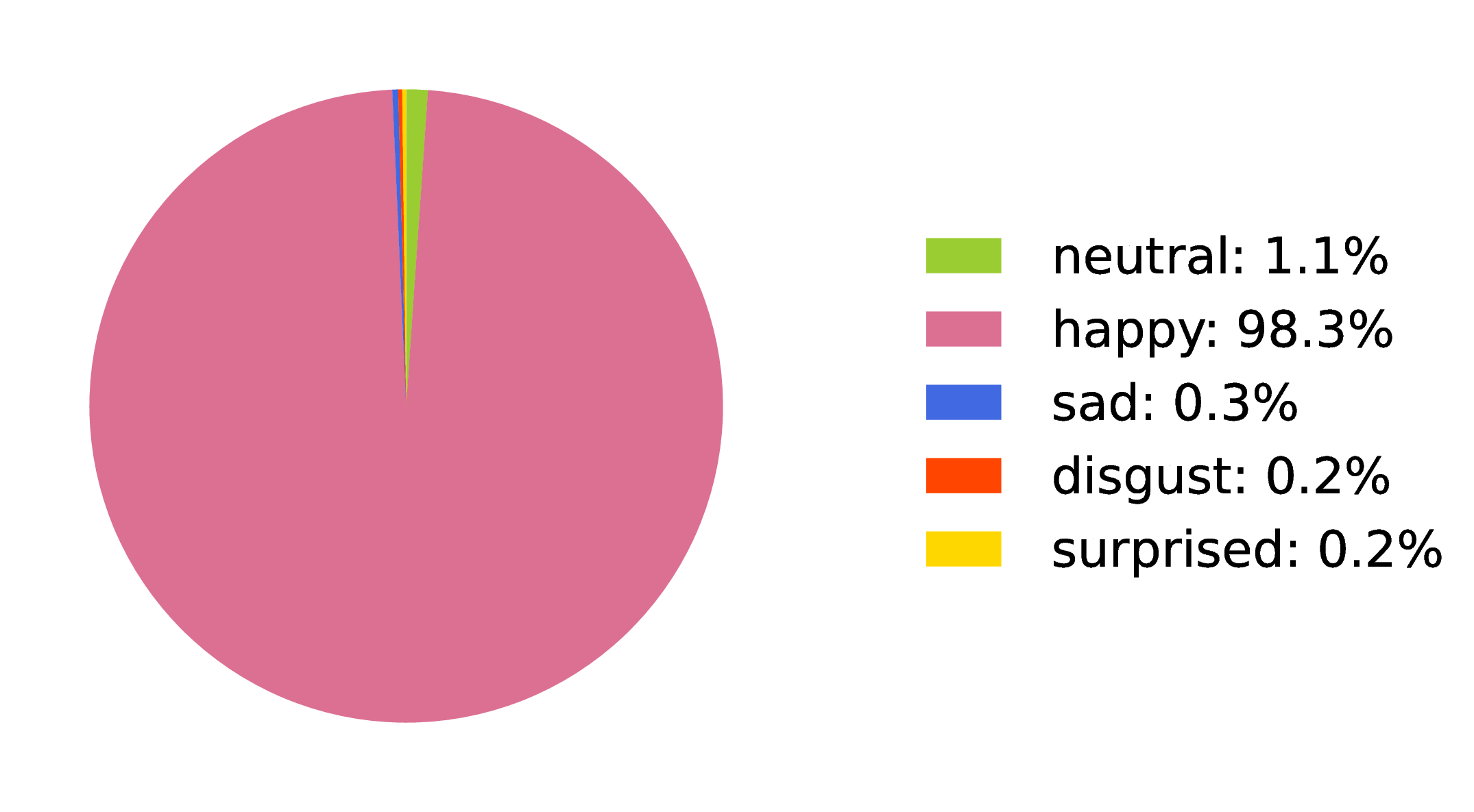}
    \caption{Emotion pie chart interacted with the robot}
    \label{result_2_navel}
  \end{subfigure}

  \vspace{1em}

  \begin{subfigure}{1.0\linewidth}
    \centering
    \includegraphics[width=1.0\linewidth]{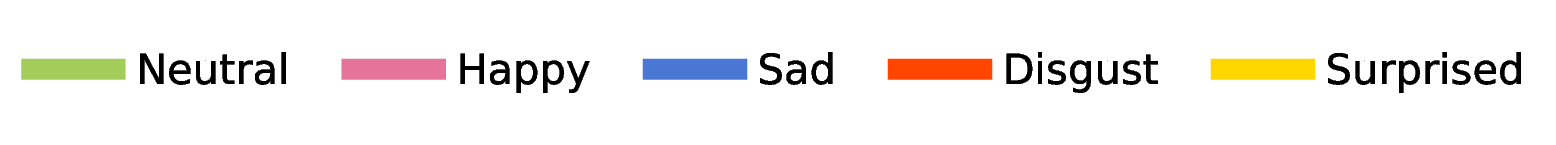}
  \end{subfigure}
  
  \caption{Emotion results of participant 1/ condition 4 \\{\small They exemplify the emotional analysis for a single participant, showing both the temporal transitions across all phases and the ratio of emotions within each interaction.}}
  \label{emotion01}
\end{figure}


\begin{figure}[htbp]
  \centering

  \begin{subfigure}{0.8\linewidth}
    \centering
    \includegraphics[width=\linewidth]{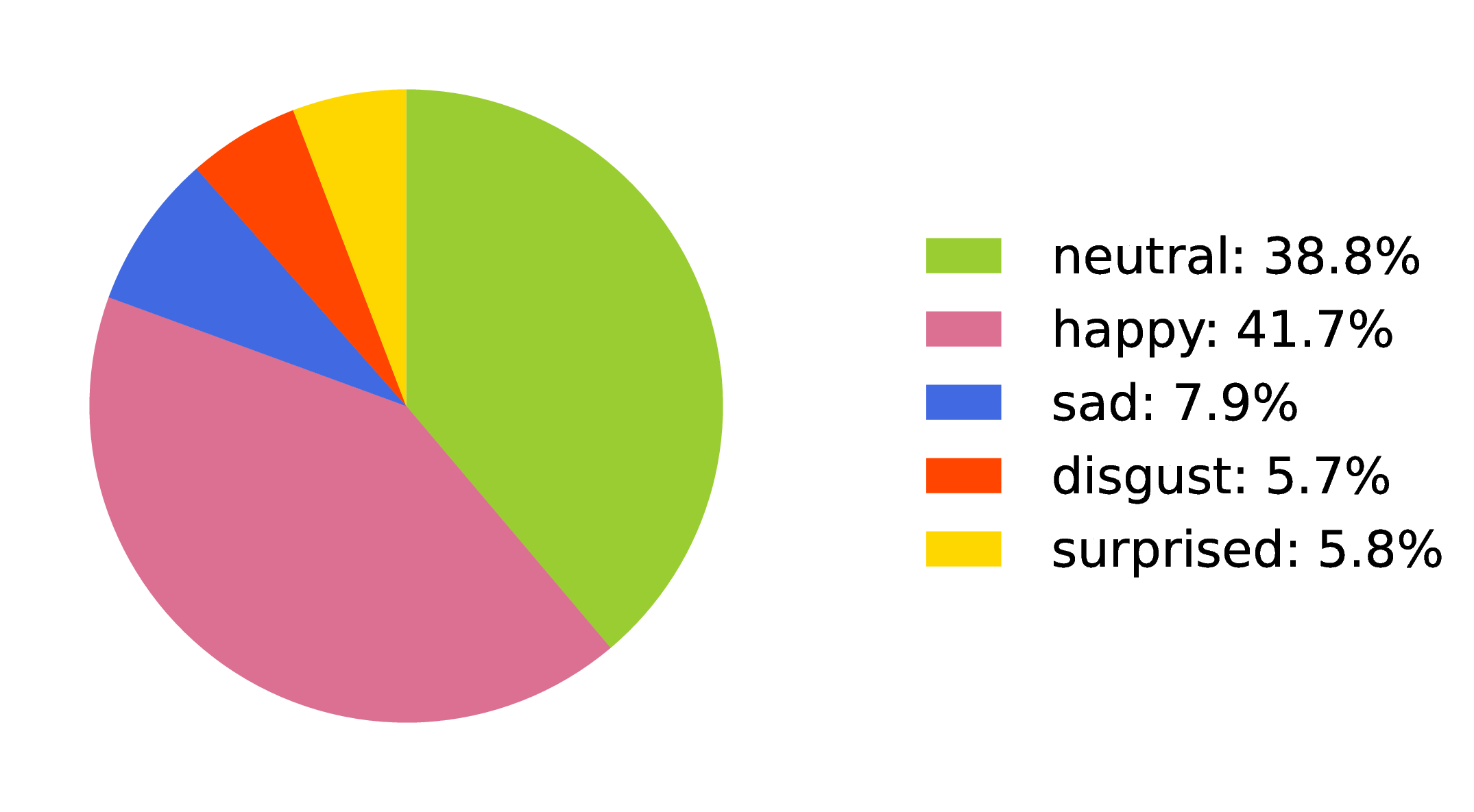}
    \caption{Emotion pie chart interacted with a human}
    \label{result_all_nurse}
  \end{subfigure}

  \vspace{1em}

  \begin{subfigure}{0.8\linewidth}
    \centering
    \includegraphics[width=\linewidth]{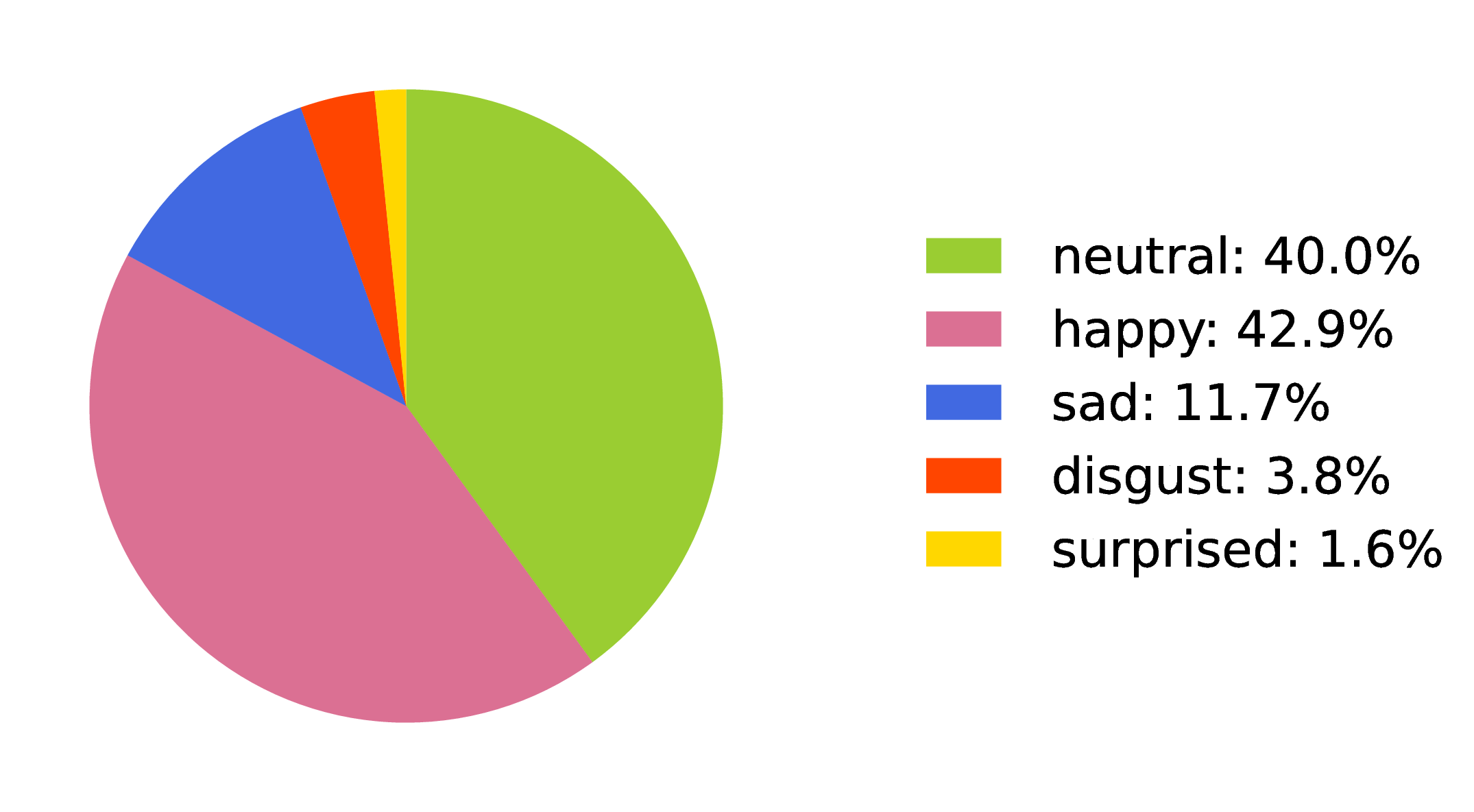}
    \caption{Emotion pie chart interacted with the robot}
    \label{result_all_navel}
  \end{subfigure}
  
  \caption{Emotion results of all participant \\{\small Emotional responses toward the robot and human partners are comparable.}}
  \label{emotion_all}
\end{figure}

The recorded data were subjected to image recognition to estimate emotions that can be read from the face. The Figs. \ref{result_1_log}, \ref{result_1_nurse} and \ref{result_2_navel} show the emotion data during the experiment for the first participant in graph form, with light green, pink, blue, red and yellow representing "Neutral", "Happy", "Sad", "Disgust", and "Surprised". In the Fig. \ref{result_1_log}, the horizontal axis shows the passage of time, and the time when the interaction with each interaction partner took place is shown. The Fig. \ref{result_1_nurse} and Fig. \ref{result_2_navel} show the percentage of emotions detected during the interaction with each interaction partner in a pie chart. For example, the Figs. \ref{result_1_log} shows that "Happy" was detected frequently during the interaction, and "Neutral" was detected frequently between Interactions 1 and 2. "Surprised" was also occasionally detected in Interaction 1 with a human. Furthermore, Figures \ref{result_1_nurse} and \ref{result_2_navel} allow for a more detailed comparison of the proportions of emotions; in Interaction 1 with a human, in addition to "Happy", there was also about 30\% of "Neutral" and "Surprised" in total, but in Interaction 2 with the robot "Happy" was detected 98\% of the time.

Results for all 35 participants are shown in the Appendix \ref{A_all_log}.\\

\vspace{1cm}
\noindent\textbf{Comparison of Interaction Partners for RQ1}

\begin{table}[t]
\centering
\caption{Statistical summary of facial expression comparisons (Chi-square test, $df=1$)}\label{stat_summary}
\begin{tabularx}{\columnwidth}{|>{\hsize=1.6\hsize}X|>{\hsize=0.6\hsize\centering\arraybackslash}X|>{\hsize=0.6\hsize\centering\arraybackslash}X|>{\hsize=0.6\hsize\centering\arraybackslash}X|>{\hsize=0.6\hsize\centering\arraybackslash}X|}
\hline
Category & Robot ($n$) & Human ($n$) & $\chi^2$ & $p$-value \\ \hline
(A)\newline Initial Happy & 21 & 22 & 0.060 & .806 \\ \hline
(B)\newline Happy $>50\%$ & 14 & 13 & 0.060 & .806 \\ \hline
(C)Higher Happy freq. & 17 & 14 & 0.521 & .470 \\ \hline
(D)\newline Negative $<25\%$ & 25 & 23 & 0.265 & .607 \\ \hline
\end{tabularx}
\vspace{1mm}
{\footnotesize N$n$ represents the total number of participants meeting the criterion.}
\end{table}

The emotional responses toward the robot and human partners were evaluated based on four categories (A--D). As illustrated in Figure \ref{emotion_all}, the overall ratio of emotional expressions remained remarkably similar between the robot and a human interaction partner. Table \ref{stat_summary} provides a statistical summary of these comparisons. Overall, the analysis revealed that the robot elicited positive facial expressions at a level comparable to a human interaction partner.  The comprehensive raw data and specific breakdowns for each participant group across all categories (A--D) and conditions (1--4) are provided in Appendix \ref{A_breakdown}. 

Across the 35 participants, the distribution of positive("Happy") and negative("Sad" and "Disgust") emotions remained consistent regardless of the partner type:
\begin{itemize}
    \item \textbf{Initial Reaction (A):} "Happy" was detected at the start of the interaction for 21 participants with the robot and 22 with a human partner. A Chi-square test indicated no significant difference ($\chi^2 = 0.06, p = .806$).
    \item \textbf{Duration of Positive Affect (B):} 14 participants smiled for more than 50\% of the time with the robot, compared to 13 with a human ($\chi^2 = 0.06, p = .806$).
    \item \textbf{Comparative Frequency (C):} A higher percentage of "Happy" was observed in 17 cases for the robot and 14 for a human ($\chi^2 = 0.52, p = .470$).
    \item \textbf{Negative Affect Suppression (D):} Negative emotions remained below 25\% for 25 participants with the robot and 23 with a human ($\chi^2 = 0.27, p = .607$).
\end{itemize}

To further investigate whether specific experimental conditions (Conditions 1--4) influenced these results, Fisher's exact tests were performed on sub-groups where numerical differences appeared larger (e.g., Category C in Condition 3, where 2/7 for robot vs. 4/7 for human). The results confirmed that all such variations were statistically non-significant ($p > .59$ in all cases). This demonstrates that any observed numerical disparities between partners within specific conditions are standard occurrences within the expected range of statistical probability.\\


\noindent\textbf{Comparison of Prompt Conditions for RQ1}
The influence of the positive prompt was analyzed by comparing the no-prompt group ($n=19$) and the prompt group ($n=16$). Table \ref{prompt_stat_summary} summarizes the statistical results for this comparison across categories B and D. 

\begin{table}[t]
\centering
\caption{Statistical comparison of facial expressions with and without positive prompts (Chi-square test, $df=1$)}\label{prompt_stat_summary}
\begin{tabularx}{\columnwidth}{|>{\hsize=1.6\hsize}X|>{\hsize=0.6\hsize\centering\arraybackslash}X|>{\hsize=0.6\hsize\centering\arraybackslash}X|>{\hsize=0.6\hsize\centering\arraybackslash}X|>{\hsize=0.6\hsize\centering\arraybackslash}X|}
\hline
Category & No Prompt ($n$) & With Prompt ($n$) & $\chi^2$ & $p$-value \\ \hline
(B)Happy $>50\%$ & 9 & 6 & 0.345 & .557 \\ \hline
(D)Negative $<25\%$ & 16 & 13 & 0.054 & .817 \\ \hline
\end{tabularx}
\vspace{1mm}
\end{table}

\begin{table}[t]
   \centering
   \caption{Comparison of results with and without prompts}\label{re_5}
   \begin{tabularx}{\columnwidth}{|X|c|c|}
   \hline
   Category & No prompt & With prompt \\ 
   \noalign{\hrule height 1.5pt}
   (B)\newline Happy $> 50\%$  & 52.9\% & 50.0\% \\ \hline
   (D)\newline Negative $< 25\%$ & 94.1\% & 86.7\% \\ \hline
   \end{tabularx}
\end{table}

As shown in Table \ref{prompt_stat_summary}, statistical analysis revealed no significant differences between the no-prompt and prompt conditions. For Category B (sustained "Happy" expressions), 9 participants in the no-prompt group and 6 in the prompt group met the criteria. Similarly, for Category D (suppression of negative emotions), no significant difference was observed, with 16 and 13 participants meeting the criteria, respectively. To further validate these results given the sample size, Fisher's exact tests were also conducted, which confirmed that the observed numerical disparities were standard occurrences within the expected range of statistical probability ($p = .734$ for Category B, $p >.999$ for Category D). These results indicate that the positive prompt intervention did not significantly influence the participants' facial expressions, suggesting that the emotional responses were primarily driven by the interaction with the partners themselves.

Although statistical analysis showed no significant difference ($p > .55$), a slight numerical trend was observed where the no-prompt conditions yielded marginally higher percentages of positive responses as shown in the Table \ref{re_5}.

\subsubsection{RQ2: Heart Rate}

\begin{figure*}[h]
    \centering

    \begin{subfigure}{0.48\textwidth}
        \centering
        \includegraphics[width=\linewidth, height=5.5cm]{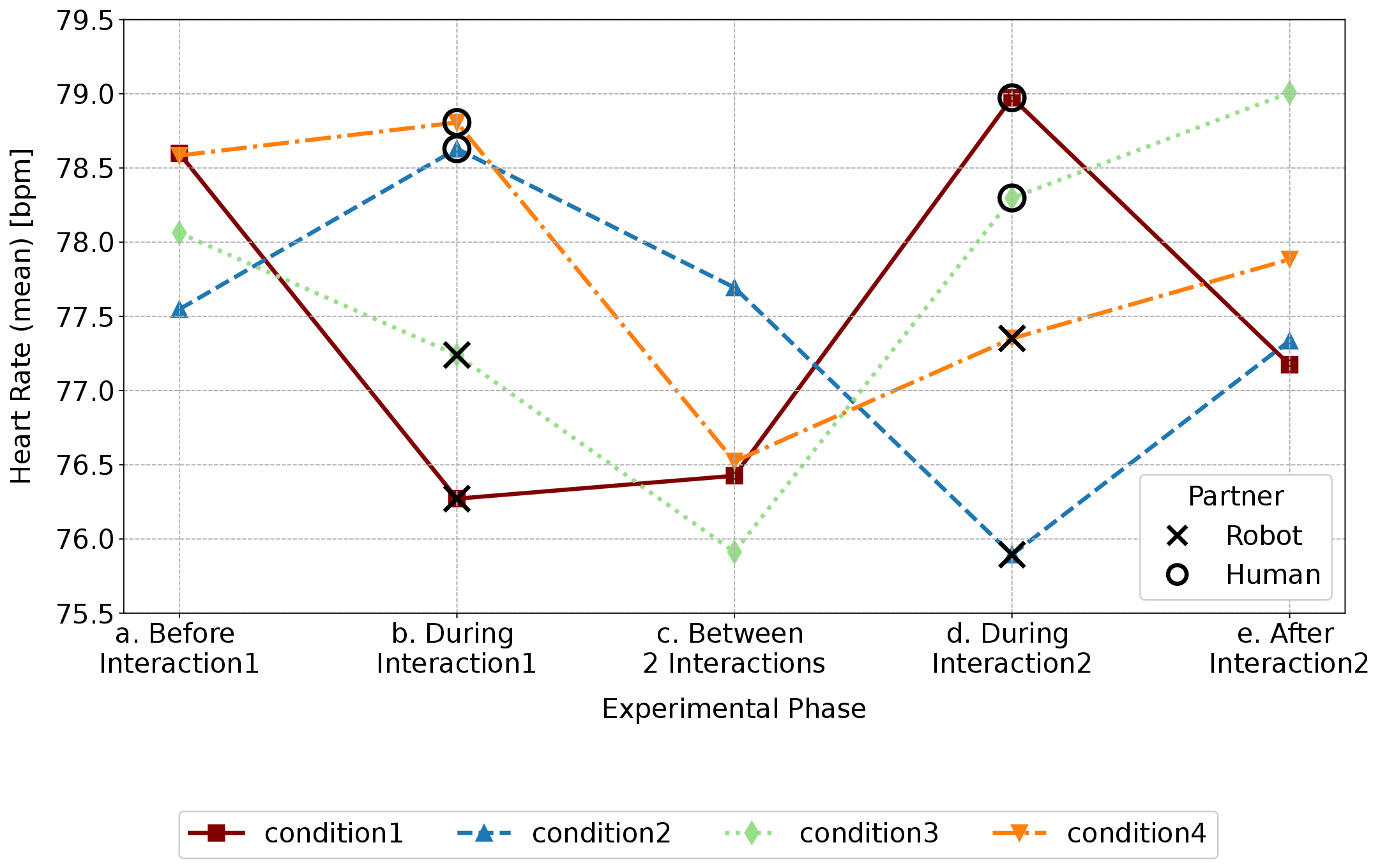}
        \caption{Mean heart rate (by condition)}
        \label{co_me}
    \end{subfigure}
    \hfill
    \begin{subfigure}{0.48\textwidth}
        \centering
        \includegraphics[width=\linewidth, height=5.5cm]{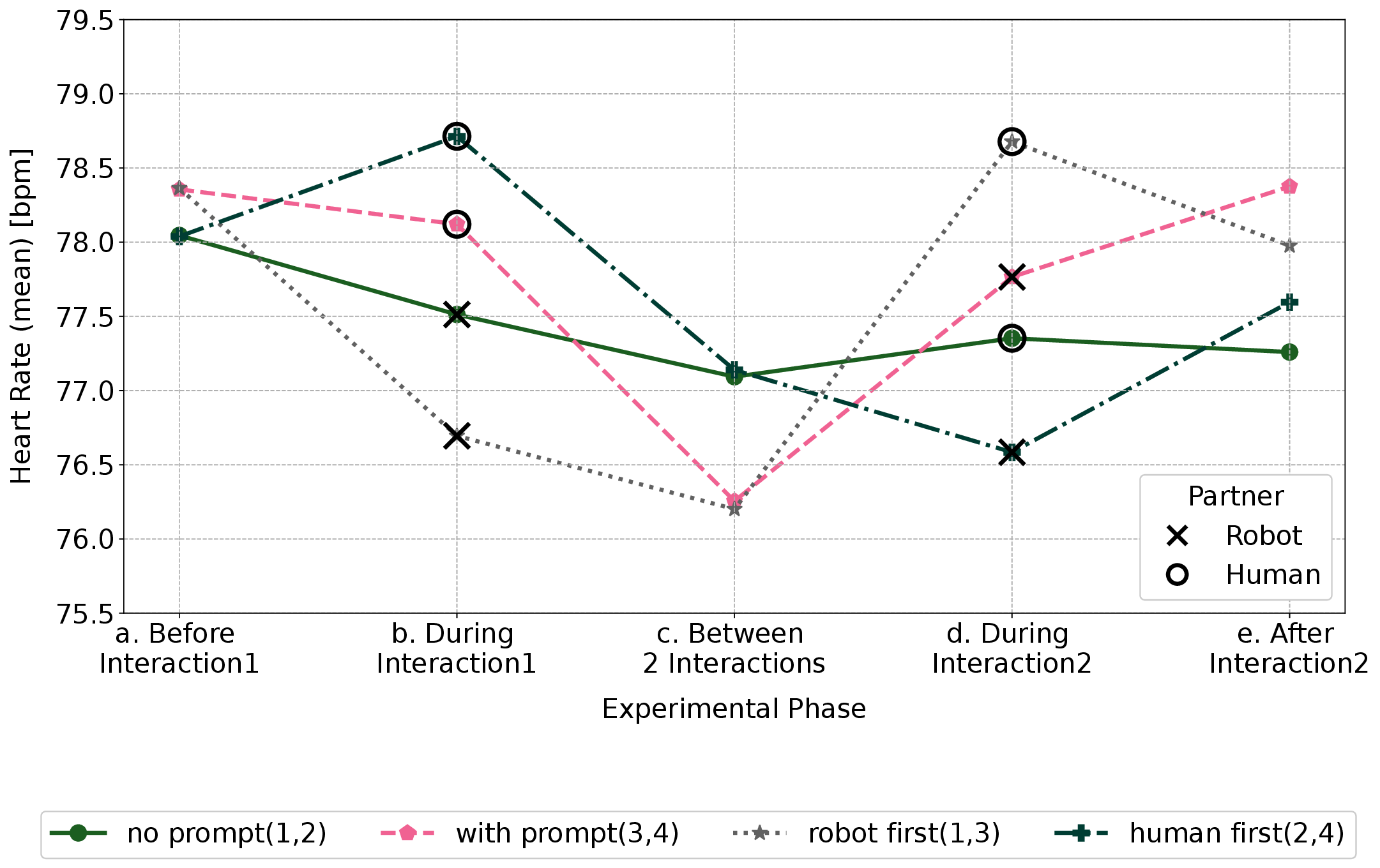}
        \caption{Mean heart rate (by group)}
        \label{gr_me}
    \end{subfigure}

    \vspace{15pt}
    
    \begin{subfigure}{0.48\textwidth}
        \centering
        \includegraphics[width=\linewidth, height=5.5cm]{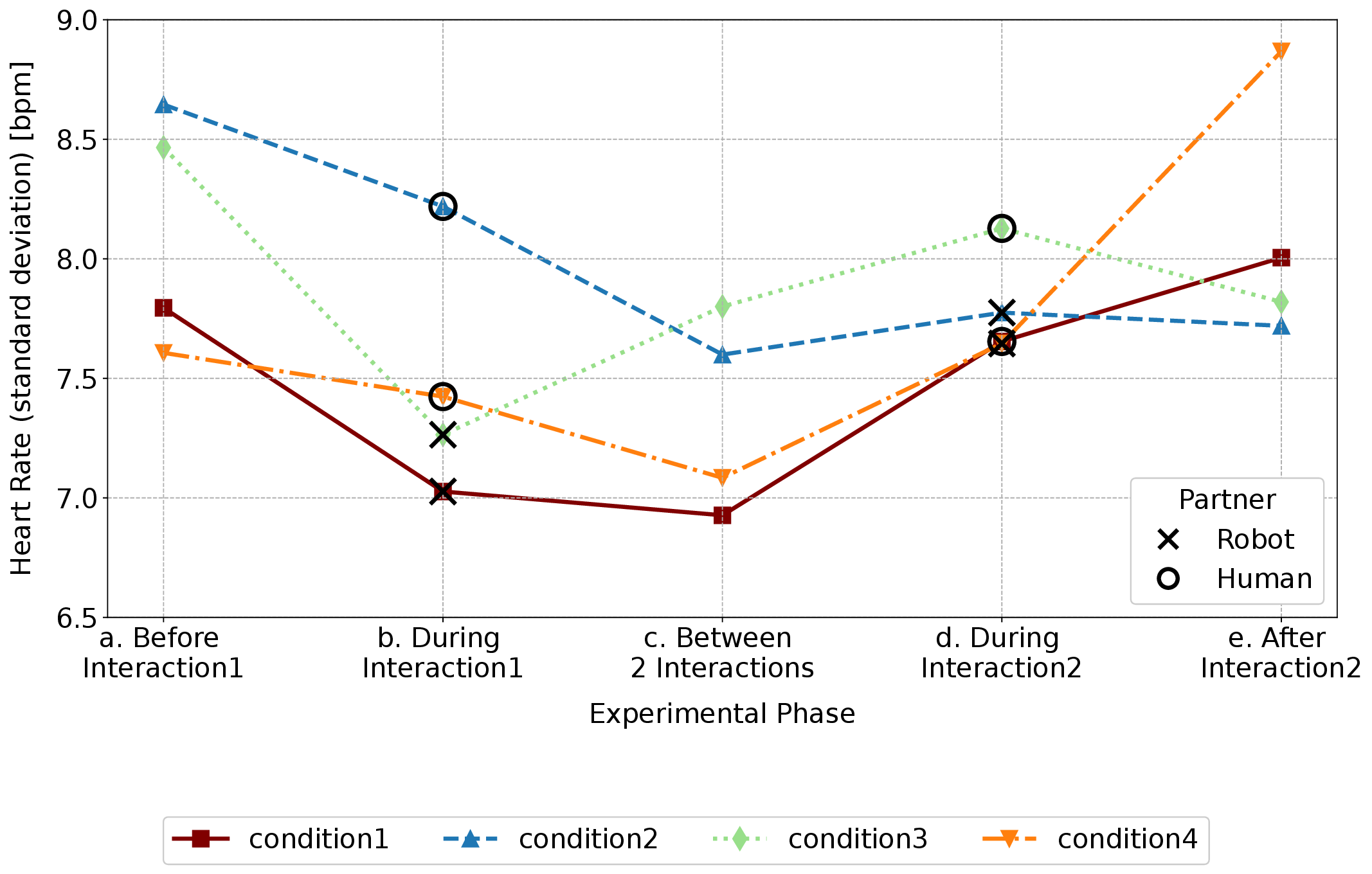}
        \caption{Heart rate SD (by condition)}
        \label{co_sd}
    \end{subfigure}
    \hfill
    \begin{subfigure}{0.48\textwidth}
        \centering
        \includegraphics[width=\linewidth, height=5.5cm]{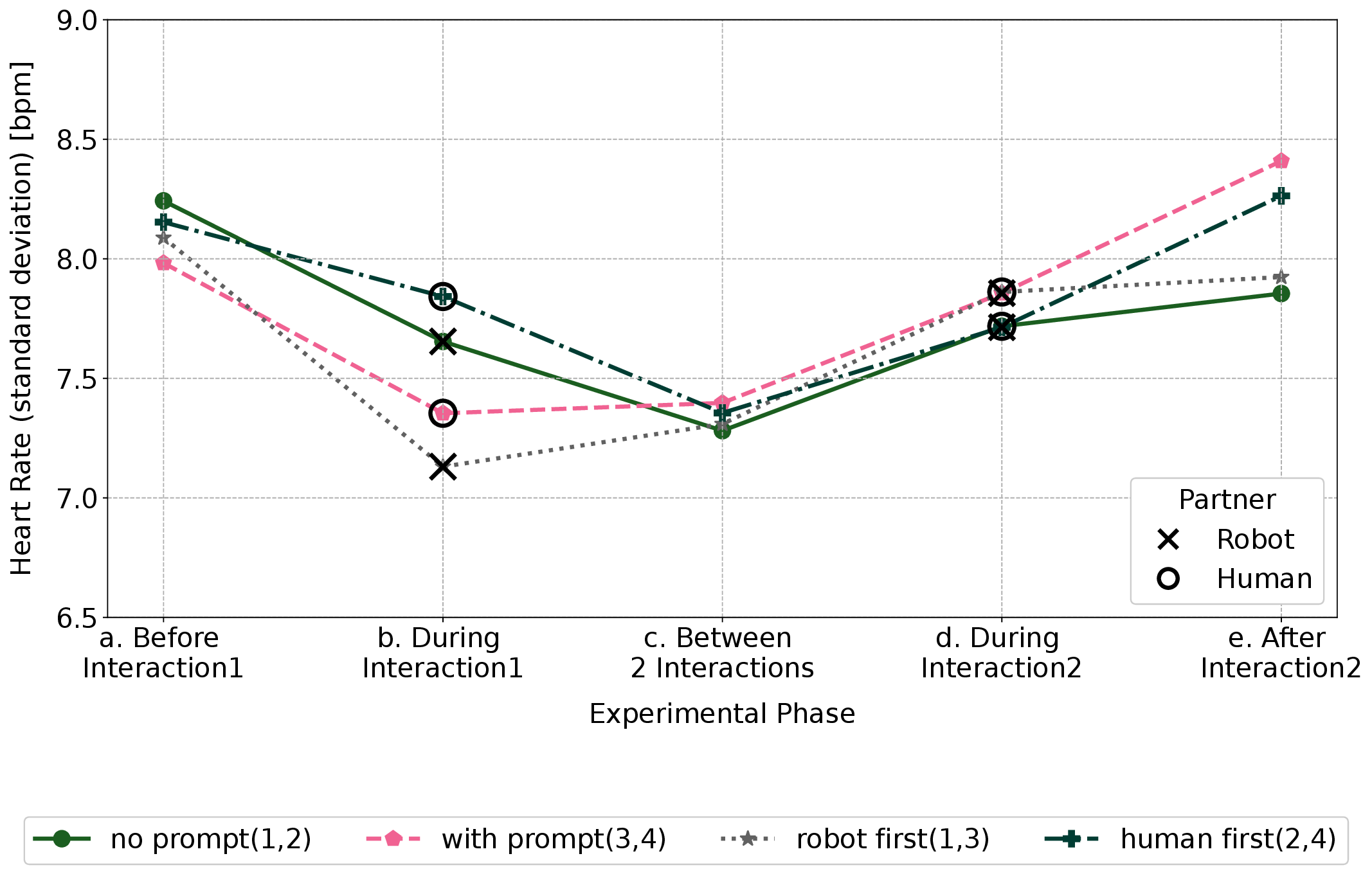}
        \caption{Heart rate SD (by group)}
        \label{gr_sd}
    \end{subfigure}

    \caption{Experimental results of heart rate: Mean and Standard Deviation (SD) for each condition and group\\{\small The robot interaction phases (marked with crosses) generally show lower and more stable heart rate values compared to human interaction phases.}}
    \label{fig:HR_combined}
\end{figure*}

Heart rate data during the experiment was collected and results are divided into sections a. Before interaction 1, b. During Interaction 1, c. Between 2 Interactions, d. During Interaction 2, and e. After Interaction 2, and the heart rate of each participant was calculated in each section. For the 4 groups reflecting the 4 conditions, mean and standard deviation of hearat rate during 5 sections were compared.

Overall, the heart rate remained relatively stable across all conditions, with no significant main effects observed for the experimental conditions ($F = 0.071, p = .975$) or the specific phases ($F = 0.959, p = .433$).

Figs. \ref{co_me} and \ref{co_sd} show the progress of mean heart rate. The mean heart rate for each individual in each section was calculated, and then the mean for all participants in the same group was calculated. Similarly, the Figs. \ref{gr_me} and \ref{gr_sd} show the standard deviation of heart rate. The standard deviation of heart rate for each individual in each section was calculated, and then the mean for all participants in the same group was calculated. Figs. \ref{co_me} and \ref{co_sd} show the results by condition, and the Figs. \ref{gr_me} and \ref{gr_sd} show the results by group (presence or absence of the positive prompt, order of interaction partners). The black crossess in the figure indicate interactions with the robot.\\

\begin{figure*}[b]
    \centering

    \begin{subfigure}{0.45\linewidth}
        \centering
        \includegraphics[width=\linewidth]{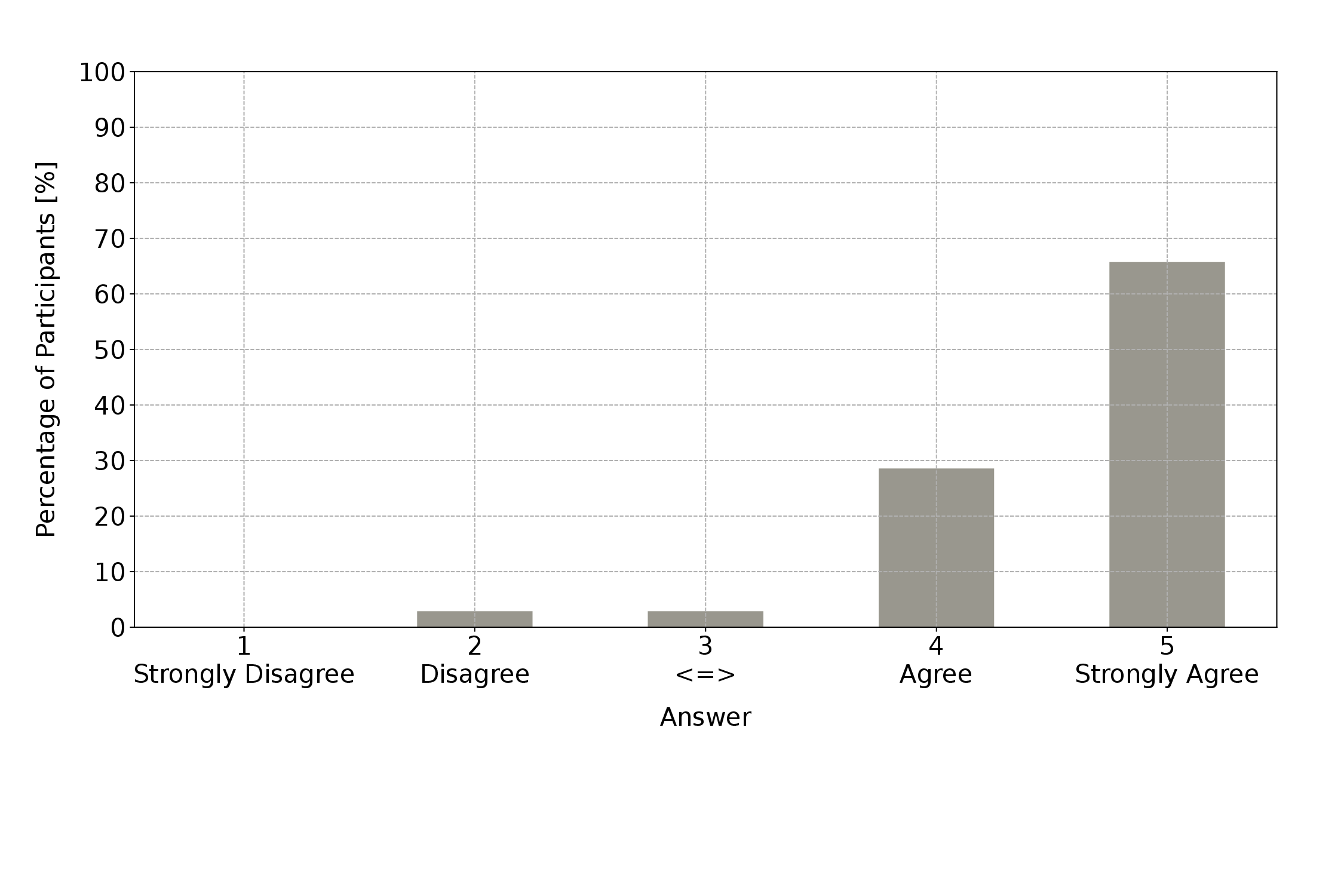}
        \vspace{-10mm}
        \caption{This robot is friendly}
        \label{re_na_a}
    \end{subfigure}
    \hfill
    \begin{subfigure}{0.45\linewidth}
        \centering
        \includegraphics[width=\linewidth]{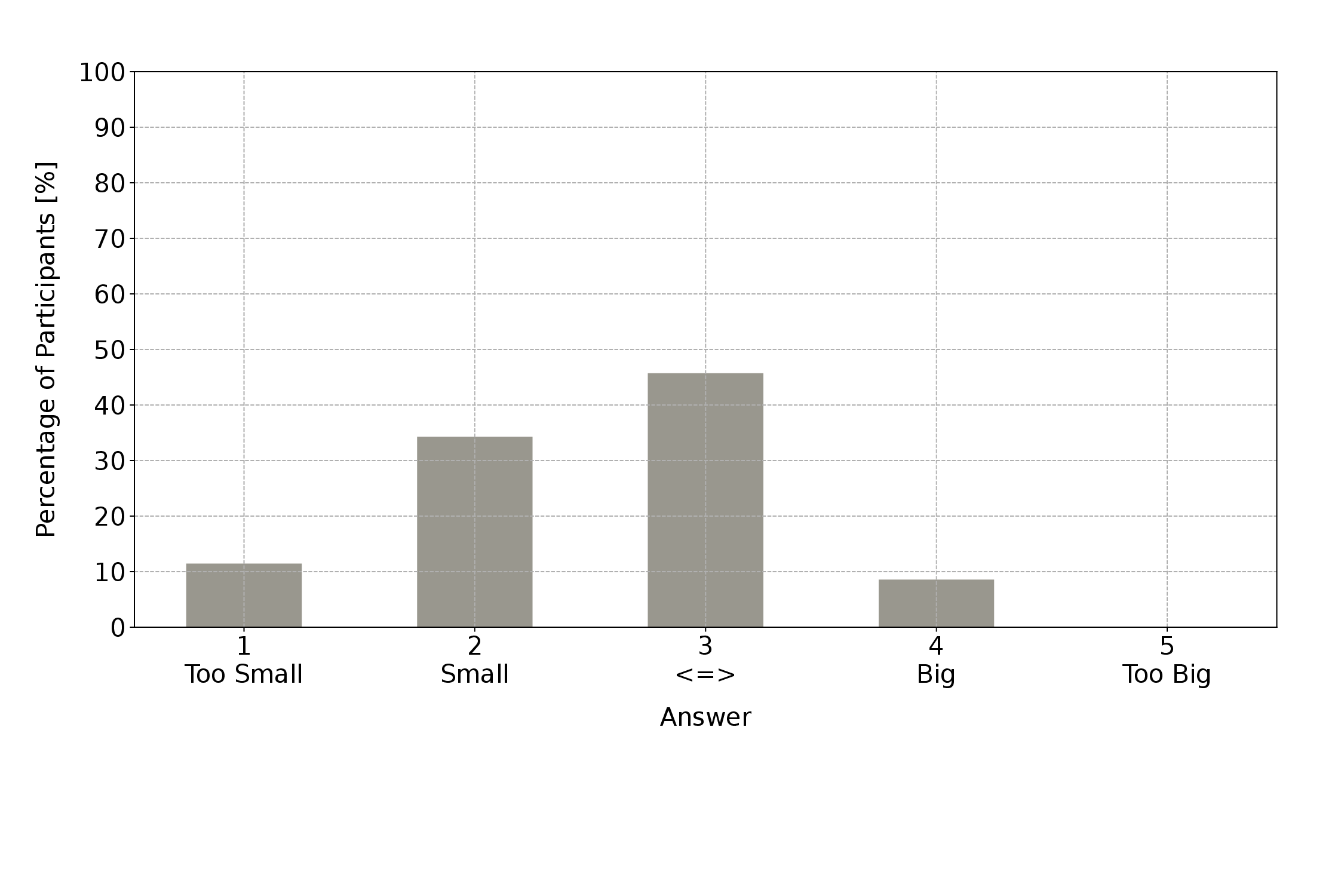}
        \vspace{-10mm}
        \caption{The size of this robot is}
        \label{re_na_b}
    \end{subfigure}

    \vspace{5mm}

    \begin{subfigure}{0.45\linewidth}
        \centering
        \includegraphics[width=\linewidth]{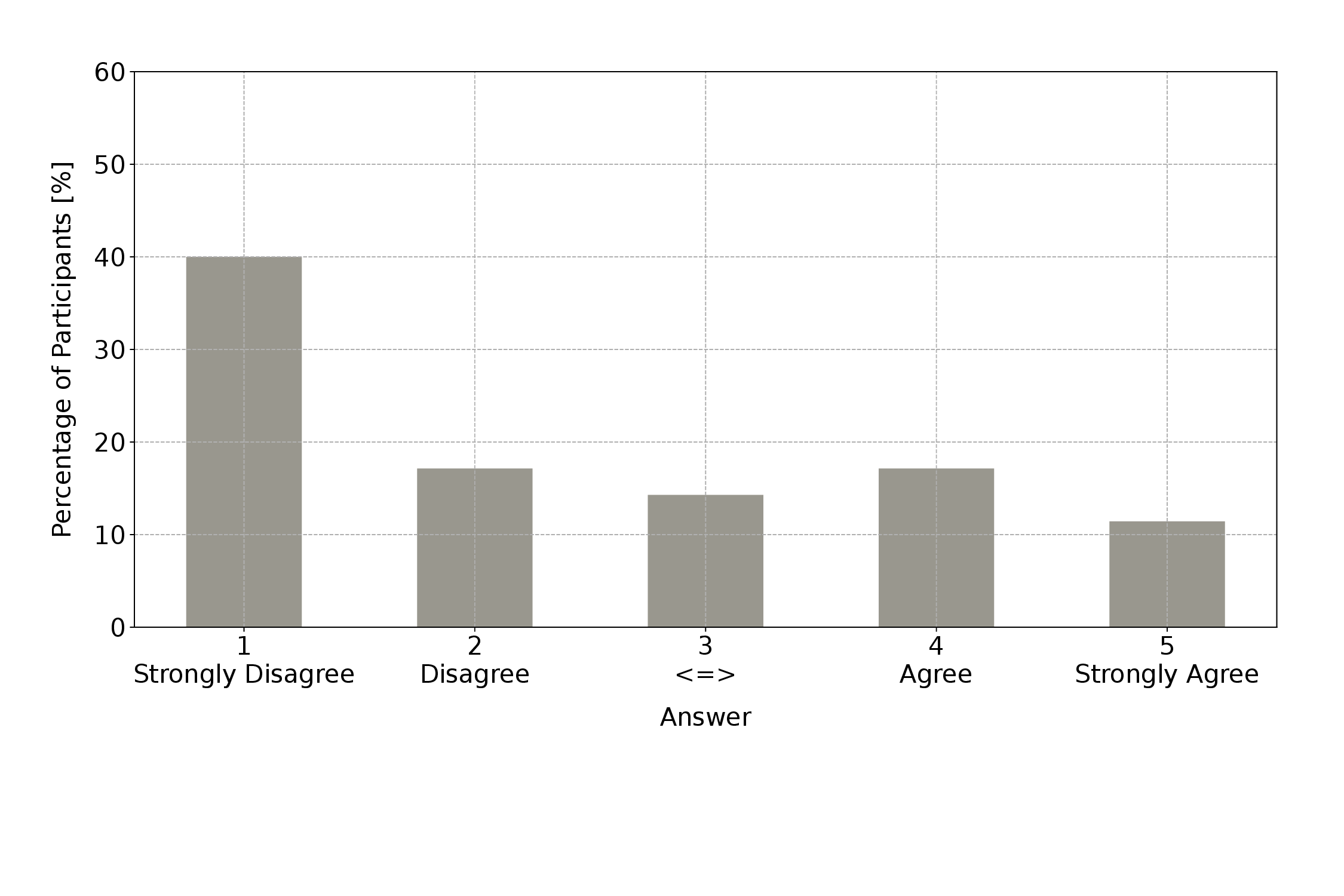}
        \vspace{-10mm}
        \caption{I want to touch this robot}
        \label{re_na_c}
    \end{subfigure}
    \hfill
    \begin{subfigure}{0.45\linewidth}
        \centering
        \includegraphics[width=\linewidth]{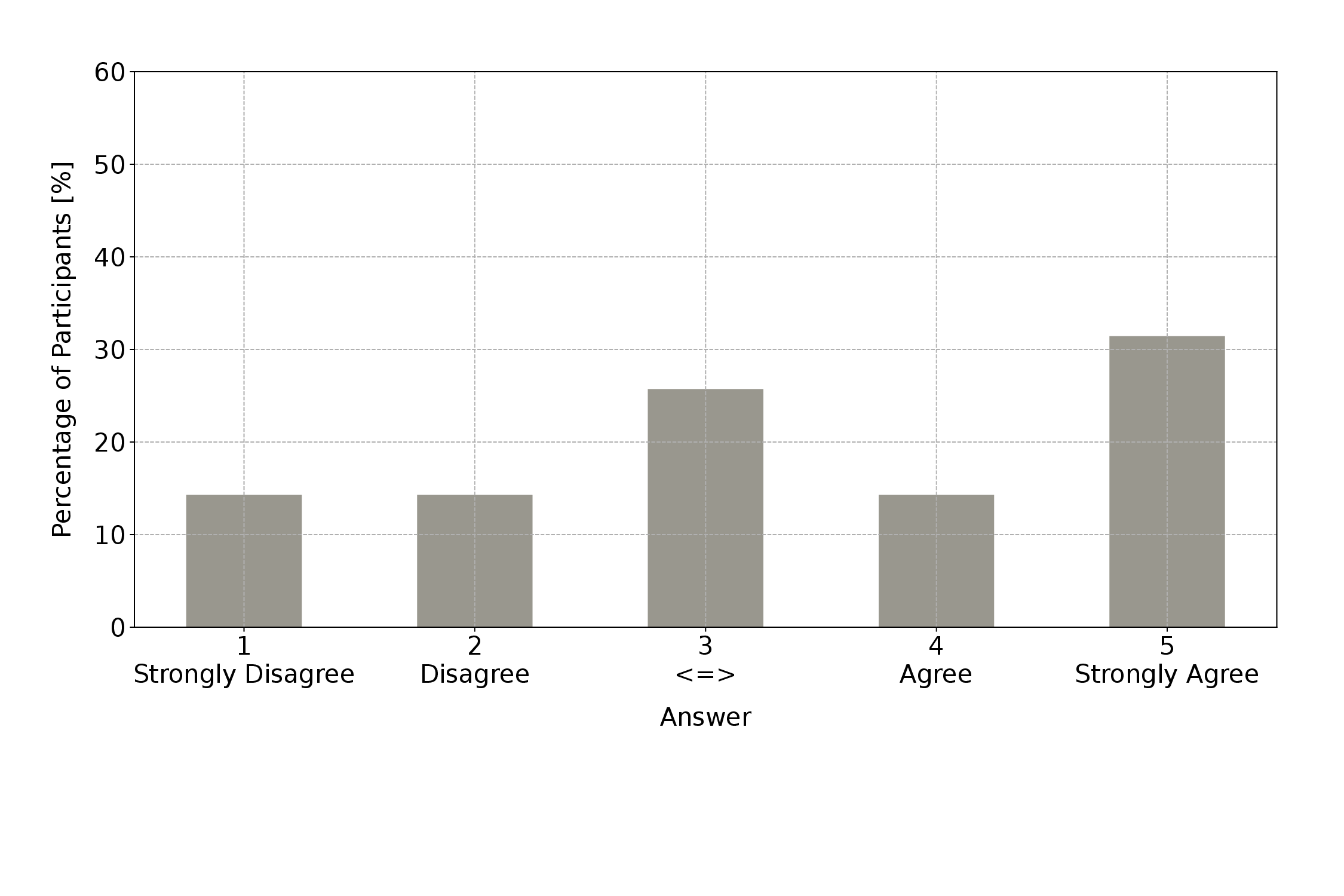}
        \vspace{-10mm}
        \caption{I want this robot to give me an attention}
        \label{re_na_d}
    \end{subfigure}

    \caption{Results of robot perception survey for all participants (1/2)}
    \label{fig:all_combined}
\end{figure*}

\begin{figure*}[t]
    \ContinuedFloat
    \centering

    \begin{subfigure}{0.45\linewidth}
        \centering
        \includegraphics[width=\linewidth]{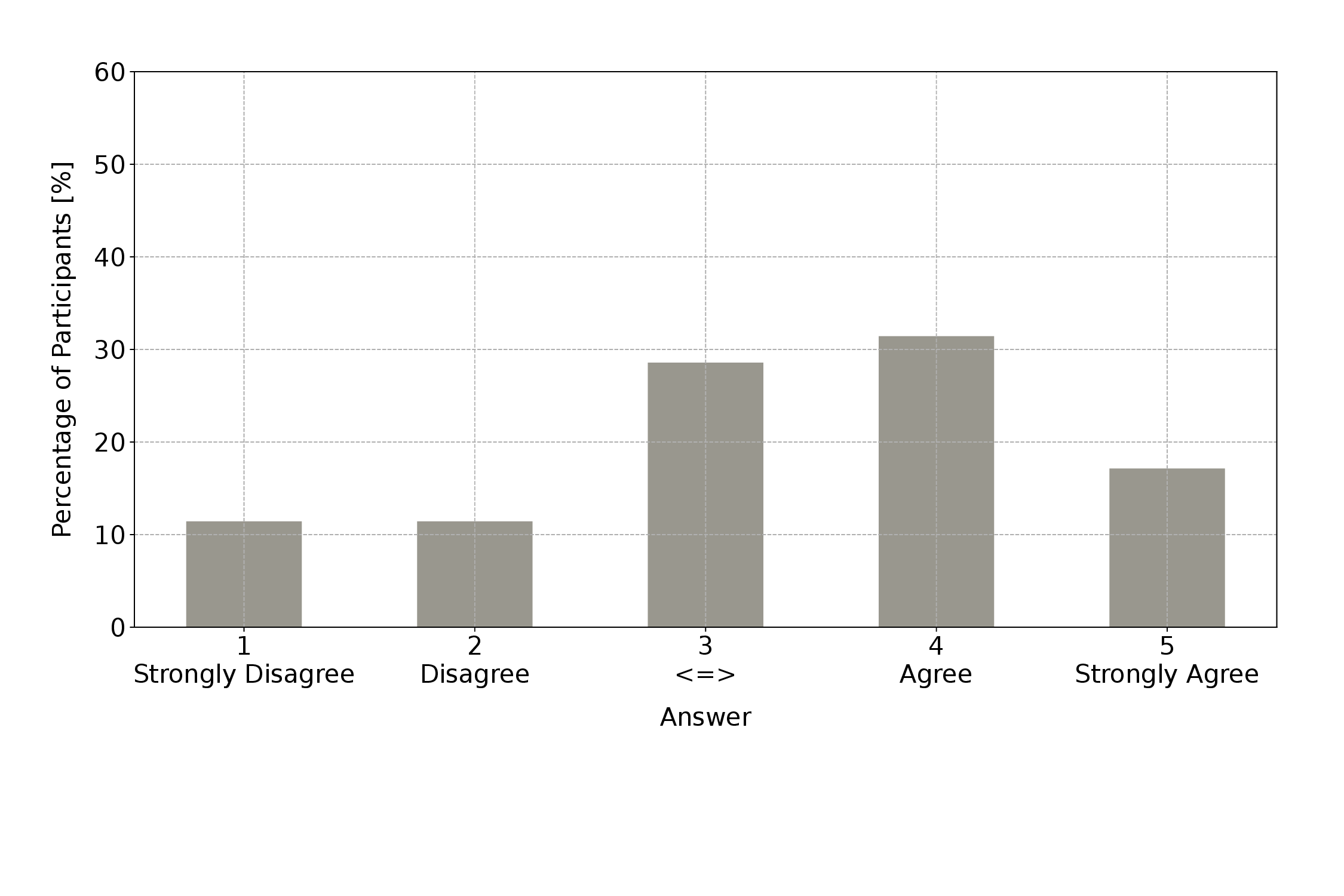}
        \vspace{-10mm}
        \caption{The way this robot speaks is natural}
        \label{re_na_e}
    \end{subfigure}
    \hfill
    \begin{subfigure}{0.45\linewidth}
        \centering
        \includegraphics[width=\linewidth]{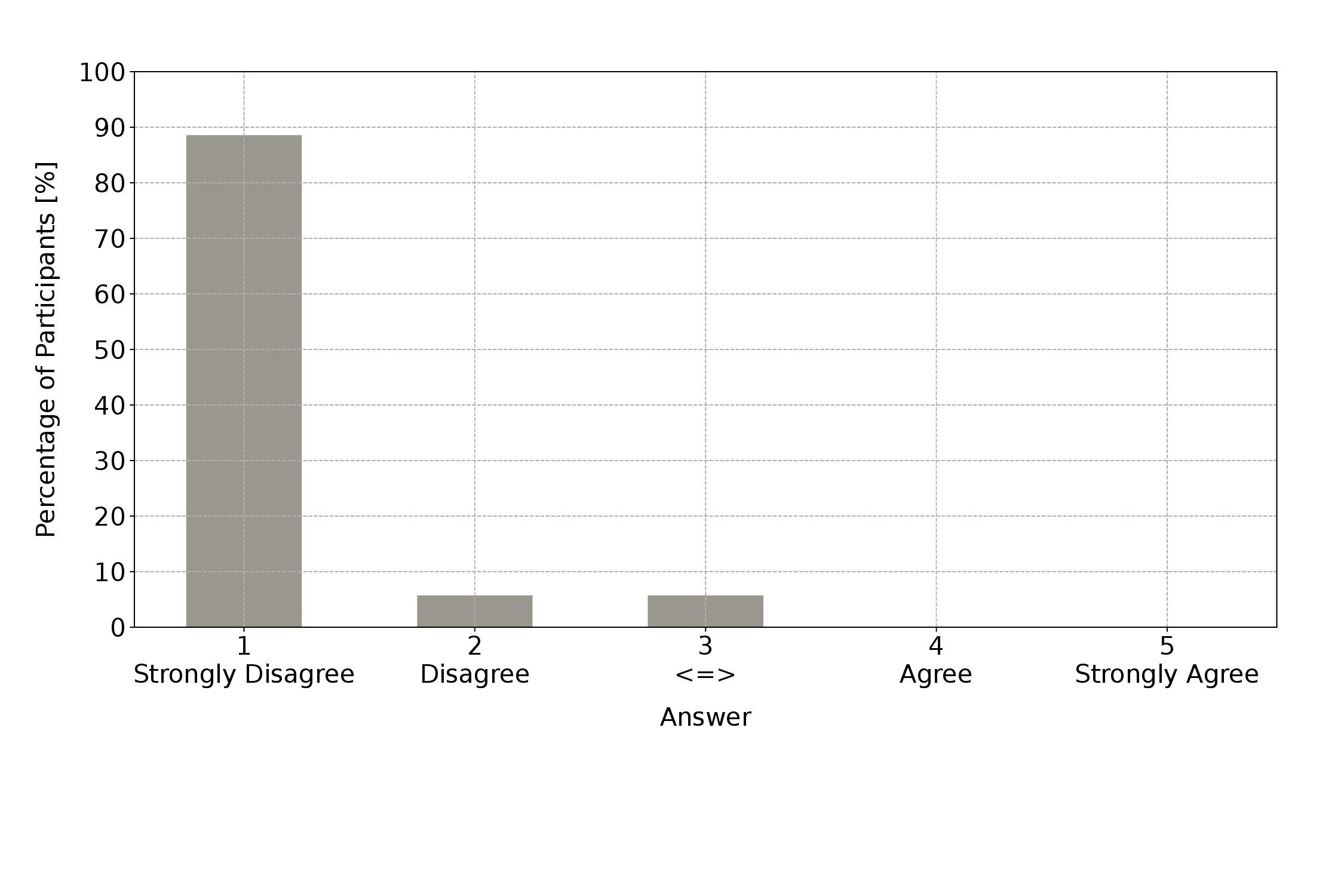}
        \vspace{-10mm}
        \caption{This robot is physically intrusive}
        \label{re_na_f}
    \end{subfigure}

    \vspace{5mm}

    \begin{subfigure}{0.45\linewidth}
        \centering
        \includegraphics[width=\linewidth]{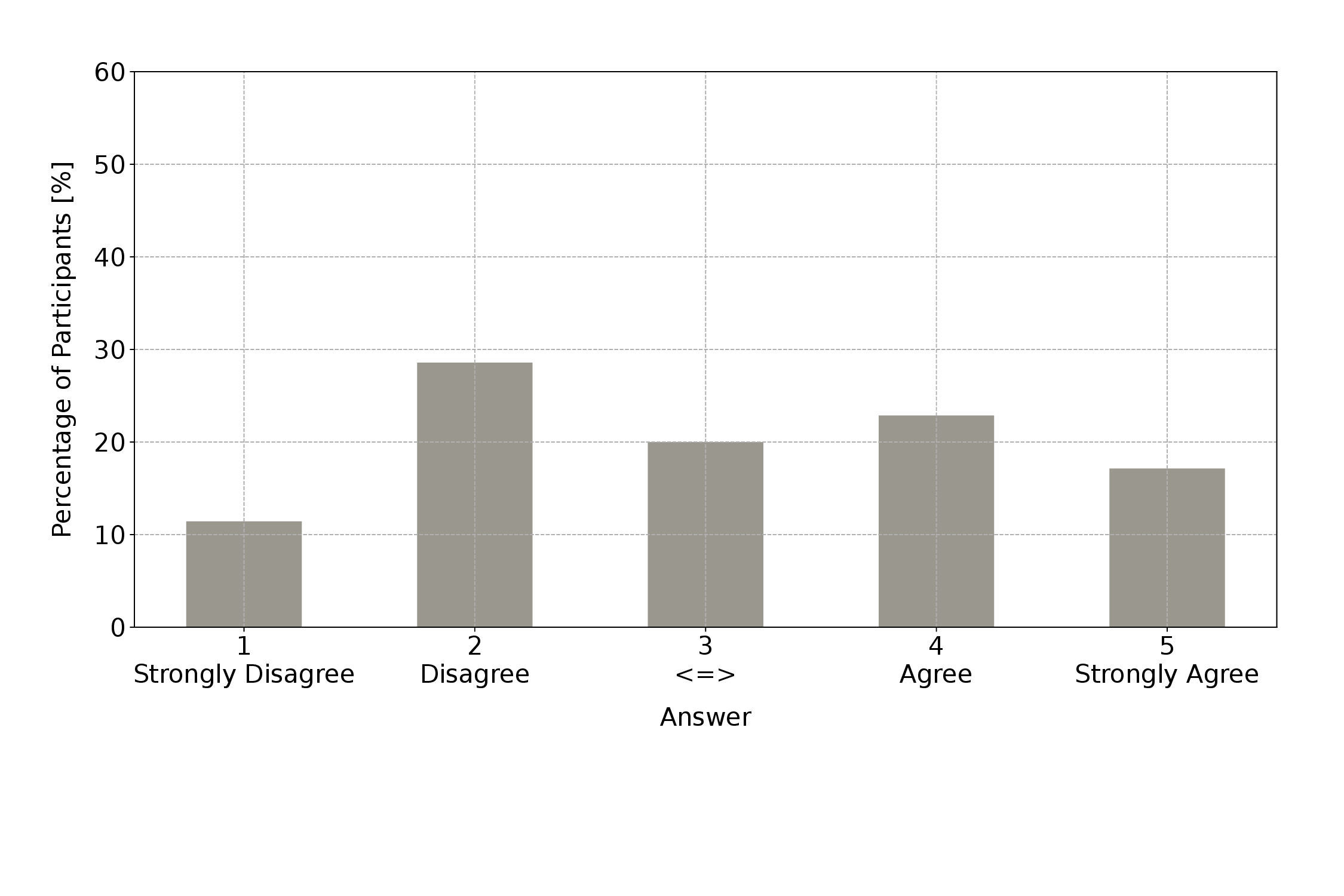}
        \vspace{-10mm}
        \caption{This robot is psychologically calming}
        \label{re_na_g}
    \end{subfigure}
    \hfill
    \begin{subfigure}{0.45\linewidth}
        \centering
        \includegraphics[width=\linewidth]{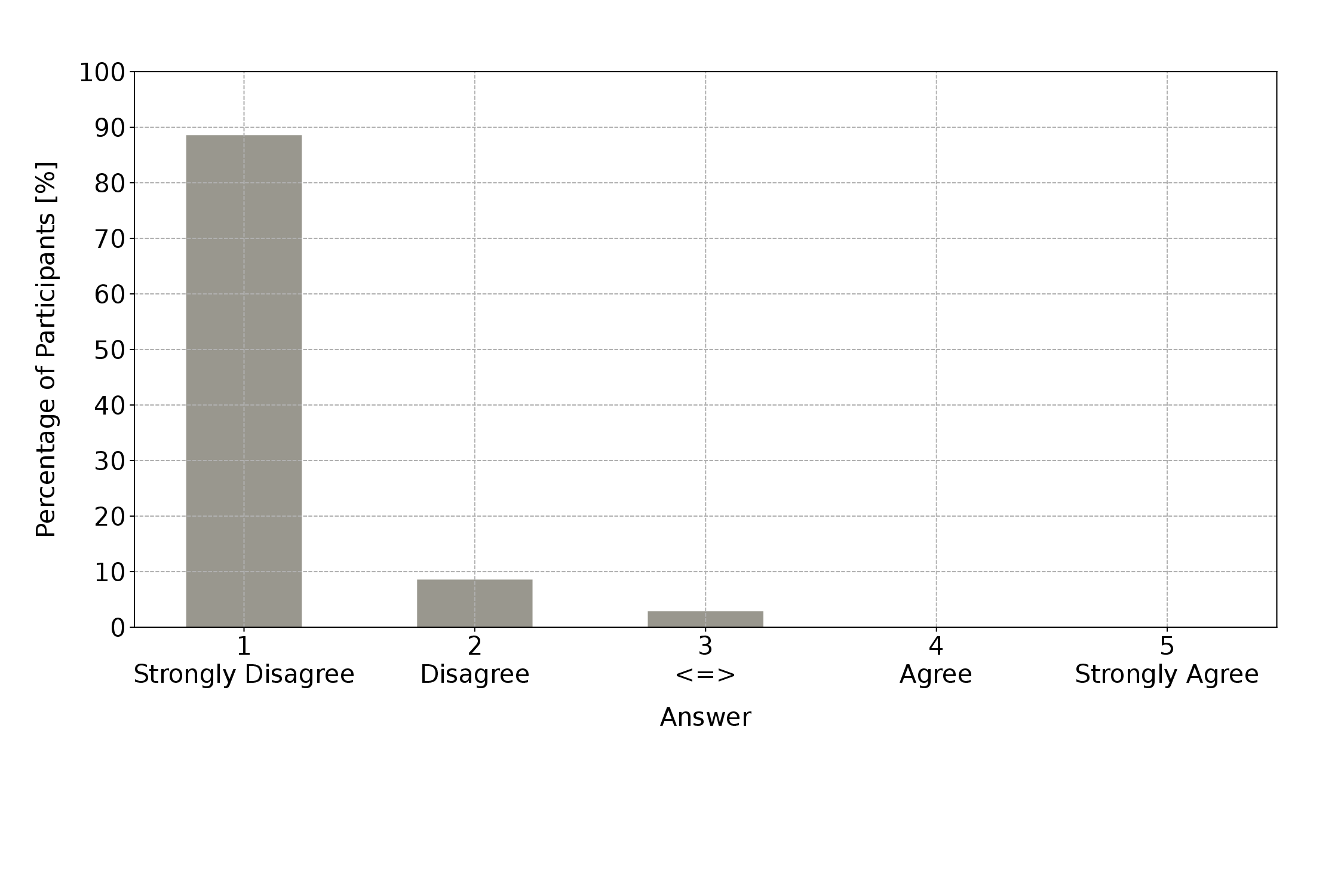}
        \vspace{-10mm}
        \caption{The sound of this robot moving is annoying}
        \label{re_na_h}
    \end{subfigure}

    \vspace{5mm}

    \begin{subfigure}{0.45\linewidth}
        \centering
        \includegraphics[width=\linewidth]{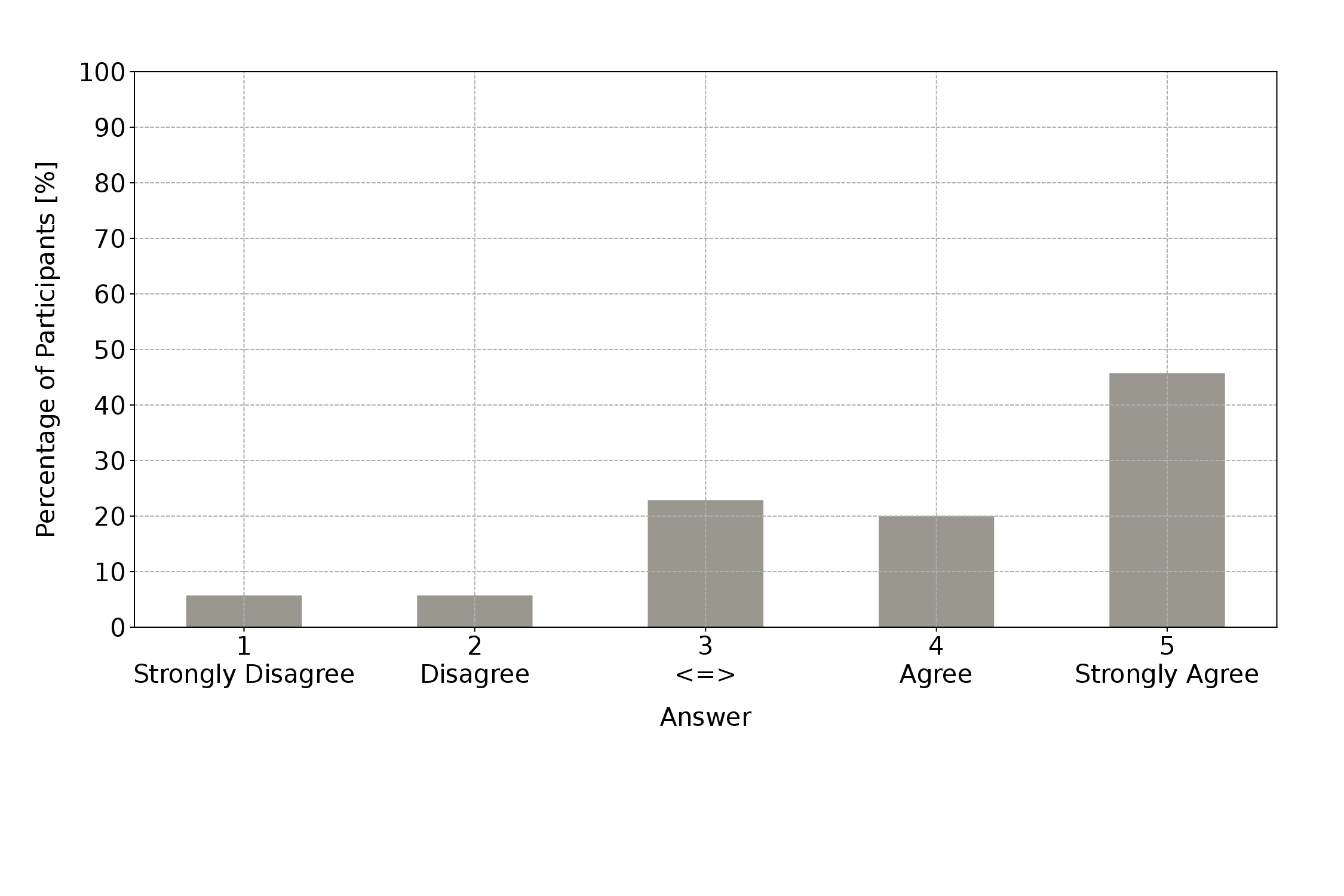}
        \vspace{-10mm}
        \caption{This robot seems safe}
        \label{re_na_i}
    \end{subfigure}
    \hspace*{\fill}

    \caption{Results of robot perception survey for all participants (2/2)\\ \small{While the robot was generally perceived as friendly (a) with low physical intrusiveness (f), results also indicate limited desire for proactive interaction (d) and some psychological anxiety (g).}}
\end{figure*}


\noindent\textbf{Comparison of Interaction Partners for RQ2}\\

\begin{table}[b]
\centering
\caption{ANOVA results for heart rate comparison between interaction partners at each phase}\label{hr_stat}
\begin{tabularx}{\columnwidth}{|X|c|c|}
\hline
Phase & $F$-statistic & $p$-value \\ \hline
a. Before Interaction 1   & 0.094 & .762 \\ \hline
b. During Interaction 1   & 2.538 & .121 \\ \hline
c. Between 2 Interactions & 0.561 & .459 \\ \hline
d. During Interaction 2   & 1.770 & .193 \\ \hline
e. After Interaction 2    & 0.229 & .636 \\ \hline
\end{tabularx}
\vspace{1mm}
{\footnotesize The distinct numerical difference between partners, though not statistically significant.}
\end{table}

Physiological responses showed a subtle numerical trend depending on the interaction partner. As shown in Fig. \ref{co_me}, mean heart rates tended to be lower during interactions with the robot compared to a human partner. For instance, in Condition 1, the mean heart rate was 76.2 bpm with the robot versus 79.0 bpm with a human. As shown in the Table \ref{hr_stat}, Statistical analysis at each phase revealed that while no results reached the threshold of $p < .05$, the difference during the interactions were the most pronounced (b. During Interaction 1: $F = 2.538, p = .121$ / d. During Interaction 2: $F = 1.770, p = .193$), suggesting a marginal trend toward lower physiological arousal when interacting with the robot. Furthermore, the SD of the heart rate, which serves as an indicator of physiological stability, was consistently smaller during robot interactions in Conditions 1, 2, and 3 (Fig. \ref{co_sd}). This lower variability suggests that participants maintained a more stable and relaxed physiological state when interacting with the robot than with a human partner.\\


\noindent\textbf{Comparison of Prompt Conditions for RQ2}\\
The influence of the positive prompt on heart rate was also evaluated. As illustrated in Fig. \ref{gr_me}, a slight increase in mean heart rate was observed when the prompt was provided. However, ANOVA results for each phase confirmed that these differences were not statistically significant (e.g., $p = .645$ for Phase B, $p = .799$ for Phase D). Similarly, no clear patterns or significant differences were found in the SD between the prompt and no-prompt groups (Fig. \ref{gr_sd}). These results indicate that the physiological impact of the prompt intervention was minimal, and the participants' cardiovascular responses were primarily influenced by the nature of the social interaction itself.\\


\noindent\textbf{RQ3: General Perception}

The overall results of the robot perception survey is shown in the Fig. \ref{fig:all_combined}. The question (a) and (f) show that most participants found the robot friendly, with only a small number finding it physically intrusive. On the other hand, the question (d) shows that few participants expressed a desire to interact with the robot by touching or talking to it. And the question (g) shows a certain number of participants felt psychologically anxiety.

To investigate the impact of experimental conditions, an ANOVA was performed on the scores for each question. No significant differences were found between conditions (e.g., $p = .772$ for friendliness, $p = .910$ for psychological calming), with the sole exception of question (h) regarding the robot's movement sound ($p = .015$). This consistency across conditions indicates that the overall trend in robot perception was stable regardless of the prompt or interaction order. All charts for each condition are provided in Appendix \ref{b_graph}.

The open-ended comments provided additional qualitative data regarding the participants' perceptions. Positive feedback included descriptions such as "cute," "interesting," and "good eyes." On the other hand, critical remarks were also recorded, including observations that the robot had a "cold expression," a "limited response repertoire," and was "not a substitute for a human conversation partner." Furthermore, some participants specifically pointed out an incongruity between the robot's appearance and the content of the dialogue, noting that the "childish appearance and non-childish questions do not match." A complete list of translated comments is provided in Appendix \ref{b_free}.


\subsection{Discussion}\label{sec3-4}
The main finding of this study is that for the participants, there is no significant difference in the stress experienced when interacting with an unfamiliar human and when interacting with a robot. Furthermore, it may even be that interacting with a robot is less stressful. At the same time, it was also noted that the robot was perceived as less reliable, suggesting that it does not serve as a complete substitute for human interaction. The following is a discussion of each outcome:\\

\subsubsection{RQ1: Facial Expressions}

\noindent\textbf{Comparison of Interaction Partners for RQ1}\\
The results show that participants smiled at the robot as frequently as they did at a human partner. Specifically, "Happy" was detected at the start of the interaction with the robot in 21 cases, a result almost identical to a human partner (22 cases). The number of participants detected more than 50\% of "Happy" during the interaction was also almost the idential(14 cases for a robot and 13 cases for a human).

A smiling expression conveys not only information about a person's emotional state but also serves as a clear signal of affiliative intent \cite{happy1}. Furthermore, The display of such positive emotions is crucial for facilitating social bonds, as emotional mimicry—which aids in understanding another's emotional state—is more likely to occur when the relationship between interlocutors is perceived as neutral or positive \cite{happy2}. Since the participants reciprocated the robot’s expressions with similar frequency to a human, it is suggested that they accepted the robot as a valid conversation partner.

So, the fact that more than half of the participants smiled for the majority of the session indicates that the robot is an effective tool for eliciting positive affect in older users.

Most participants' negative reactions remained below 25\%. Notably, in Conditions 1 and 2 where the robot was the first interaction partner, the number of participants with negative emotions was slightly lower than in other conditions. This suggests that the robot may alleviate the social pressure or discomfort often associated with answering private questions.

Furthermore, detailed review of the video data revealed that facial expressions categorized as "Sad" or "Disgust" frequently occurred when participants were concentrating on understanding a question or thinking of an answer. These results may not indicate negative affect, but rather reflect cognitive effort and concentration, as eyebrow movements (e.g., brow lowering) are often signal a lack of immediate understanding \cite{negative1} or mental effort during task engagement \cite{negative2}. Since automated recognition tools frequently categorize these brow-based movements as negative, it is likely that these detections primarily represent the participants' intensive concentration. Therefore, these results should be interpreted as a sign of intensive concentration during the task.\\

\noindent\textbf{Comparison of Prompt Conditions for RQ1}\\
The statistical analysis revealed no significant difference between the presence and absence of positive prompts, indicating that the intervention did not fundamentally alter the participants' emotional displays. However, looking at the raw percentages, the no-prompt conditions showed a slightly higher proportion of "Happy" expressions (Category B) and a more stable suppression of negative emotions (Category D).

While the prompt was intended to encourage positive thinking, the fact that it did not lead to a statistically significant increase in positive affect suggests several possibilities. One reason might be that the prompts were almost identical each time, which could have caused participants to perceive the interaction as repetitive or redundant. Previous studies on conversational agents have reported that repetitive responses can negatively affect user experience and dialogue quality \cite{prompt1}. Moreover, monotonous repetitive utterances may quickly cause users to lose interest in continuing the dialogue \cite{prompt2}. Additionally, the extra dialogue required for the prompt made the conversation longer, which might have subtly interfered with the natural flow of the interaction, as multi-turn dialogues for simple tasks can impose additional interaction burden on users \cite{prompt3}. These numerical trends, though not statistically significant, suggest that natural and concise dialogue may be just as effective—if not more so—than structured psychological interventions for maintaining a positive user experience in social robotics.


\subsubsection{RQ2: Heart rate}
\noindent\textbf{Comparison of Interaction Partners for RQ2}\\
Heart rates during interactions with the robot tended to be lower and the standard deviation smaller. Although the ANOVA did not show a statistically significant difference at the $p < .05$ level, a marginal trend was observed during the first interaction (Phase B, $p = .121$), where the heart rate was notably lower with the robot. This suggests that interactions with the robot were less tense than those with a human, allowing participants to converse in a more relaxed state. This suggests that interactions with the robot were less tense than interactions with a human, allowing participants to converse in a relaxed state. This interpretation is consistent with well-established findings in psychophysiology showing that reduced heart rate and diminished short-term variability are markers of decreased sympathetic arousal and greater calmness \cite{HR}. Such physiological indicators are commonly used to infer reduced levels of stress and tension in social interactions. A calmer state may be advantageous in that it may facilitate both more pleasant conversations and more accurate information provision by older adults.\\

\noindent\textbf{Comparison of Prompt Conditions for RQ2}\\
Regarding the prompt intervention, heart rates were slightly higher when the positive prompt was provided, though no significant difference was observed ($p > .64$). This slight elevation could be interpreted in two ways: it may reflect a subtle increase in anxiety due to the additional questioning, or conversely, a positive state of physiological arousal (excitement) triggered by reflecting on personal topics such as family. However, given that there were no significant differences in the standard deviation and no clear patterns emerged across other phases, the influence of the positive prompt on cardiovascular response in this study was not considered noteworthy. These results suggest that while the prompt did not cause significant distress, its current form may not be powerful enough to fundamentally alter the physiological state of the participants. Future research could explore whether modifying the content or delivery of the prompt—making it more personalized or less redundant—might elicit a more distinct physiological effect.\\


\subsubsection{RQ3: General Perception}

33 out of 35 participants found the robot friendly and easy to use, suggesting its potential as an interaction partner. According to the comments, the robot's "sweet" and "cute" appearance, combined with its clear speech and responsive eye movements, created a pleasant and non-judgmental atmosphere. Some participants even noted that the robot's presence could be beneficial for well-being or as a stimulating companion in nursing homes, especially for those who are mentally fit and interested in technology. In other words, a friendly aura and responsive non-verbal cues are essential for a robot to be accepted in a social context.

However, participants did not necessarily actively interact with the robot, revealing a clear distinction between a "technical device" and a "human partner." A recurring sentiment was that while the robot is a helpful tool, it is not a substitute for human interaction ("people cannot be replaced yet"). Specifically, th
e "childish appearance" combined with "non-childish questions," such as inquiries regarding life satisfaction, created a mismatch that led to perceptions of the robot as "machine-like" or "inauthentic." Matching a robot’s appearance and behavior to its specific task is crucial for managing user expectations and improving cooperation \cite{match}. In this study, the serious nature of the questions likely conflicted with the participants' expectations of a child-like entity, contributing to the reported psychological unease and the perception of a "cold expression" or "limited repertoire."

These results suggest that while a child-like appearance is advantageous for reducing social pressure—as evidenced by the lower heart rates and suppressed negative expressions—it may also limit the depth of engagement if the dialogue strategy is not carefully aligned with that persona. To bridge this gap, future designs must ensure that the robot's appearance and behavioral role are consistent. Furthermore, it is essential to define the robot's role clearly as a supplement to, rather than a replacement for, human contact. Acknowledging the current limitations of social robots while leveraging their ability to provide a "safe" and non-judgmental environment will be key to their successful integration into healthcare settings for the older adults.


\subsubsection{Overall Evaluation}
Based on the findings from facial expressions (RQ1), herat rate (RQ2), and general perception (RQ3), the overall evaluation of the interaction is discussed by integrating psychological and physiological perspectives.\\

\noindent\textbf{Comparison of Interaction Partners}\\
Consistent with the emotional responses in RQ1 and the physiological data in RQ2, there was no significant difference in the overall results between interacting with a human and the robot. This stability suggests that the participants accepted the robot as a legitimate interaction partner. Notably, using the same experimental setup and participant group, Mayer et al. analyzed a different set of metrics and similarly found no significant differences across these conditions. This is reinforced by the findings of Mayer et al., who analyzed the same participant group using different metrics and similarly found no significant psychological burden imposed by the robot \cite{mayer}.\\

\noindent\textbf{Comparison of Prompt Conditions}\\
Regarding the presence of prompts, the results from RQ1: Facial Expressions and RQ2: Heart Rate indicate that the positive prompts did not produce the expected emotional or physiological benefits. Although numerical trends in RQ1 and RQ2 showed slightly more positive responses without the prompts, these were statistically non-significant. Furthermore, similar to the results observed for the interaction partners, there were no significant differences in psychological or physiological stress markers regardless of the presence or absence of prompts \cite{mayer}. This is likely due to the redundant nature of the prompts and the increased interaction length, which may have led to participant fatigue rather than enhanced well-being. This suggests that for older users, concise and natural dialogue is more effective than structured psychological interventions.\\

\noindent\textbf{Limitations and Future Directions}\\
While the robot is perceived as a friendly and low-stress partner for structured tasks such as health surveys, its effectiveness is currently limited by a mismatch between its childlike appearance and adult-oriented conversational content. To avoid over-generalization, it must be noted that while the robot provides a relaxed environment—sometimes even more so than a human partner—it is not yet a functional substitute for the flexible and natural communication offered by a human. Future improvements should focus on aligning the robot's persona with the interaction context to overcome the identified communicative gaps.

\section{Conclusion and Future Work}\label{sec4}
\subsection{Conclusion}\label{sec4-1}
This study evaluated the effectiveness of social robots as interaction partners for older adults compared to human caregivers. The primary findings, addressing our two overarching research questions, are summarized below:

\begin{itemize}
    \item \textbf{RQ1 \& RQ2 (Partner Comparison)}: Interaction with the robot elicited facial emotions comparable to those observed with human nurses, suggesting high acceptance of the robot as a conversation partner. Physiological data showed lower mean heart rates and smaller standard deviations during robot encounters, indicating a more relaxed state.
    \item \textbf{RQ1 \& RQ2 (Positive Prompts)}: The positive prompts did not produce significant emotional or physiological benefits. Instead, increased interaction length appeared to lead to participant fatigue.
    \item \textbf{RQ3 (General Perception)}: Participants generally perceived the robot as friendly, but its acceptance was hindered by a mismatch between its childlike appearance and adult-oriented conversational content.
\end{itemize}

In conclusion, our results demonstrate that social robots can effectively support structured care tasks without imposing an additional psychological burden on old users. This conclusion, derived from our unique physiological and emotional metrics, reinforces and is consistent with the psychological approach \cite{mayer}. To bridge the identified communicative gaps, future development must ensure that a robot's visual persona is harmonized with its conversational role.

\subsection{Future Work}\label{sec4-2}
To enhance the effectiveness of robot-led interactions, future research should focus on the following:

\begin{itemize}
\item \textbf{Persona and Task Alignment}: Improving robot design to ensure its appearance, voice, and behavior are harmoniously aligned with the social context and age of the user. Specifically, future studies should employ humanoid robots with more mature appearances and voices to resolve the mismatch between a childlike persona and adult-oriented conversational content.
\item \textbf{Optimization of Interaction Content}: Focusing on facilitating natural and fluid conversation to avoid redundant prompts and maximize emotional engagement. Advanced AI is expected to enable this evolution by generating real-time, adaptive reactions tailored to each user.
\item \textbf{Physical and Emotional Support}: Exploring the role of physical contact or more nuanced non-verbal cues to better simulate the holistic support provided by human caregivers.
\end{itemize}

\backmatter

\bmhead{Supplementary information}

\bmhead{Acknowledgements}

We are deeply grateful to Beat Simon Vincent Iwan, Andrea Tempel and Elias Staatz (Heidelberg University) for their essential support in performing the experiments presented in this paper. Their diligent contribution was crucial to the successful completion of this research.

\vspace{15mm}
\section*{Declarations}

\begin{itemize}
\item \textbf{Funding}: This work was supported by the Hector Fellow Academy and the Heidelberg Karlsruhe Strategic Partnership (HEIKA) and also: Dres. Majic/Majic-Schlez-Foundation at Medical Faculty Heidelberg to CJM.
\item \textbf{Conflict of interest}: The authors declare that they have no competing interests.
\item \textbf{Ethical approval}: Ethical approval was granted by the Ethics Committee of the Medical Faculty at Heidelberg University (S-290/2024). All participants provided informed consent in accordance with the Declaration of Helsinki. The larger overall project is preregistered (https://osf.io/9dpxw) and approved.
\item  \textbf{Informed consent}: Informed consent was obtained from all individual participants included in the study.
\item \textbf{Consent for publication}: The participants provided informed consent regarding the publication of their anonymized data and any images included in this manuscript.
\item \textbf{Data availability}: The datasets generated and analyzed during the current study are not publicly available due to the privacy of the elderly participants but are available from the corresponding author on reasonable request.
\item \textbf{Materials availability}: The social robot Navel and other sensors used in this study are a commercially available product. The specific conversational scenarios and questionnaire items used during the interaction are available from the corresponding author on reasonable request.
\item \textbf{Code availability}: The code used for the robotic system setup and data analysis is available from the corresponding author on reasonable request.
\item \textbf{Author contributions}: All authors contributed to the study conception and design. HY conducted the technical setup, data collection, and analysis for the robotic system. CJM and CR were responsible for the psychological experimental setup, data collection, and analysis. The first draft of the manuscript was written by HY, and all authors contributed to subsequent revisions. All authors read and approved the final manuscript.
\end{itemize}

\begin{appendices}
\section{Results of Facial Expression}\label{appendix_emotion}
\subsection{Emotion Log and Pie Chart}\label{A_all_log}
The 35 experimental participants, 9, 10, 7, and 9 were randomly assigned to conditions 1 to 4, respectively. There is no record of the robot for experiments 05 and 17, and a human nurse for experiments 07 and 27, due to recording failure. The parts that were not recorded were due to participants not facing the camera and a failure of the video camera recording. Light green, pink, blue, red and yellow representing ’neutral’, ’happy’, ’sad’, ’disgust’, and ’surprised’, respectively.

\begin{figure*}[p]
\centering
\includegraphics[width=0.92\textwidth]{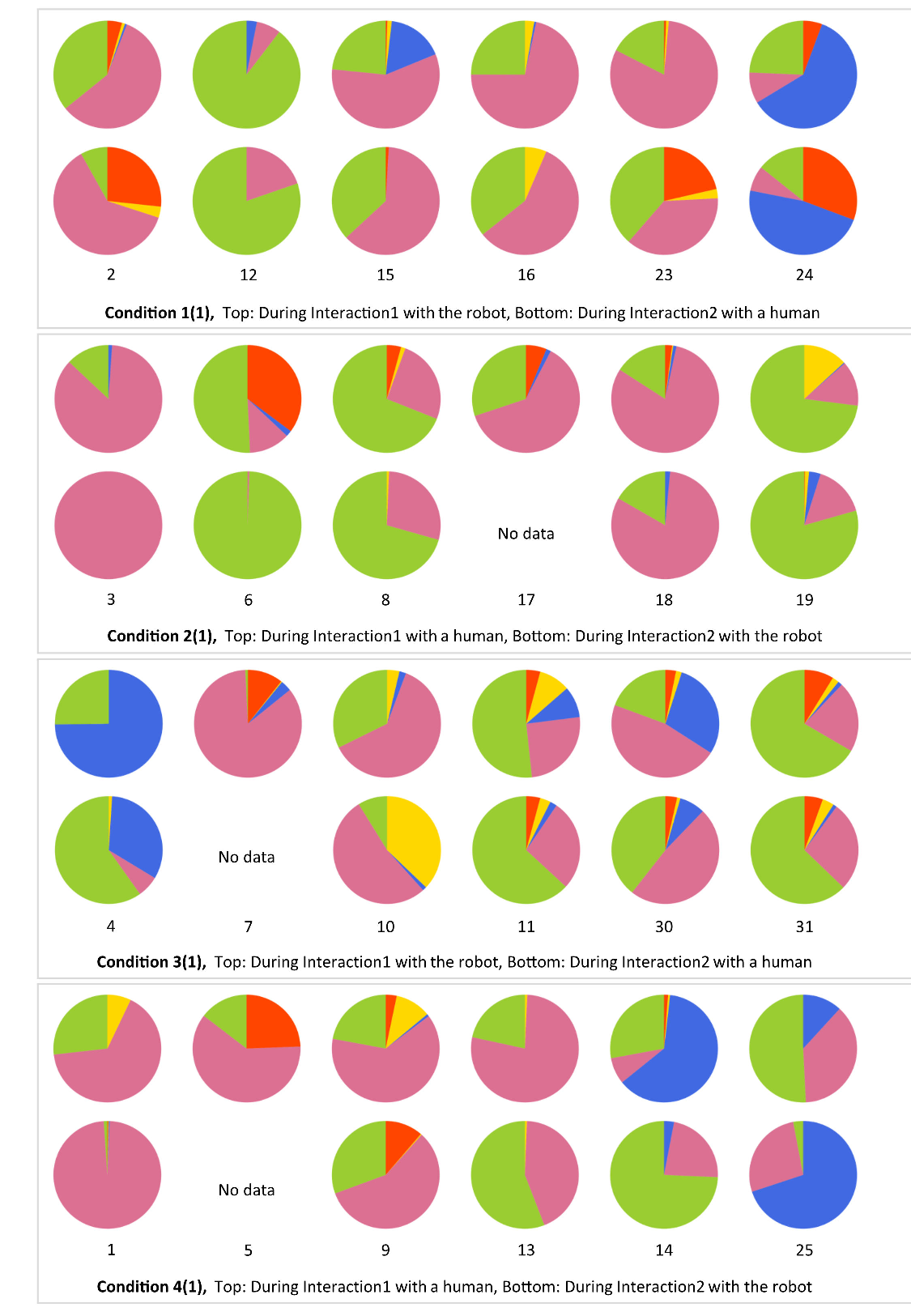}
\caption{Pie Chart of Emotion(1)}\label{pie1}
\end{figure*}

\clearpage 

\begin{figure*}[p]
\ContinuedFloat
\centering
\includegraphics[width=0.92\textwidth]{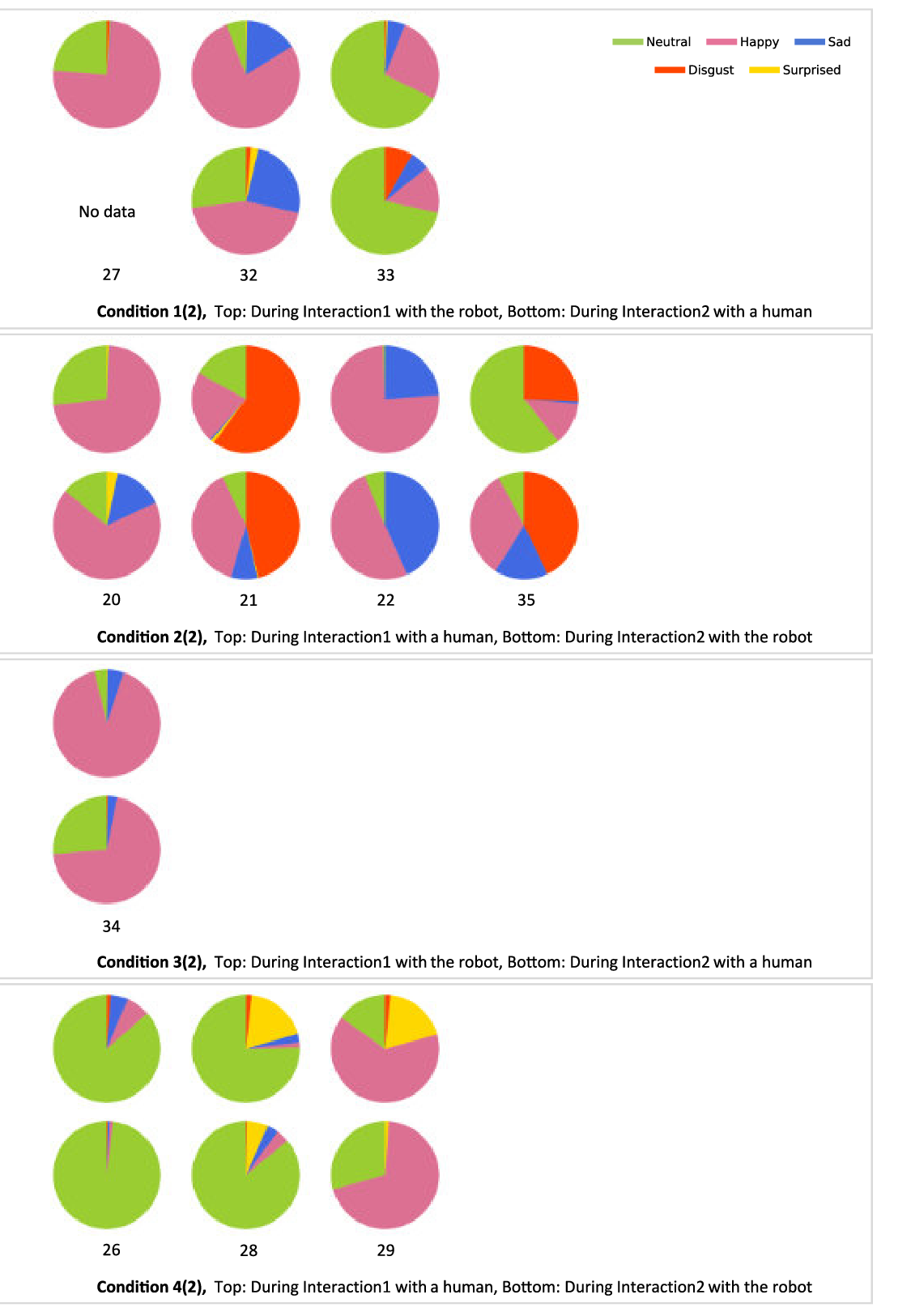}
\caption{Pie Chart of Emotion(2)}\label{pie2}
\end{figure*}

\begin{figure*}[p]
\centering
\includegraphics[width=0.92\textwidth]{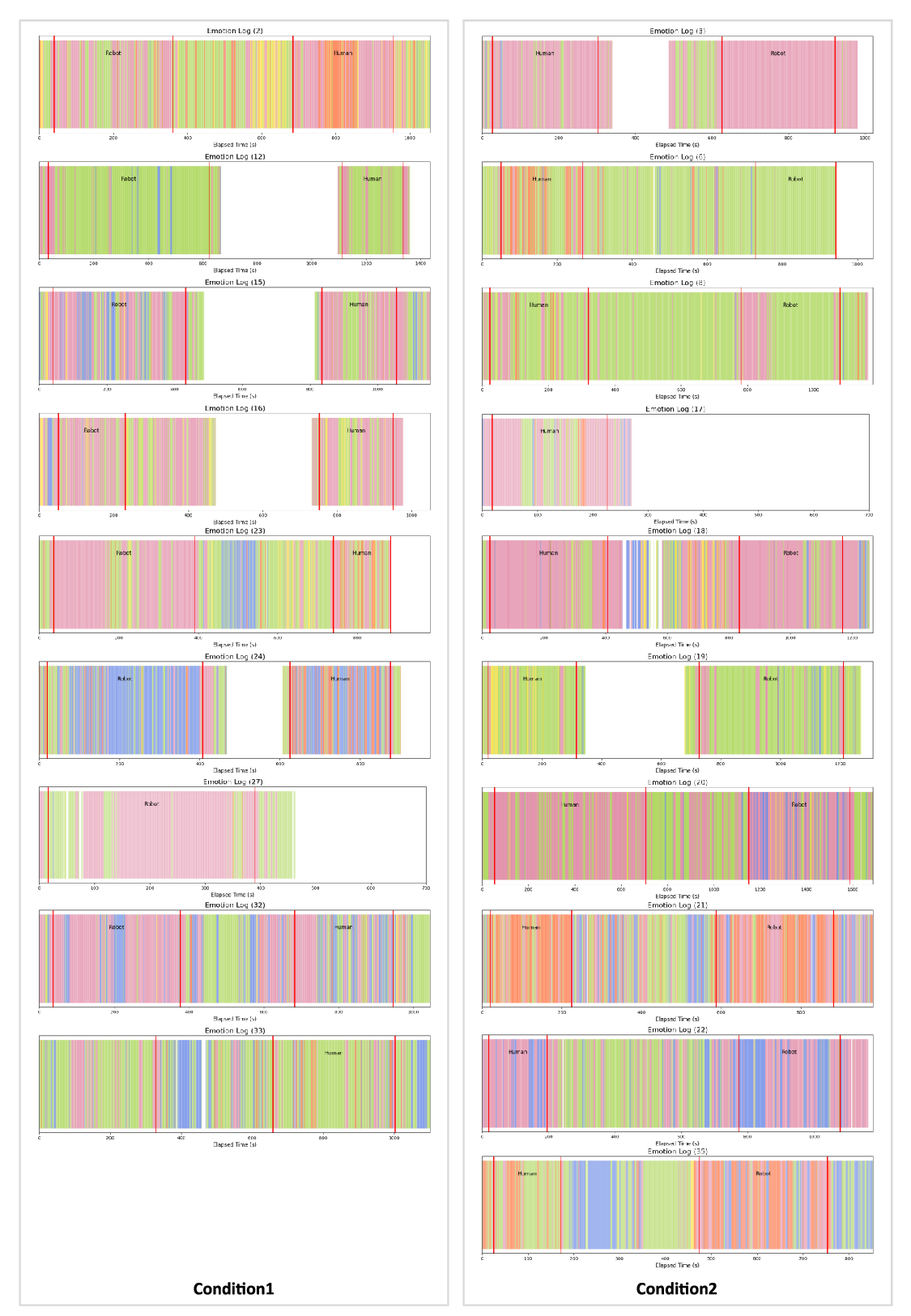}
\caption{Emotion Log(1)}\label{log1}
\end{figure*}

\clearpage 

\begin{figure*}[p]
\ContinuedFloat
\centering
\includegraphics[width=0.92\textwidth]{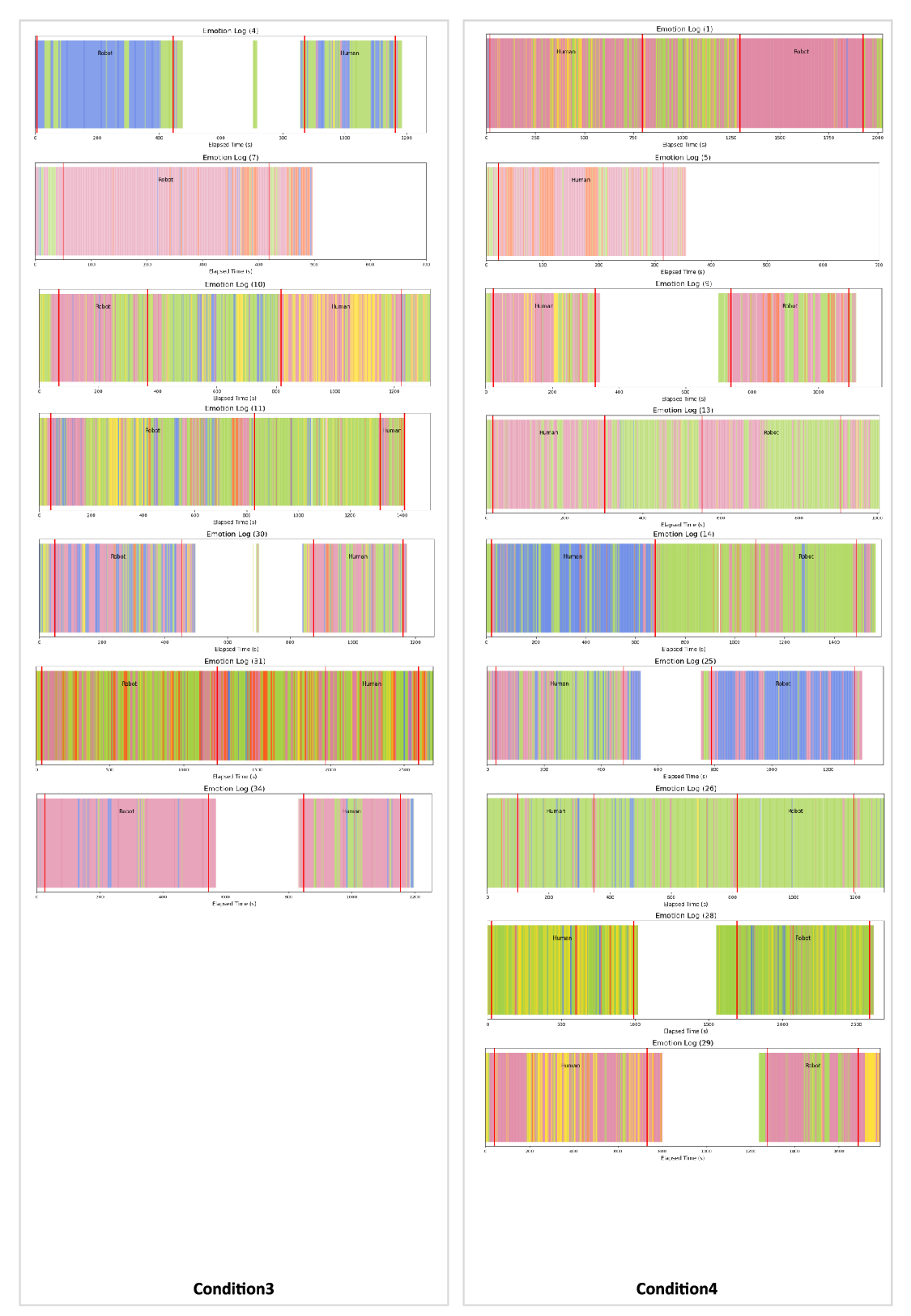}
\caption{Emotion Log(2)}\label{log2}
\end{figure*}

\clearpage
\subsection{Breakdown for All Categories}\label{A_breakdown}

\begin{center}
\small

\begin{minipage}{\textwidth}
\centering
\captionof{table}{(A) The number of participants detected 'Happy' when the interaction partner came into the room}\label{re_1}
\begin{tabularx}{\textwidth}{|X||c|c|c|c|c|}
\hline
\multicolumn{1}{|c||}{} & condition1 & condition2 & condition3 & condition4 & SUM \\ 
\noalign{\hrule height 1.5pt}
Only Robot              & 2 & 0 & 0 & 1 & 3 \\ \hline
Only Human              & 0 & 3 & 1 & 0 & 4 \\ \hline
Both Robot and Human    & 5 & 5 & 4 & 4 & 18 \\ \hline
Neither Robot nor Human & 1 & 1 & 1 & 3 & 6 \\ \hline
\textcolor{gray!50}{Fail to record} & \textcolor{gray!50}{1} & \textcolor{gray!50}{1} & \textcolor{gray!50}{1} & \textcolor{gray!50}{1} & \textcolor{gray!50}{4} \\ \hline
\noalign{\hrule height 1.5pt}
\textbf{SUM}            & \textbf{9} & \textbf{10} & \textbf{7} & \textbf{9} & \textbf{35} \\ \hline
\end{tabularx}
\end{minipage}

\vspace{1.2em}

\begin{minipage}{\textwidth}
\centering
\captionof{table}{(B) The number of participants detected more than 50\% of 'Happy' during the interaction}\label{re_2}
\begin{tabularx}{\textwidth}{|X||c|c|c|c|c|}
\hline
\multicolumn{1}{|c||}{} & condition1 & condition2 & condition3 & condition4 & SUM \\ 
\noalign{\hrule height 1.5pt}
Only Robot              & 2 & 0 & 0 & 0 & 2 \\ \hline
Only Human              & 0 & 0 & 0 & 1 & 1 \\ \hline
Both Robot and Human    & 3 & 4 & 2 & 3 & 12 \\ \hline
Neither Robot nor Human & 3 & 5 & 4 & 4 & 16 \\ \hline
\textcolor{gray!50}{Fail to record} & \textcolor{gray!50}{1} & \textcolor{gray!50}{1} & \textcolor{gray!50}{1} & \textcolor{gray!50}{1} & \textcolor{gray!50}{4} \\ \hline
\noalign{\hrule height 1.5pt}
\textbf{SUM}            & \textbf{9} & \textbf{10} & \textbf{7} & \textbf{9} & \textbf{35} \\ \hline
\end{tabularx}
\end{minipage}

\vspace{1.2em}

\begin{minipage}{\textwidth}
\centering
\captionof{table}{(C) Number of participants with a higher percentage of 'Happy' during the interaction}\label{re_3}
\begin{tabularx}{\textwidth}{|X||c|c|c|c|c|}
\hline
\multicolumn{1}{|c||}{} & condition1 & condition2 & condition3 & condition4 & SUM \\ 
\noalign{\hrule height 1.5pt}
Robot                   & 5 & 6 & 2 & 4 & 17 \\ \hline
Human                   & 3 & 3 & 4 & 4 & 14 \\ \hline
\textcolor{gray!50}{Fail to record} & \textcolor{gray!50}{1} & \textcolor{gray!50}{1} & \textcolor{gray!50}{1} & \textcolor{gray!50}{1} & \textcolor{gray!50}{4} \\ \hline
\noalign{\hrule height 1.5pt}
\textbf{SUM}            & \textbf{9} & \textbf{10} & \textbf{7} & \textbf{9} & \textbf{35} \\ \hline
\end{tabularx}
\end{minipage}

\vspace{1.2em}

\begin{minipage}{\textwidth}
\centering
\captionof{table}{(D) The number of participants detected less than 25\% of 'Sad' and 'Disgust' during the interaction}\label{re_4}
\begin{tabularx}{\textwidth}{|X||c|c|c|c|c|}
\hline
\multicolumn{1}{|c||}{} & condition1 & condition2 & condition3 & condition4 & SUM \\ 
\noalign{\hrule height 1.5pt}
Only Robot              & 3 & 2 & 0 & 1 & 6 \\ \hline
Only Human              & 0 & 2 & 1 & 1 & 4 \\ \hline
Both Robot and Human    & 5 & 4 & 4 & 6 & 19 \\ \hline
Neither Robot nor Human & 0 & 1 & 1 & 0 & 2 \\ \hline
\textcolor{gray!50}{Fail to record} & \textcolor{gray!50}{1} & \textcolor{gray!50}{1} & \textcolor{gray!50}{1} & \textcolor{gray!50}{1} & \textcolor{gray!50}{4} \\ \hline
\noalign{\hrule height 1.5pt}
\textbf{SUM}            & \textbf{9} & \textbf{10} & \textbf{7} & \textbf{9} & \textbf{35} \\ \hline
\end{tabularx}
\end{minipage}

\end{center}

\clearpage
\section{Results of Questionnaire}\label{appendix_questionnaire}
\subsection{Results of Question (a)-(i)}\label{b_graph}
Figure 7 shows the overall question results, and Figure \ref{fig:all_combined_each} shows the survey results for each condition.


\begin{figure*}[b]
    \centering

    \begin{subfigure}{0.45\linewidth}
        \centering
        \includegraphics[width=\linewidth]{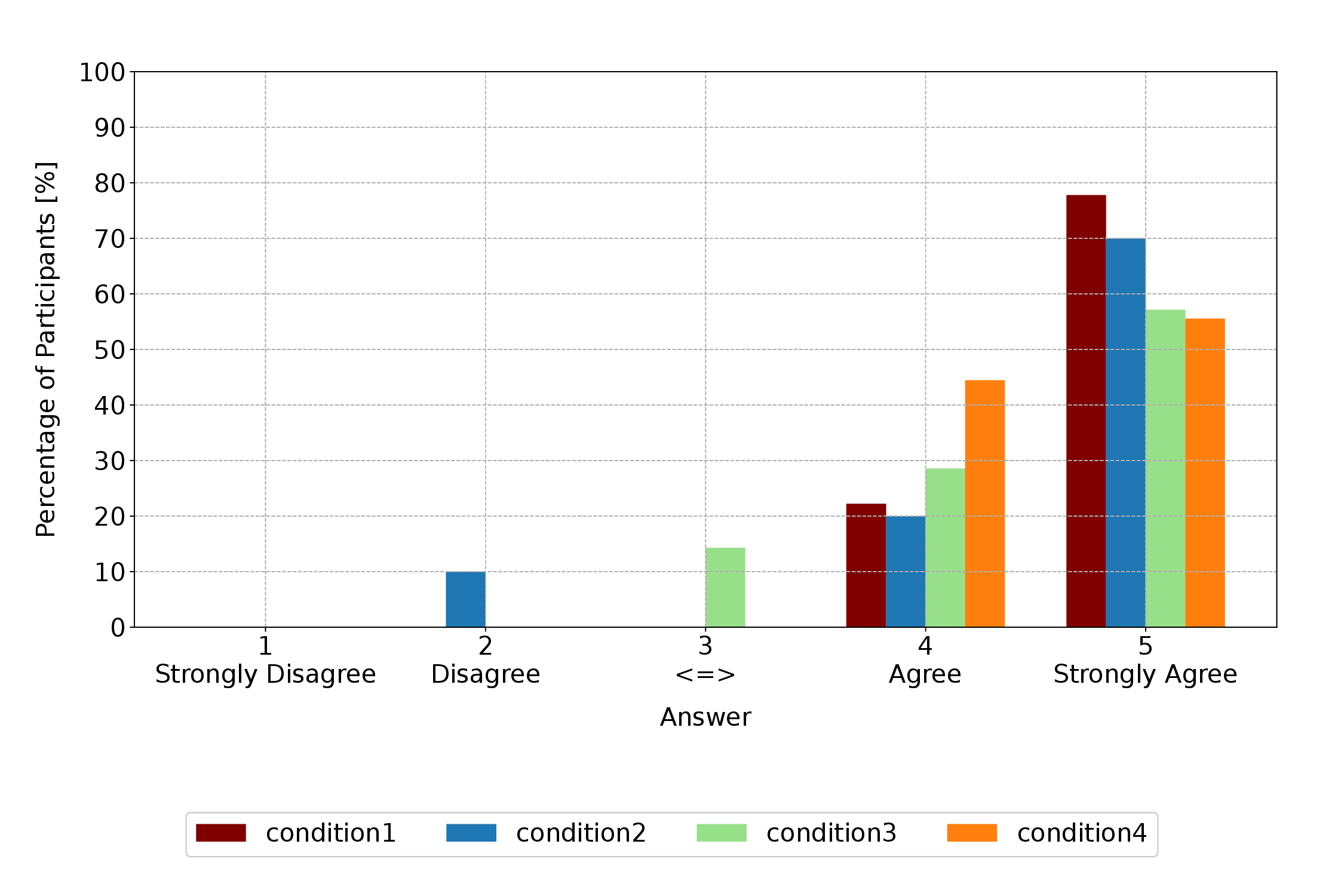}
        \caption{This robot is friendly}
        \label{re_na_a}
    \end{subfigure}
    \hfill
    \begin{subfigure}{0.45\linewidth}
        \centering
        \includegraphics[width=\linewidth]{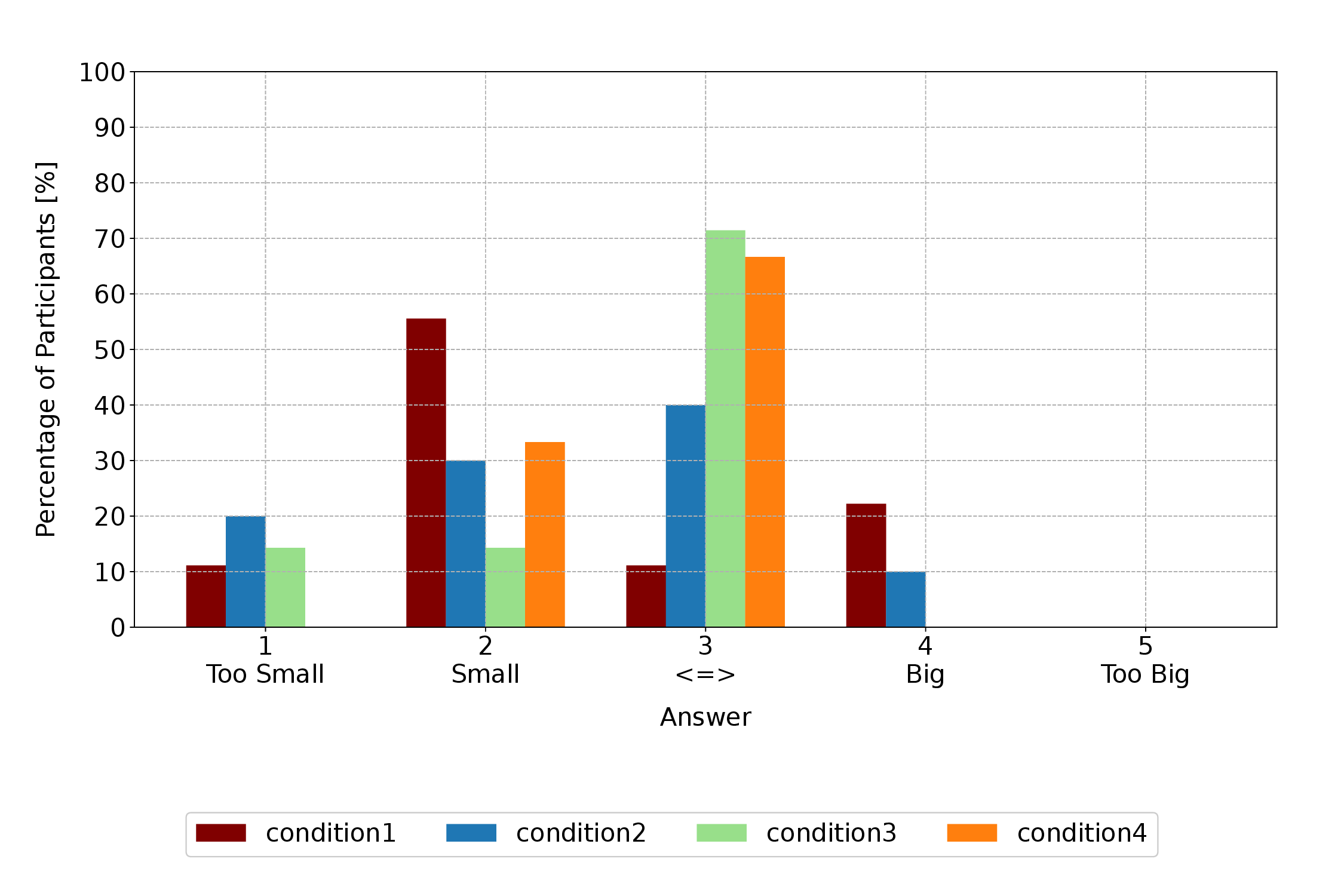}
        \caption{The size of this robot is}
        \label{re_na_b}
    \end{subfigure}

    \vspace{5mm}

    \begin{subfigure}{0.45\linewidth}
        \centering
        \includegraphics[width=\linewidth]{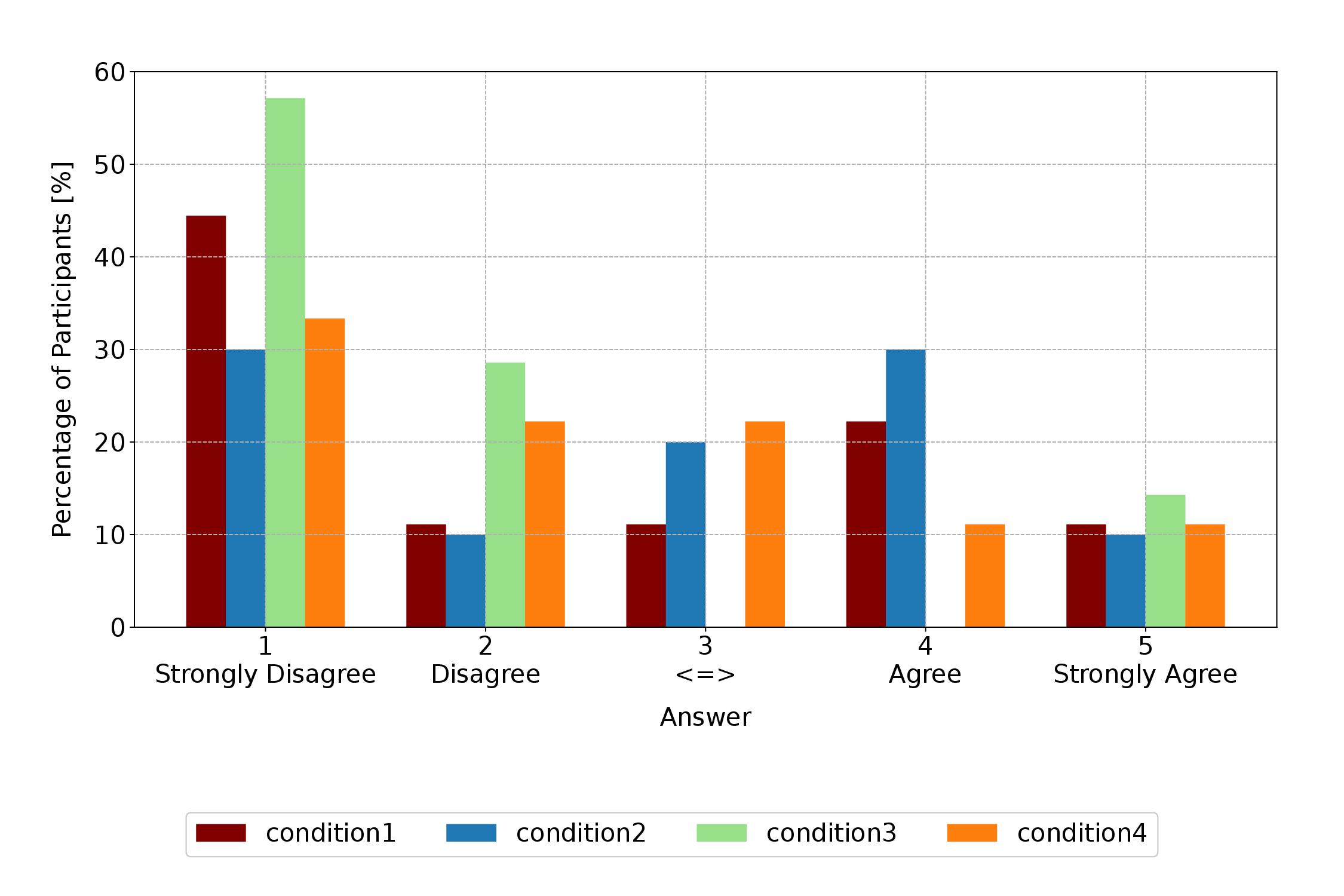}
        \caption{I want to touch this robot}
        \label{re_na_c}
    \end{subfigure}
    \hfill
    \begin{subfigure}{0.45\linewidth}
        \centering
        \includegraphics[width=\linewidth]{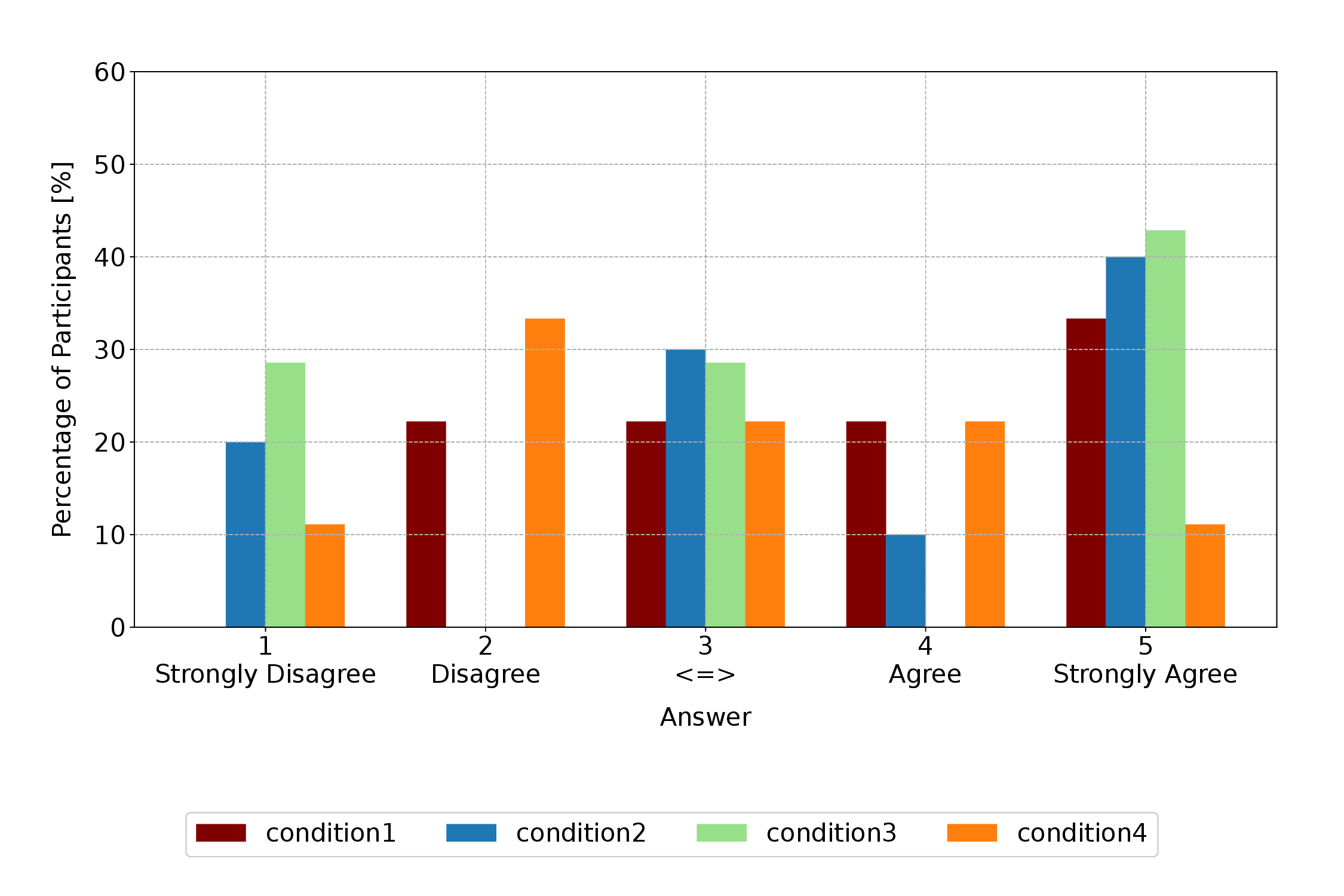}
        \caption{I want this robot to give me an attention}
        \label{re_na_d}
    \end{subfigure}

    \caption{Results of robot perception survey for each condition (1/2)}
    \label{fig:all_combined_each}
\end{figure*}

\begin{figure*}[t]
    \ContinuedFloat
    \centering

    \begin{subfigure}{0.45\linewidth}
        \centering
        \includegraphics[width=\linewidth]{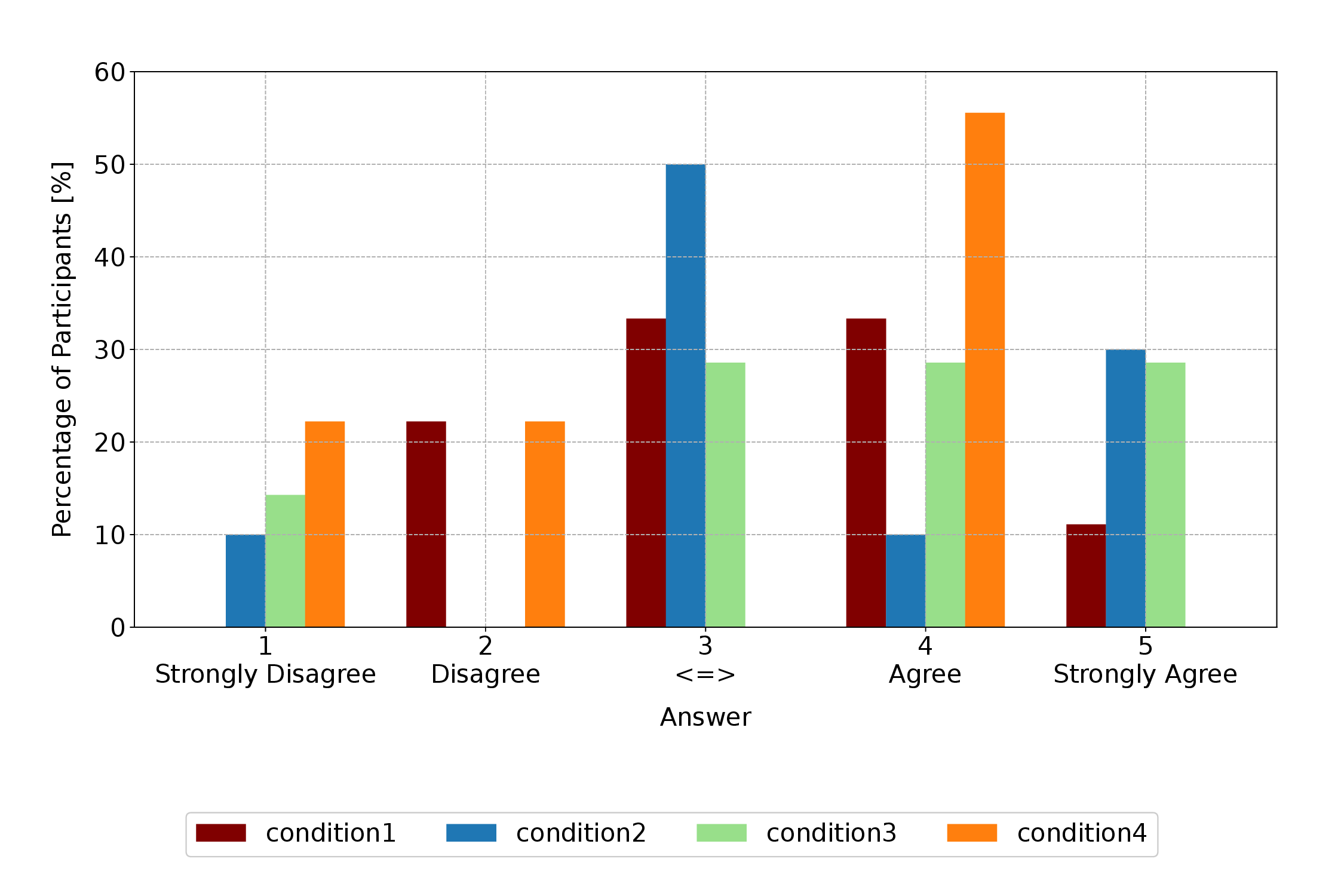}
        \caption{The way this robot speaks is natural}
        \label{re_na_e}
    \end{subfigure}
    \hfill
    \begin{subfigure}{0.45\linewidth}
        \centering
        \includegraphics[width=\linewidth]{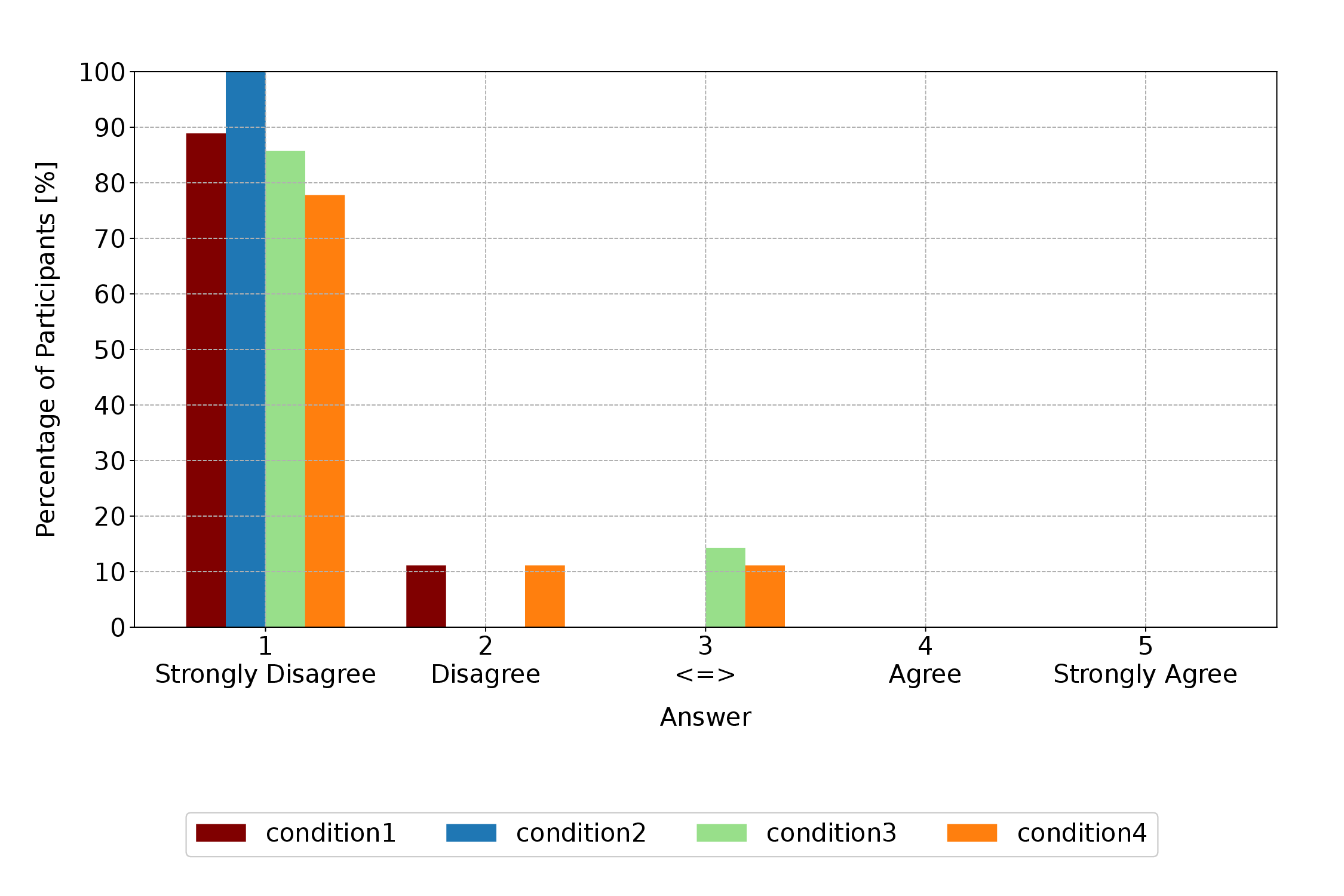}
        \caption{This robot is physically intrusive}
        \label{re_na_f}
    \end{subfigure}

    \vspace{5mm}

    \begin{subfigure}{0.45\linewidth}
        \centering
        \includegraphics[width=\linewidth]{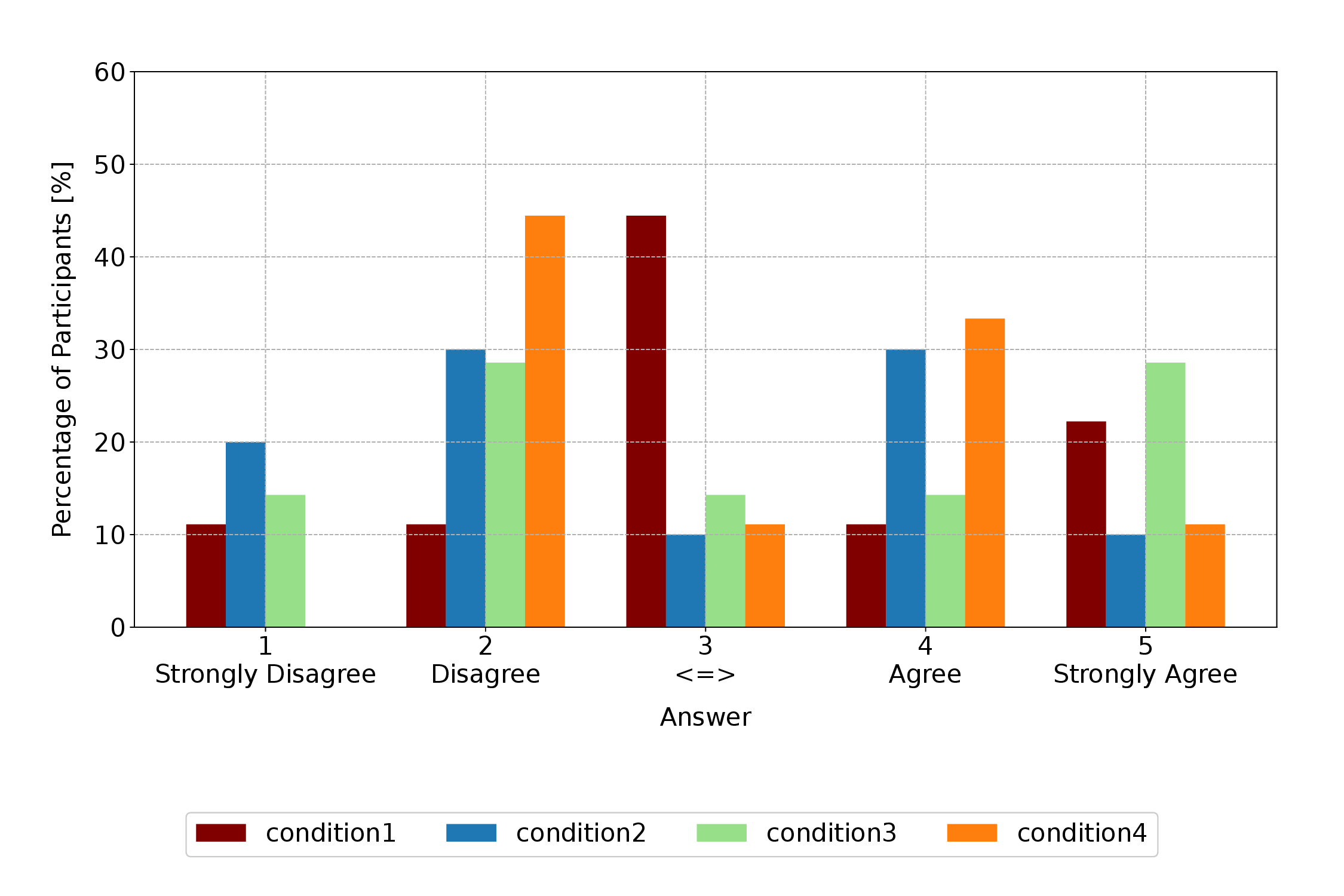}
        \caption{This robot is psychologically calming}
        \label{re_na_g}
    \end{subfigure}
    \hfill
    \begin{subfigure}{0.45\linewidth}
        \centering
        \includegraphics[width=\linewidth]{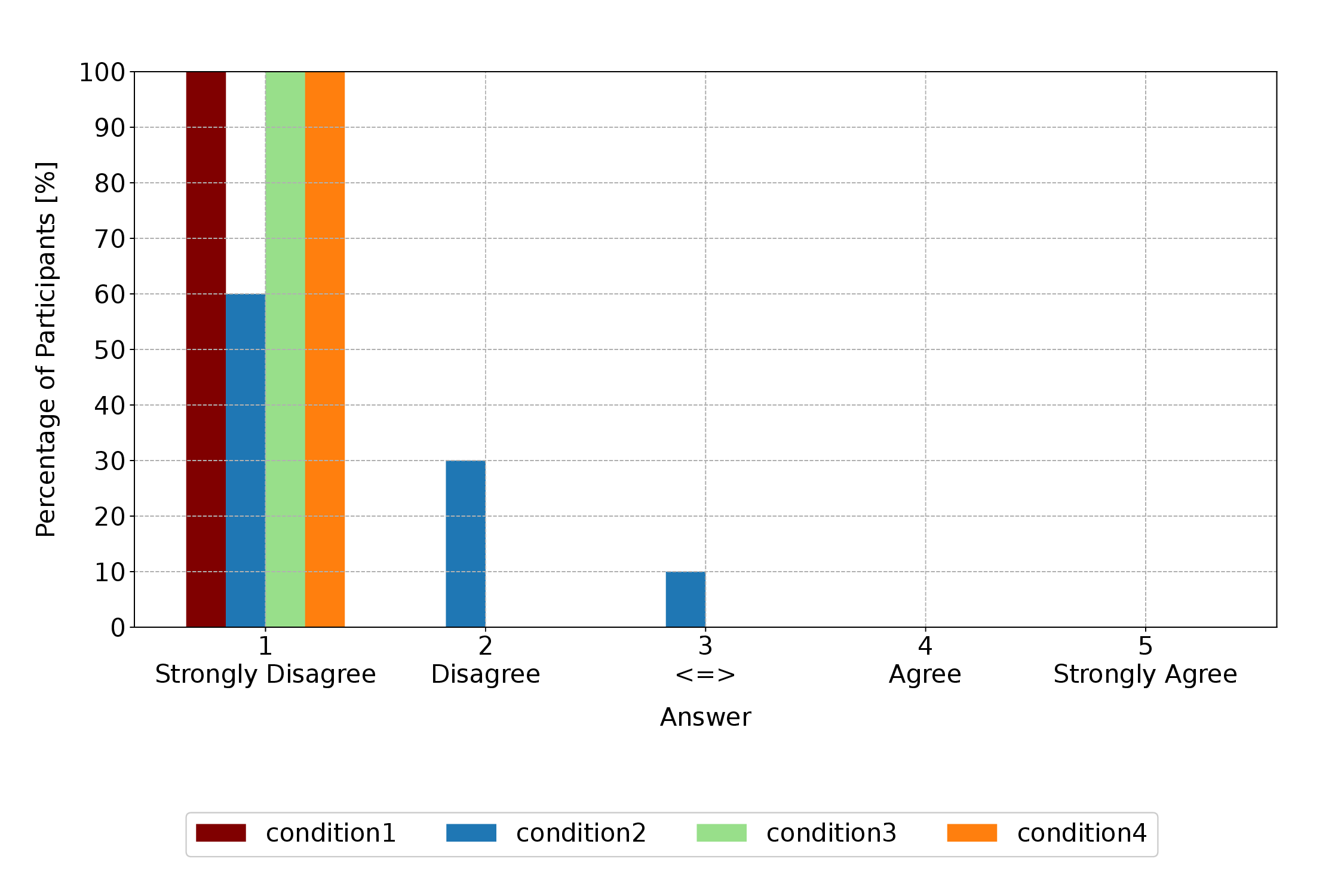}
        \caption{The sound of this robot moving is annoying}
        \label{re_na_h}
    \end{subfigure}

    \vspace{5mm}

    \begin{subfigure}{0.45\linewidth}
        \centering
        \includegraphics[width=\linewidth]{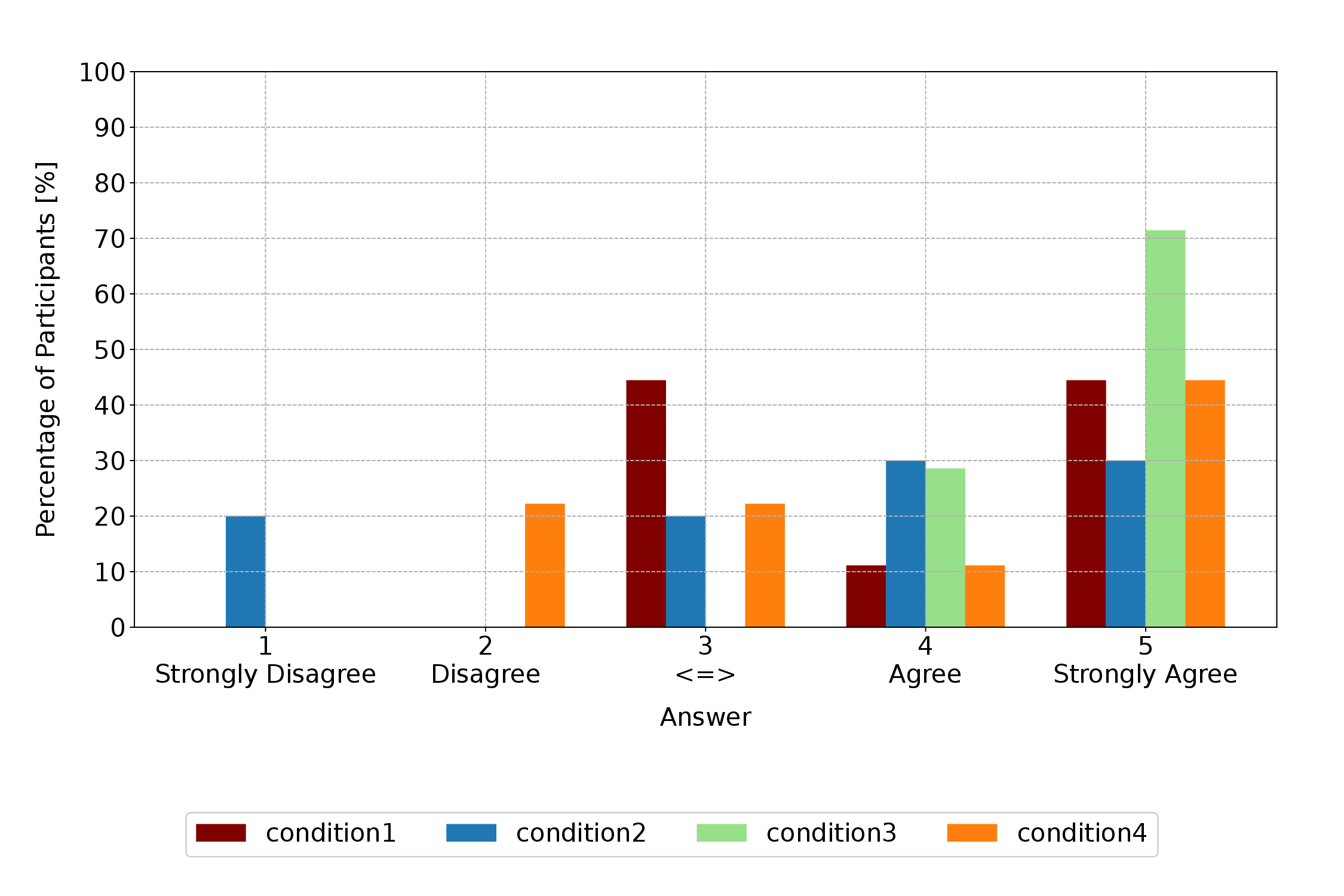}
        \caption{This robot seems safe}
        \label{re_na_i}
    \end{subfigure}
    \hspace*{\fill}

    \caption{Results of robot perception survey for each condition (2/2)}
\end{figure*}

\clearpage
\subsection{Results of Free Comments}\label{b_free}
All comments were answered in German. Here are the comments in English translation:\\

\noindent\textbf{Condition 1}
\begin{itemize}
\item problems with emphasis, happy, Emma should ask if she can be heard acoustically, or if she should speak louder or with headphones
\item has something childish, machine language, conversation with Emma not taken too seriously. Emma's eye reaction is very good \& beneficial for well-being. Good for routine things, very sensitive things: people cannot be replaced yet. Data protection?
\item one-sided, too much of a machine, no feeling that belongs
\item sweet, already seen on TV, variety with Emma in the retirement home
\item can't shake hands, VP wanted to say goodbye
\item very sweet, cute, pleasant size. Speaks clearly and loud enough for hearing aids
\item eye play was nice, funny, interesting. Hat is nice.
\item it bothered him that Emma closed her eyes, you don't know what she's thinking, Emma has a friendly aura, when she speaks she moves her lips, he doesn't know if that's effective, perhaps a rigid, friendly mouth is less distracting
\end{itemize}

\vspace{\baselineskip}
\noindent\textbf{Condition 2}
\begin{itemize}
\item limited repertoire of answers/reaction options
\item cold eyes and facial expression and eye movement, not very emotional when asked questions, you can tell that they are building blocks, sober
\item looks funny (rolls eyes)
\item previously imagined it to be more mechanical, it was more pleasant
\item quite funny, would prefer people. Depending on the situation, if I had no one (no specialist), yes. But it is not a substitute.
\item interesting, baby schema, she prefers people, has no function, not judgmental
\item would like a real conversation, "to communicate with her as a bit of a joke" -> a bit disappointed, easy to understand, clearly spoken
\item nice to have her in a nursing home. Great for mentally fit and interested people for interaction
\item voice well chosen, hat very nice, answered the right thing
\item don't know how Emma would deal with different answers, Emma was friendly and showed nice reactions even if the person didn't sound so enthusiastic, facial expressions positive, she would have liked gestures
\end{itemize}

\vspace{\baselineskip}
\noindent\textbf{Condition 3}
\begin{itemize}
\item insignificant, no real interaction possible, potential is there
\item very sweet and cute
\item other interactions with people, slowed conversation
\item belly is not nice (preferably closed), unfriendly when technology is visible
\item empathetic, optimistic that Emma can support, you don't feel alone
\item there is something formulaic, the longer he talks the longer Emma takes, thought she would be a bit bigger, impression that it is a toy, remains a technical device, interesting robot
\item could imagine contact depending on the situation (e.g. very lonely, that would be good), use in geriatrics is easy to imagine
\end{itemize}

\vspace{\baselineskip}
\noindent\textbf{Condition 4}
\begin{itemize}
\item not to be taken seriously, not a conversation partner
\item human form, eye/mouth movement -> fake, unnecessary, inauthentic
\item nice if she would smile, funny, the eyes are pleasant (it "thinks"), you could make eyes more human, e.g. light blue (easier to read mood), slightly smaller eyes
\item interesting, funny, curious, difficult for every day, more practical tasks can be imagined board games, good thing for support in care
\item compared to humans, he felt like he was on a hotline, had the feeling that he wasn't getting anywhere, no real contact with Emma, fundamental reservations about robots of this kind
\item size is good, not scary or creepy, I could well imagine Emma doing a survey/anamnesis, someone would have to evaluate it because a human is necessary, AI, closed eyes at the end is stupid
\end{itemize}
 
\end{appendices}

\bibliography{sn-bibliography}

@online{population,
  author      = "{United Nations, Department of Economic and Social Affairs, Population Division}",
  title       = "World Population Prospects 2024: Summary of Results (Advance Unedited version)",
  institution = "United Nations",
  year        = "2024",
  address     = "New York, NY, USA",
  isbn        = "9789210031691",
  note        = "eISBN: 9789211065138",
}

@online{aging,
  author       = {{Statistisches Bundesamt (Destatis)}},
  title        = {Pflege im Fokus: Zahl der Pflegekräfte steigt weiter an},
  year         = {2024},
  note          = "https://www.destatis.de/DE/Presse/\\Pressemitteilungen/2024/12/PD24\_478\\\_224.html",
  urldate      = {2025-09-05}
}

@online{demand_supply,
  author       = {{Statistisches Bundesamt (Destatis)}},
  title        = {Pflegekräftebedarf steigt deutlich bis 2049},
  year         = {2024},
  note          = "https://www.destatis.de/DE/Presse/\\Pressemitteilungen/2024/01/PD24\_033\\\_23\_12.html",
  urldate      = {2025-09-05}
}

@online{elderly_care,
  author       = {{Statistisches Bundesamt (Destatis)}},
  title        = {Medianentgelte in der Pflegebranche steigen überdurchschnittlich},
  year         = {2025},
  note          = "https://www.destatis.de/DE/Presse/\\Pressemitteilungen/Zahl-der-Woche/2025/\\PD25\_19\_p002.html",
  urldate      = {2025-09-05}
}

@article{backpain1,
  author  = "Tinubu, B. M. S. and Mbada, C. E. and Oyeyemi, A. L. and Fabunmi, A. A.",
  title   = "Work-Related Musculoskeletal Disorders among Nurses in Ibadan, South-west Nigeria: a cross-sectional survey",
  journal = "BMC Musculoskeletal Disorders",
  volume  = "11",
  pages   = "12",
  year    = "2010",
  doi     = "10.1186/1471-2474-11-12"
}

@article{backpain2,
  author = "Vinstrup, J. and Jakobsen, M. D. and Madeleine, P. and Andersen, L. L.",
  title  = "Physical exposure during patient transfer and risk of back injury \& low-back pain: prospective cohort study",
  journal= "BMC Musculoskeletal Disorders",
  year   = "2020",
  volume = "21",
  pages  = "715",
  doi    = "10.1186/s12891-020-03731-2"
}

@article{mental,
  author = {Abukari K. and Pammla M. P.},
  title = {A literature-based study of patient-centered care and communication in nurse-patient interactions: barriers, facilitators, and the way forward},
  journal = {BMC Nursing},
  volume = {20},
  number = {1},
  pages = {158},
  year = {2021},
  doi = {10.1186/s12912-021-00684-2},
  publisher = {Springer Nature}
}

@article{survey1,
  author  = "Flandorfer, P.",
  title   = "Population Ageing and Socially Assistive Robots for Elderly Persons: The Importance of Sociodemographic Factors for User Acceptance",
  journal = "International Journal of Population Research",
  volume  = "2012",
  pages   = "1--13",
  year    = "2012",
  doi     = "10.1155/2012/829835"
}

@article{survey2,
  author  = "Bemelmans, R. and Gelderblom, G. J. and Jonker, P. and de Witte, L.",
  title   = "Socially Assistive Robots in Elderly Care: A Systematic Review into Effects and Effectiveness",
  journal = "Journal of the American Medical Directors Association",
  volume  = "13",
  number  = "2",
  pages   = "114--120",
  year    = "2012",
  doi     = "10.1016/j.jamda.2010.10.002"
}

@article{robear1,
  author  = "Mukai, T.",
  title   = "Transfer assistance devices RIBA, ROBEAR",
  journal = "Journal of the Japan Society of Mechanical Engineers",
  volume  = "119",
  number  = "1166",
  pages   = "42",
  year    = "2016",
  doi     = "10.1299/jsmemag.119.1166_42"
}

@inproceedings{robear2,
  author    = "Mukai, T. and Toki, S. and Obinata, G. and others",
  title     = "Development of High-functionality Nursing-care Assistant Robot ROBEAR for Patient-transfer and Standing Assistance",
  booktitle = "Proceedings of the IEEE/RSJ International Conference on Intelligent Robots and Systems",
  year      = "2017"
}

@article{lio,
  author		= "Justinas Mi{\v{s}}eikis",
  title		= "Lio-A Personal Robot Assistant for Human-Robot Interaction and Care Applications",
  journal		= "IEEE Robotics and Automation Letters",
  volume		= "5",
  issue       = "4",
  pages		= "5339--5346",
  year		= "2020",
  doi			= "10.1109/LRA.2020.3007462"
}

@article{paro2,
  author		= "Shibata, T.",
  title		= "Therapeutic Seal Robot as Biofeedback Medical Device: Qualitative and Quantitative Evaluations of Robot Therapy in Dementia Care",
  journal		= "Proceedings of the IEEE",
  volume		= "100",
  issue       = "8",
  pages		= "2527--2538",
  year		= "2012",
  doi			= "10.1109/JPROC.2012.2200559"
}

@article{paro3,
  author		= "Wada, K. and Shibata, T. and Saito, T. and Sakamoto, K. and Tanie, K",
  title		= "Psychological and Social Effects of One Year Robot Assisted Activity on Elderly People at a Health Service Facility for the Aged",
  year		= "2005",
  doi			= "10.1109/ROBOT.2005.1570535",
  note		= "Paper presented at Proceedings of the 2005 IEEE International Conference on Robotics and Automation, Barcelona, Spain, 18--22 April 2005"
}

@article{paro4,
  author		= " Šabanović, S. and Bennett, C.C. and Chang, W.-L. and Huber, L.",
  title		= "PARO robot affects diverse interaction modalities in group sensory therapy for older adults with dementia",
  year		= "2013",
  doi			= "10.1109/ICORR.2013.6650427",
  note		= "Paper presented at 2013 IEEE 13th International Conference on Rehabilitation Robotics (ICORR), Seattle, WA, USA, 24--26 June 2013"
}

@article{paro5,
  author		= "Bemelmans, R. and Gelderblom, G. J. and Jonker, P. and de Witte, L.",
  title		= "Effectiveness of Robot Paro in Intramural Psychogeriatric Care: A Multicenter Quasi-Experimental Study",
  journal		= "The Society for Post-Acute and Long-Term Care Medicine",
  volume		= "16",
  issue       = "11",
  pages		= "946--950",
  year		= "2015",
  doi			= "10.1016/j.jamda.2015.05.007"
}

@article{qol,
  author		= "Guevara, J. and Fernandez, M. and Balbuena, J. and Arroyo, D. and Davila, M. and Arce, D.",
  title		= "The Role of Socially Assistive Robots for the Improvement of Quality of Life in Elderly People: A Systematic Review",
  journal		= "IEEE Access",
  volume		= "14",
  pages		= "2431--2450",
  year		= "2026",
  doi			= "10.1109/ACCESS.2025.3649801"
}

@article{healthcare,
  author		= "Fasola, J. and Mataric, M. J.",
  title		= "Using Socially Assistive Human–Robot Interaction to Motivate Physical Exercise for Older Adults",
  journal		= "Proceedings of the IEEE",
  volume		= "100",
  issue       = "8",
  pages		= "2512--2526",
  year		= "2012",
  doi			= "10.1109/JPROC.2012.2200539"
}

@article{trust,
  author		= "Romeo, M. and Torre, I. and Le Maguer, S. and Sleat, A. and Cangelosi, A. and Leite, I.",
  title		= "The Effect of Voice and Repair Strategy on Trust Formation and Repair in Human-Robot Interaction",
  journal		= "ACM Transactions on Human-Robot Interaction",
  volume		= "14",
  issue       = "2",
  pages		= "1--22",
  year		= "2025",
  doi			= "10.1145/3711938"
}

@article{engagement,
  author		= "Moshikina, L. and Trickett, S. and Trafton, J.",
  title		= "Social Engagement in Public Places: A Tale of One Robot ",
  journal		= "HRI'14",
  pages		= "382--389",
  year		= "2014",
  doi			= "10.1145/2559636.2559678"
}

@article{appearance,
  author		= "Tuncer, S. and Licoppe, C. and Luff, P. and Heath, C.",
  title		= "Recipient design in human–robot interaction: the emergent assessment of a robot’s competence",
  journal		= "AI \& SOCIETY",
  volume		= "39",
  issue       = "4",
  year		= "2023",
  doi			= "10.1007/s00146-022-01608-7"
}

@article{appearance2,
  author		= "Barnes, J. and FakhrHosseini, M. and Jeon, M. and Park, C. and Howard, A.",
  title		= "The influence of robot design on acceptance of social robots",
  journal		= "2017 14th International Conference on Ubiquitous Robots and Ambient Intelligence (URAI)",
  pages		= "51--55",
  year		= "2017",
  doi			= "10.1109/URAI.2017.7992883"
}

@article{age,
  author  = "Chien, SE. and Chu, L. and Lee, HH. and Yang, CC. and Lin, FH. and Yang, PL. and Wang, TM. and Yeh, SL.",
  title   = "Age Difference in Perceived Ease of Use, Curiosity, and Implicit Negative Attitude toward Robots",
  journal = "ACM Transactions on Human-Robot Interaction",
  volume  = "8",
  number  = "2",
  pages   = "1--19",
  year    = "2019",
  doi     = "10.1145/3311788"
}

@article{elderly,
  author  = "Heerink, M. and Kröse, B. and Evers, V. and Wielinga, B.",
  title   = "Assessing Acceptance of Assistive Social Agent Technology by Older Adults: the Almere Model",
  journal = "International Journal of Social Robotics",
  volume  = "2",
  pages   = "361--375",
  year    = "2010",
  doi     = "10.1007/s12369-010-0068-5",
  Publisher = {Springer Nature}
}

@article{nursing_home,
  author  = "Rettinger, L. and Fürst, A. and Kupka-Klepsch, E. and Mühlhauser, K. and Haslinger-Baumann, E. and Werner, F. ",
  title   = "Observing the Interaction between a Socially-Assistive Robot and Residents in a Nursing Home",
  journal = "International Journal of Social Robotics",
  volume  = "16",
  pages   = "403--413",
  year    = "2023",
  doi     = "10.1007/s12369-023-01088-9",
  Publisher = {Springer Nature}
}

@article{trust2,
  author		= "Cho, M and Kim, D. and Jang, M. and Lee, J. and Kim, J. and Yun, W. and Yoon, Y. and Jang, J. and Park, C. and Ko, W. and Jang, J. and Yoon, H. and Lee, D. and Jang, C.",
  title		= "Evaluating Human-Care Robot Services for the Elderly: An Experimental Study",
  journal		= "International Journal of Social Robotics",
  volume		= "16",
  pages		= "1561--1587",
  year		= "2024",
  doi			= "10.1007/s12369-024-01157-7",
  Publisher = {Springer Nature}
}

@article{HRI,
  author = {Bartosz Sawik and S{\l}awomir Tobis and Ewa Baum and Aleksandra Suwalska and Katarzyna Stachnik and Marta Cildoz and Alba Agustin and Katarzyna Wieczorowska-Tobis},
  title = {Robots for Elderly Care: Review, Multi-Criteria Optimization Model and Qualitative Case Study},
  journal = {Healthcare},
  volume = {11},
  number = {9},
  pages = {1286},
  year = {2023},
  doi = {10.3390/healthcare11091286},
  publisher = {MDPI}
}

@article{positive_prompt,
  author		= "Eckstein, M. and Stoßel, G. and Gerchen, M. F. and Bilek, E. and Kirsch, P. and Ditzen, B.",
  title		= "Neural responses to instructed positive couple interaction: an fMRI study on compliment sharing",
  journal		= "Social Cognitive and Affective Neuroscience",
  volume		= "100",
  issue       = "18(1)",
  pages		= "1--9",
  year		= "2023",
  doi			= "10.1093/scan/nsad005"
}

@article{mayer,
  author		= "Mayer, C. J. and Raithel, C. and Yamamoto, H. and Buchner, T. and Iwan, B. S. V. and Tempel, A. and Staatz, E. and Schommer, F. and Schmetterer, M. and Misok, D. and Werner, C. and Ditzen, B. and Mombaur, K. and Eckstein, M.",
  title		= "Trust, stress, and oxytocin: Psychophysiological responses of older adults to social robot interactions",
  journal		= "Under submission",
  year		= "2026",
  note    	= "Submitted for publication"
}

@article{navel,
  author		= "Nonoda, S. and Okada, K.",
  title		= "Navel - a social robot with verbal and nonverbal communication skills",
  year		= "2023",
  doi			= "10.1145/3544549.3583898",
  note			= "Paper presented at CHI '23: CHI Conference on Human Factors in Computing Systems, Hamburg, Germany, 23--28 April 2023"
}

@article{navel2,
  author  = "Nardelli, A. and Maccagni, G. and Minutoli, F. and Sgorbissa A. and Recchiuto, C.",
  title   = "Towards Intuitive Interaction: Cognitive Architecture for Artificial Personality, Emotional Intelligence, and Cognitive Capabilities",
  journal = "International Journal of Social Robotics",
  volume  = "17",
  pages   = "2211--2228",
  year    = "2025",
  doi     = "10.1007/s12369-025-01260-3",
  Publisher = {Springer Nature}
}

@article{emotion1,
  author  = "Faria, Diego R. and Vieira, Mario and Faria, Fernanda C.C. and Premebida, Cristiano",
  title   = "Affective facial expressions recognition for human-robot interaction",
  journal = "2017 26th IEEE International Symposium on Robot and Human Interactive Communication (RO-MAN)",
  pages   = "805--810",
  year    = "2017",
  doi     = "10.1109/ROMAN.2017.8172395",
}

@article{emotion2,
  author  = "Green, H. N. and Iqbal, T.",
  title   = "Examining Physiological Response and Facial Expression as Indicators of Trust in a Robot Partner",
  journal = "ACM Transactions on Human-Robot Interaction",
  volume  = "32",
  issue   = "2",
  pages   = "1--24",
  year    = "2025",
  doi     = "10.1145/3773895",
}

@article{heartrate2,
  author  = "Lei, G. and Cheng, W. and Yin, X. and Wu, Y.",
  title   = "The dataset of multi-target vital signs monitored by FMCW radar",
  journal = "Data in Brief",
  volume  = "57",
  pages   = "111027",
  year    = "2024",
  doi     = "10.1016/j.dib.2024.111027",
}

@article{heartrate3,
  author  = "Hassanpour, A. and Yang, B.",
  title   = "Contactless Vital Sign Monitoring: A Review Towards Multi-Modal Multi-Task Approaches",
  journal = "Sensors 2025",
  volume  = "25(15)",
  pages   = "4792",
  year    = "2025",
  doi     = "10.3390/s25154792",
}

@article{facenet,
  author		= "Chen, S. and Liu, Y. and Gao, X. and Han, Z.",
  title		= "MobileFaceNets: Efficient CNNs for Accurate Real-Time Face Verification on Mobile Devices",
  year		= "2018",
  booktitle 	= "Biometric Recognition",
  pages	    = "428--438",
  doi			= "10.1007/978-3-319-97909-0_46",
}

@article{likert,
  author		= "Nonoda, S. and Okada, K.",
  title		= "Psychological assessments of paired-comparison and Likert formats:Comparing their reliability, validity, and ease of answering questions",
  journal		= "The Japanese Journal of Psychology",
  volume		= "93",
  issue       = "6",
  pages		= " 526--535",
  year		= "2023",
  doi			= "10.4992/jjpsy.93.20231"
}

@article{multi,
  author		= "Bethel, C. L. and Salomon, K. and Murphy, R. R. and Burke, J. L.",
  title		= "Survey of Psychophysiology Measurements Applied to Human-Robot Interaction",
  journal		= "RO-MAN 2007 - The 16th IEEE International Symposium on Robot and Human Interactive Communication",
  pages		= "732--737",
  year		= "2007",
  doi			= "10.1109/ROMAN.2007.4415182."
}

@article{multi2,
  author		= "Kidd, C. D. and Breazeal, C.",
  title		= "Human-Robot Interaction Experiments: Lessons learned",
  pages		= "141--142",
  year		= "2005",
}

@article{MMSE,
  author		= "Folstein, M. F.and Folstein, S. E. and McHugh, P. R.",
  title		= "''Mini-mental state''. A practical method for grading the cognitive state of patients for the clinician",
  journal		= "Journal of Psychiatric Research",
  volume		= "12",
  issue       = "3",
  pages		= "189--198",
  year		= "1975",
  doi			= "10.1016/0022-3956(75)90026-6",
}

@article{CATS,
  author		= "Heinssen, Jr. and Robert K. and Glass, C. R. and Knight, L. A.",
  title		= "Assessing computer anxiety: Development and validation of the Computer Anxiety Rating Scale",
  journal		= "Computers in Human Behavior",
  volume		= "3",
  issue       = "1",
  pages		= "49--59",
  year		= "1987",
  doi			= "10.1016/0747-5632(87)90010-0"
}

@article{PANAS,
  author		= "Watson, D. and Clark, L. A. and Tellegen, A.",
  title		= "Development and validation of brief measures of positive and negative affect: The PANAS scales",
  journal		= "Journal of Personality and Social Psychology",
  volume		= "54",
  issue       = "6",
  pages		= "1063--1070",
  year		= "1988",
  doi			= "10.1037/0022-3514.54.6.1063"
}

@article{GDS,
  author		= "Almeida, O. P. and Almeida, S. A.",
  title		= "Short versions of the geriatric depression scale: a study of their validity for the diagnosis of a major depressive episode according to ICD-10 and DSM-IV",
  journal		= "International journal of geriatric psychiatry",
  volume		= "14",
  issue       = "10",
  pages		= "858--865",
  year		= "1999",
  note		= "https://doi.org/10.1002/(SICI)1099-1166(199910)14:10<858::AID-GPS35>3.0.CO;2-8"
}

@article{LSNS,
  author		= "Lubben, J. and Blozik, E. and Gillmann, G. and Iliffe, S. and von Renteln Kruse, W. and Beck, J. C. and Stuck, A. E.",
  title		= "Performance of an abbreviated version of the Lubben Social Network Scale among three European community-dwelling older adult populations",
  journal		= "The Gerontologist",
  volume		= "46",
  issue       = "4",
  pages		= "503--513",
  year		= "2006",
  doi			= "10.1093/geront/46.4.503"
}

@article{hrv1,
  author		= "Taelman, J.and Vandeput, S. and Spaepen, A. and Van Huffel, S. ",
  title		= "Influence of Mental Stress on Heart Rate and Heart Rate Variability",
  journal		= "4th European Conference of the International Federation for Medical and Biological Engineering",
  pages		= "1366–1369",
  year		= "2009",
  doi			= "10.1007/978-3-540-89208-3_324",
  publisher   = "Springer"
}

@article{hrv2,
  author		= "Kim, H. G. and Cheon, E. J. and Bai, D. S. and Lee, Y. H. and Koo, B. H.",
  title		= "Stress and Heart Rate Variability: A Meta-Analysis and Review of the Literature",
  journal		= "Psychiatry investigation",
  volume		= "15",
  issue       = "3",
  pages		= "235--245",
  year		= "2018",
  doi			= "10.30773/pi.2017.08.17"
}

@article{HR,
  author		= "Thayer, J. F. and Lane, R. D.",
  title		= "A model of neurovisceral integration in emotion regulation and dysregulation",
  journal		= "Journal of Affective Disorders",
  volume		= "61",
  issue       = "3",
  pages		= "201--216",
  year		= "2000",
  doi			= "10.1016/S0165-0327(00)00338-4"
}

@article{oz,
  author		= "Sequeira, P. and Alves-Oliveira, P. and Ribeiro, T. and Di Tllio, E. and Petisca, S. and Melo, F. S. and Castellano, G. and Paiva, A.",
  title		= "Discovering social interaction strategies for robots from restricted-perception Wizard-of-Oz studies",
  journal		= "2016 11th ACM/IEEE International Conference on Human-Robot Interaction (HRI)",
  pages		= "197--204",
  year		= "2016",
  doi			= "10.1109/HRI.2016.7451752"
}

@article{happy1,
  author		= "Hess, U. and Blairy, S. and Kleck, R. E.",
  title		= "The Influence of Facial Emotion Displays, Gender, and Ethnicity on Judgments of Dominance and Affiliation",
  journal		= "Journal of Nonverbal Behavior",
  volume		= "24",
  pages		= "265–-283",
  year		= "2000",
  doi			= "10.1023/A:1006623213355",
  publisher   = "Springer Nature"
}

@article{happy2,
  author		= "Hess, U. and Fischer, A.",
  title		= "Emotional Mimicry as Social Regulation.",
  journal		= "Personality and Social Psychology Review",
  volume		= "17",
  issue       = "2",
  pages		= "142-157",
  year		= "2013",
  doi			= "10.1177/1088868312472607"
}

@article{negative1,
  author		= "Hömke, P. and Levinson, S. C. and Emmendorfer, A. K. and Holler, J.",
  title		= "Eyebrow movements as signals of communicative problems in human face-to-face interaction",
  journal		= "Royal Society open science",
  volume		= "12",
  issue       = "3",
  pages		= "241632",
  year		= "2025",
  doi			= "10.1098/rsos.241632"
}

@article{negative2,
  author		= "Grafsgaard, J. and Wiggins, J. and Boyer, K. and Wiebe, E. and Lester, J.",
  title		= "Automatically Recognizing Facial Indicators of Frustration: A Learning-Centric Analysis",
  journal		= "Proceedings - 2013 Humaine Association Conference on Affective Computing and Intelligent Interaction, ACII 2013",
  pages		= "159--165",
  year		= "2013",
  doi			= "10.1109/ACII.2013.33"
}

@article{prompt1,
  author		= "Shi, W. and Li, Y. and Sahay, S. and Yu, Z.",
  title		= "Refine and Imitate: Reducing Repetition and Inconsistency in Persuasion Dialogues via Reinforcement Learning and Human Demonstration",
  journal		= "EMNLP (Findings)",
  pages		= "3478-3492",
  year		= "2021",
  doi			= "10.48550/arXiv.2012.15375"
}

@article{prompt2,
  author		= "Yang, J. and Kikuchi, H. and Uegaki, T. and Kikuchi, H.",
  title		= "The Effect of the Repetitive Utterances Complexity on User’s Perceived Empathy and Desire to Continue Dialogue by a Chat-oriented Dialogue System",
  journal		= "Proceedings of the 9th International Conference on Human-Agent Interaction (HAI ’21)",
  pages		= "241--244",
  year		= "2021",
  doi			= "10.1145/3472307.3484651"
}

@article{prompt3,
  author		= "Hu, Y. and Qu, Y. and Maus, A. and Mutlu, B.",
  title		= "Polite or Direct? Conversation Design of a Smart Display for Older Adults Based on Politeness Theory",
  journal		= "CHI '22: Proceedings of the 2022 CHI Conference on Human Factors in Computing Systems",
  volume		= "307",
  pages		= "1--15",
  year		= "2022",
  doi			= "10.1145/3491102.3517525"
}

@article{match,
  author		= "Goetz, J. and Kiesler, S. and Powers, A.",
  title		= "Matching robot appearance and behavior to tasks to improve human-robot cooperation",
  journal		= "The 12th IEEE International Workshop on Robot and Human Interactive Communication, 2003. Proceedings. ROMAN 2003.",
  pages		= "55--60",
  year		= "2003",
  doi			= "10.1109/ROMAN.2003.1251796"
}

\end{document}